%% file: main.tex
\documentclass{article} 
\usepackage{arxiv_v1,times}

\usepackage{hyperref}
\usepackage{url}

\usepackage{amsmath}
\usepackage{amssymb}
\usepackage{amsthm}
\usepackage{booktabs}
\usepackage{graphicx}
\usepackage{tikz}
\usepackage{wrapfig}
\usepackage{subcaption}
\usepackage{enumitem}
\usepackage{multirow}

\AtBeginDocument{
    \setlength{\abovedisplayskip}{0.5pt}
    \setlength{\belowdisplayskip}{0.5pt}
    \setlength{\abovedisplayshortskip}{0.5pt}
    \setlength{\belowdisplayshortskip}{0.5pt}
}

\usepackage{titlesec}
\titlespacing{\paragraph}{0pt}{0.01\baselineskip}{0.5em}

\newtheorem{definition}{Definition}
\newtheorem{theorem}{Theorem}
\newtheorem{lemma}{Lemma}
\newtheorem{proposition}{Proposition}
\newtheorem{corollary}{Corollary}
\newtheorem{assumption}{Assumption}

\title{Where Should Learning Live? Regret Analysis for Text, Parameter, and Joint Adaptation}

\author{Jungtaek Kim$^{1}$\qquad
Jaden Park$^{2}$\qquad
Jongwon Jeong$^{2}$\\
\textbf{Emmanouil-Vasileios Vlatakis-Gkaragkounis}$^{2,}$\thanks{Co-corresponding authors.}\qquad
\textbf{Kangwook Lee}$^{3,4,}$\footnotemark[1]\\
$^{1}$University of Arkansas\quad
$^{2}$University of Wisconsin--Madison\quad
$^{3}$KRAFTON\quad
$^{4}$Ludo Robotics
}

\iclrfinalcopy
\begin{document}

\maketitle

\begin{abstract}
Language-model systems can select prompts and parameter adapters repeatedly during deployment, using feedback to guide later choices. Earlier choices may change the context or environment inherited by later configurations. We study selection among preconstructed candidates and compare cumulative loss with the best fixed prompt--parameter pair used from the beginning. This excess loss separates into three terms: selection regret within the available family, the gap from restricting that family, and the benefit or harm of the actual deployment history. Selection is evaluated using the history each candidate would have produced if used throughout, which separates candidate quality from the effects of switching. When loss functions are fixed independently of selector randomness, past choices affect loss for only a bounded number of rounds, and every candidate's own-history loss is revealed, limiting switches yields sublinear expected regret within the selected family. We also quantify the costs and additional assumptions of independent probing and deployment-only feedback. Controlled experiments show that fewer switches can either lower or raise deployed loss, depending on whether inherited histories are harmful or helpful. With Qwen3.5-9B on Spider, text-only selection has lower within-family regret than joint selection, yet produces more failures because its candidate family excludes stronger configurations.
\end{abstract}

\section{Introduction}

Language-model agents edit code, execute tests, and use tool feedback~\citep{YangJ2024neurips}, while recent benchmarks study learning across sequences of related tasks~\citep{continuallearningbench2026}.
Consider a coding agent that handles successive tasks in a persistent workspace.
Before each task, a selector chooses an instruction prompt and a model checkpoint or adapter.
The agent edits code, runs tools, and receives feedback such as whether its changes pass the task's tests.
Feedback guides the next configuration choice, while the edited code and tool history remain in the workspace.
The next configuration therefore inherits the consequences of earlier choices.
\emph{How should we compare text-only, parameter-only, and joint adaptation in such a system?}

Each opportunity to select and execute a configuration is a \emph{deployment round}, indexed by $t$.
For example, a round can be one coding task, one question in Spider~\citep{YuT2018emnlp}, or one dialogue turn in SParC~\citep{YuT2019acl}.
We consider a sequence of $T$ rounds, which can share state and need not be independent.
The selector uses earlier feedback to choose a configuration, executes it, incurs a loss, and uses the available feedback to update future selection probabilities.
Minimizing cumulative loss over these repeated choices gives an online learning problem~\citep{cesabianchi2006prediction}.
Recent methods construct configurations through reflective prompt optimization~\citep{agrawal2026gepa}, joint prompt--parameter optimization~\citep{BoX2026iclr}, and alternating harness--weight optimization~\citep{KimH2026whale}.
In our setting, candidates are constructed before evaluation.
During evaluation, online learning updates their selection probabilities without constructing new prompts or adapters.

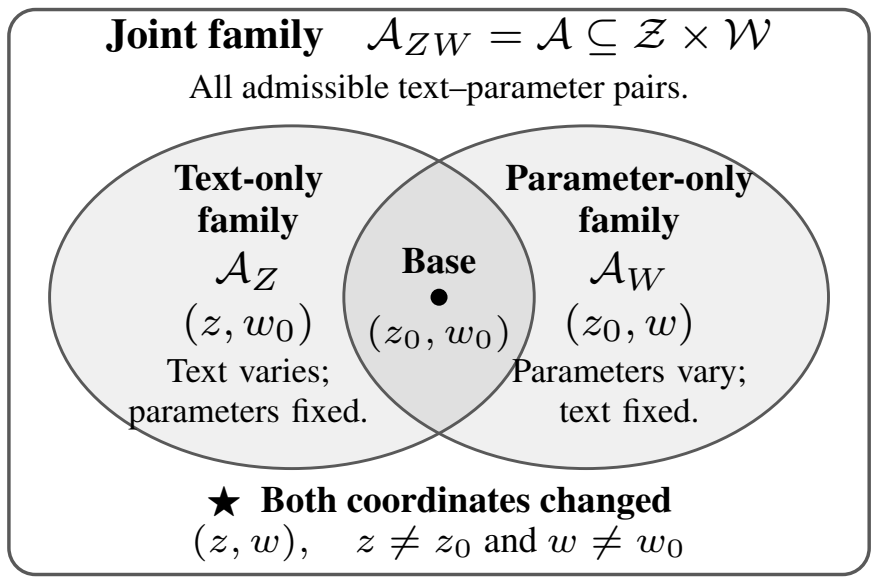
\begin{wrapfigure}{R}{0.41\textwidth}
    \vspace{-15pt}
    \centering
    \input{figures/adaptation_families}
    \vspace{-15pt}
    \caption{Text-only and parameter-only families within the joint family.}
    \label{fig:adaptation-families}
    \vspace{-20pt}
\end{wrapfigure}

A \emph{configuration} pairs reusable text, such as prompts, demonstrations, and retrieval contents, with deployable parameters, such as checkpoints or adapters~\citep{brown2020language,lewis2020rag,hu2022lora}.
Text-only selection keeps the base parameters, parameter-only selection keeps the base text, and joint selection can change both.
Figure~\ref{fig:adaptation-families} illustrates these configuration families.
The workspace, generated history, and tool sessions constitute \emph{operational state}, which evolves even under a fixed configuration.
Retaining feedback across trials can influence later behavior without changing model parameters~\citep{shinn2023reflexion}.

Our comparator is the best available joint configuration used for all $T$ rounds from the common initial condition, with its own resulting history.
This own-history comparison follows the policy-regret perspective~\citep{arora2012policy}.
For example, a coding agent that switches to configuration $b$ after using $a$ can encounter different code from an agent that used $b$ throughout.

This comparison raises three separate questions.
(i) \emph{How well does the selector choose among its available candidates, when each is evaluated on the history it would have produced if used throughout?}
(ii) \emph{How much better is the best joint candidate than the best candidate in the selected family?}
(iii) \emph{Does the actual sequence of choices create a more or less favorable history than those fixed-configuration evaluations?}
We call the corresponding terms \emph{diagonal selection regret}, \emph{family restriction}, and the \emph{signed state effect}.
Their sum is exactly the deployed excess loss.
\emph{Diagonal} refers to repeating a single configuration throughout the history,
and \emph{signed} emphasizes that inherited state can reduce as well as increase loss.

Our strongest feedback model evaluates every candidate on its own fixed-configuration history.
Ordinary deployment feedback need not supply those counterfactual losses.
We also study independent probes of individual candidates and updates based only on deployed outcomes.
For state-dependent guarantees, we require \emph{bounded influence}: choices sufficiently far in the past no longer affect current loss, including through hidden state.
These assumptions connect our analysis to online learning with memory~\citep{anava2015memory}.
We also compare joint sampling and coordinatewise conditional selection: they agree for additive losses under a product prior, but interactions can separate their regret even when both eventually select the optimum.
The online-learning methods are established~\citep{geulen2010regret,AvnerO2012icml,ChengD2023icml}.
Our analysis connects their guarantees to adaptation comparators, information requirements, and signed state effects.

Controlled studies show that reducing switches helps when mixed histories are harmful and hurts when they are beneficial, even with the same per-round selection probabilities.
With Qwen3.5-9B~\citep{qwen35modelcard} on Spider, text-only selection incurs less within-family regret than joint selection but more deployed loss because its candidates are weaker.
On recursive SParC, earlier generated queries in Structured Query Language (SQL) supply the inherited state.
Lazy exponential weights (EW) substantially reduces switching with only a small change in estimated expected loss.

Our main contributions are summarized as follows.
\begin{itemize}[leftmargin=*]
    \item \emph{Loss accounting.}
We separate selection, family restriction, and state effects relative to a common fixed joint comparator; see Section~\ref{sec:protocol}.
    \item \emph{Selection and state guarantees.}
We connect selection bounds to configuration families and bound expected excess loss for a selector that limits switches under exogenous losses, bounded influence, and full counterfactual feedback; see Sections~\ref{sec:selection}--\ref{sec:state}.
    \item \emph{Weaker feedback.}
We quantify the costs and additional assumptions of approximate evaluation, sparse probes, and deployment-only feedback; see Section~\ref{sec:feedback}.
    \item \emph{Empirical comparisons.}
Controlled studies and SQL evaluations show how candidate availability and inherited state change the interpretation of selection regret and switching; see Section~\ref{sec:experiments}.
\end{itemize}

\section{Configurations, Comparators, and Loss Decomposition}
\label{sec:protocol}

\paragraph{Configurations and Candidate Families.}
A protocol $\Xi$ declares the executor, available data, resource budgets, and permitted updates.
An \emph{admissible} configuration is one allowed and executable under these rules.
Write an admissible configuration as $a=(z,w)\in\mathcal A\subseteq\mathcal Z\times\mathcal W$, where $z$ is reusable text and $w$ specifies deployable parameters.
Learned continuous prompt tensors also belong to $w$.
Relative to a base pair $(z_0,w_0)$, the text-only family $\mathcal A_Z=\mathcal A\cap(\mathcal Z\times\{w_0\})$ holds parameters fixed.
The parameter-only family $\mathcal A_W=\mathcal A\cap(\{z_0\}\times\mathcal W)$ holds text fixed.
Their union is $\mathcal A_\cup=\mathcal A_Z\cup\mathcal A_W$, while the joint family is $\mathcal A_{ZW}=\mathcal A$.
We select from finite candidates $\mathcal C_{ZW}\subseteq\mathcal A$ constructed before evaluation and containing the base pair, with $\mathcal C_e=\mathcal C_{ZW}\cap\mathcal A_e$ for $e\in\{Z,W,\cup,ZW\}$.

\paragraph{One Deployment Round.}
At round $t\in\{1,2,\ldots,T\}$, the selector chooses $A_t\in\mathcal C_e$ using the feedback available from earlier rounds.
The executor applies $A_t$ to task input $x_t$ in operational state $s_t$, producing outputs and tool actions, incurring a normalized loss $\ell_t\in[0,1]$, and advancing the state to $s_{t+1}$.
Loss can indicate task failure, while state can retain generated code or dialogue history.
The permitted feedback updates future selection probabilities, while execution changes the operational state.
Incurring a loss does not imply observing every candidate's loss.
The formal executor and state transition are provided in Appendix~\ref{sec:app-theory-accounting}.

\paragraph{Fixed-Configuration Histories and Feedback.}
Fix a common initial condition and an exogenous random tape, independent of selector randomness, determining losses $F_t(a_{1:t})\in[0,1]$ for histories $a_{1:t}=(a_1,\ldots,a_t)$.
For the deployed history, $\ell_t=F_t(A_{1:t})$.
Each alternative history reruns the state and any deployment-responsive tasks.
The \emph{diagonal loss} of a fixed configuration $a$ is
\begin{equation}
    d_t(a)=F_t(a,\ldots,a).
    \label{eq:common-diagonal-loss}
\end{equation}
This is its round-$t$ loss if it had been used from the beginning.
After switching from $a$ to $b$, for example, the round-two deployed loss is $F_2(a,b)$, whereas the diagonal loss of the current choice is $d_2(b)=F_2(b,b)$.
Full diagonal feedback reveals every $d_t(a)$ after round $t$,
sparse feedback probes individual candidates independently of deployment,
and deployment-only feedback reveals the realized loss $\ell_t$.
The accounting identity requires no observation of counterfactual losses, while the learning guarantees state their feedback assumptions separately.

\paragraph{Comparator and Loss Accounting.}
Let $D_T(a)=\sum_{t=1}^T d_t(a)$ be the cumulative loss of using $a$ throughout.
The best fixed value within family $e$ is $V_{T,\mathcal C}^e=\min_{a\in\mathcal C_e}D_T(a)$.
Our common comparator has value $V_{T,\mathcal C}^{ZW}$: hindsight selects one joint candidate for the entire horizon, with its own induced history.
Write $L_T^{\mathrm{dep}}=\sum_{t=1}^T F_t(A_{1:t})$ for the learner's cumulative deployed loss and $L_T^{\mathrm{diag}}=\sum_{t=1}^T d_t(A_t)$ for its choices evaluated on their respective fixed-configuration histories.
The latter sum need not describe one deployment trajectory.
Adding and subtracting it and the best within-family value gives the decomposition.

\begin{proposition}[Selection, restriction, and state accounting]
\label{prop:common-accounting}
For every trajectory with $A_t\in\mathcal C_e$,
\begin{equation}
    L_T^{\mathrm{dep}}-V_{T,\mathcal C}^{ZW}
    =\underbrace{L_T^{\mathrm{diag}}-V_{T,\mathcal C}^e}_{\text{diagonal selection}}
    +\underbrace{V_{T,\mathcal C}^e-V_{T,\mathcal C}^{ZW}}_{\text{family restriction}}
    +\underbrace{L_T^{\mathrm{dep}}-L_T^{\mathrm{diag}}}_{\text{signed state effect}}.
    \label{eq:common-accounting}
\end{equation}
Only family restriction is necessarily nonnegative; the identity also holds in expectation.
\end{proposition}
Selection compares choices within the available family,
restriction charges its gap to the joint family,
and state accounts for the actual history's benefit or harm.
Stateless systems have zero state effect.
For fixed diagonal tables, identical action marginals give equal expected diagonal loss, while different deployment histories can help or hurt.
Appendix~\ref{sec:app-accounting-conventions} details sampling residuals and comparator averaging.
The proofs appear in Appendix~\ref{sec:app-proofs}.

\section{Selection and Family Restriction}
\label{sec:selection}

\paragraph{Selection and Prior Complexity.}
\label{sec:finite-selection}

For stateless full-information selection, exponential weights with a positive prior $\mu_e$ on $\mathcal C_e$ and learning rate $\eta>0$ uses the following distribution:
\begin{equation}
    q_t(a)=\mu_e(a)e^{-\eta\sum_{r<t}d_r(a)}/\sum_{b\in\mathcal C_e}\mu_e(b)e^{-\eta\sum_{r<t}d_r(b)}.
    \label{eq:finite-ew}
\end{equation}
Sample $A_t\sim q_t$, then observe every $d_t(a)$.
The following theorem also covers standard-Borel configuration spaces with prior $\mu$ fixed before evaluation and measurable losses $L_t:\mathcal A\to[0,1]$ fixed before the current draw and fully revealed afterward.
Here $p_t(\mathrm da)\propto e^{-\eta\sum_{r<t}L_r(a)}\mu(\mathrm da)$.
Write $\overline L_t=\int L_t\,\mathrm dp_t$ and $R_T(Q)=\sum_t\int L_t\,\mathrm d(p_t-Q)$, and let $C_Q=\operatorname{KL}(Q\,\|\,\mu)$ denote the Kullback--Leibler (KL) divergence.
The comparator holds one draw from fixed $Q$ throughout, with $R_T(a)=R_T(\delta_a)$.

\begin{theorem}[Prior-complexity regret]
\label{thm:prior-complexity}
For every $\eta>0$ and fixed $Q\ll\mu$ with $C_Q<\infty$, prior-initialized exponential weights satisfies the following:
\begin{equation}
    R_T(Q) \le \operatorname{KL}(Q\,\|\,\mu)/\eta+\eta T/8.
    \label{eq:general-kl-regret}
\end{equation}
For $C_Q>0$, $\eta=\sqrt{8C_Q/T}$ gives the following:
\begin{equation}
    R_T(Q)\le\sqrt{TC_Q/2}.
    \label{eq:general-kl-regret-optimized}
\end{equation}
\end{theorem}
Setting $L_t=d_t$ and $\mu=\mu_e$, the best finite candidate $a_e^\star$ has $C_{\delta_{a_e^\star}}=\log(1/\mu_e(a_e^\star))$.
Thus,
\begin{equation}
    \mathbb E L_T^{\mathrm{dep}}-V_{T,\mathcal C}^{ZW}
    \le\log(1/\mu_e(a_e^\star))/\eta+\eta T/8+V_{T,\mathcal C}^{e}-V_{T,\mathcal C}^{ZW}.
    \label{eq:finite-common-bound}
\end{equation}
Uniform priors give a selection-regret bound of $\sqrt{T\log|\mathcal C_e|/2}$, but restriction can remain linear.

Appendix~\ref{sec:measurable-selection} separates text and parameter prior complexity, treats continuous comparators by localization, and discusses computational limits.

\paragraph{Restriction and Mode Selection.}
\label{sec:mode-restriction}

To extend the finite-candidate comparison to all admissible configurations, let $V_T^e=\inf_{a\in\mathcal A_e}\sum_tL_t(a)$.
Define $\Delta_e^{\mathrm{mode}}=V_T^e-V_T^\cup$, $\Delta^{\mathrm{joint}}=V_T^\cup-V_T^{ZW}$, and $\mathsf{Apx}_T^e(Q_e)=\sum_t\int L_t\,\mathrm dQ_e-V_T^e$.
These nonnegative gaps distinguish comparator approximation, mode restriction, and access to joint configurations.

\begin{proposition}[Selection--approximation--restriction decomposition]
\label{prop:selection-restriction}
For any learner and comparator $Q_e$ on $\mathcal A_e$, $e\in\{Z,W\}$, the regret $R_T^{ZW}=\sum_t\overline L_t-V_T^{ZW}$ satisfies the following:
\begin{equation}
    R_T^{ZW}=R_T(Q_e)+\mathsf{Apx}_T^e(Q_e)+\Delta_e^{\mathrm{mode}}+\Delta^{\mathrm{joint}}.
    \label{eq:main-selection-restriction}
\end{equation}
EW on $\mathcal A_e$ with prior $\mu_e$ bounds $R_T(Q_e)$ by $\operatorname{KL}(Q_e\,\|\,\mu_e)/\eta+\eta T/8$ whenever this KL is finite.
\end{proposition}

A mixture prior also permits online mode selection, charging the comparator mode an additional prior cost; see Corollary~\ref{cor:mixture-mode-selection}.
Candidate construction remains separate: Appendix~\ref{sec:candidate-discovery} distinguishes localization from discovery without claiming convergence of prompt search or adapter training.

\paragraph{Joint and Coordinatewise Selection.}
\label{sec:coordinate-selection}

Fix finite Cartesian candidates $\mathcal C=\mathcal Z_c\times\mathcal W_c$, a positive product prior, and exogenous stateless losses $d_t:\mathcal C\to[0,1]$.
Both selectors observe every loss and start from $A_1\sim\mu$.
For deterministic $\eta_t>0$, let $q_t(z,w)\propto\pi(z)\rho(w)e^{-\eta_t\sum_{r<t}d_r(z,w)}$.
Joint sampling draws from $q_t$,
while one coordinate sweep, for $t\ge2$, uses fresh draws:
\begin{equation}
    Z_t\sim q_t(\cdot\mid W_{t-1}),\qquad
    W_t\sim q_t(\cdot\mid Z_t).
\end{equation}
Only the final pair is deployed.
Write $R_T^{\mathrm{joint}},R_T^{\mathrm{coord}}$ for expected regret against the same best fixed candidate.
In this comparison, restriction and state effects vanish.

\begin{proposition}[Equivalence under additive losses]
\label{prop:coordinate-additive-equivalence}
If $d_t(z,w)=g_t(z)+h_t(w)$, one sweep has law $q_t$ for every deterministic schedule, so $R_T^{\mathrm{coord}}=R_T^{\mathrm{joint}}$.
\end{proposition}

\begin{proposition}[Separation under a fixed learning rate]
\label{prop:coordinate-fixed-rate-separation}
On $\mathcal C=\{0,1\}^2$ with uniform prior, let $d_t=f$, where $f(0,0)=\Delta$, $f(1,0)=f(0,1)=\Delta+b$, $f(1,1)=0$, and $\Delta,b>0$, $\Delta+b\le1$.
For fixed $\eta_t=\eta>0$ and every $T\ge1$,
\begin{equation}
    R_T^{\mathrm{joint}}\le1+3/\eta,\qquad
    R_T^{\mathrm{coord}}\ge(\Delta T/4)\exp\!\left(-2/(e^{\eta b}-1)\right).
\end{equation}
\end{proposition}

\begin{corollary}[Separation despite eventual optimal selection]
\label{cor:coordinate-convergent-separation}
For the same construction with $\Delta>b$, use $\eta_t=\log(4(t-1))/(b(t-1))$ for $t\ge2$.
Then $R_T^{\mathrm{joint}}=O(1)$ and $R_T^{\mathrm{coord}}=\Omega(\sqrt T)$, although both selectors deploy $(1,1)$ on every sufficiently late round almost surely.
\end{corollary}
These results concern the specified conditional sampler.
More sweeps or global refreshes can change its behavior.
Appendix~\ref{sec:app-coordinate-selection} provides refined constants,
and Section~\ref{sec:experiments} examines these settings.

\section{State Effects and Policy Regret}
\label{sec:state}

Policy regret groups selection and state in~\eqref{eq:common-accounting}: $R_T^{\mathrm{pol}}(Q)=\mathbb E L_T^{\mathrm{dep}}-\mathbb E\int D_T(a)\,\mathrm dQ(a)$, where the comparator draws once from $Q$ and deploys it from the common initial condition~\citep{arora2012policy}.
Point regret takes $Q=\delta_a$.
Evaluating comparators on the learner's realized state instead gives external regret.

\begin{proposition}[Zero external regret and linear policy regret]
\label{prop:external-policy-separation}
A deterministic two-configuration persistent-state system admits a trajectory with zero realized-state external regret and policy regret $T-1$ against a fixed configuration.
\end{proposition}

\begin{proposition}[Linear policy regret]
\label{prop:irreversible-state}
For every optimizer whose first action uses only common pre-round history and private coins, and every $T\geq2$, some deterministic two-configuration environment gives expected policy regret at least $(T-1)/2$, when its index and transition kernel are unknown before that action.
\end{proposition}

\begin{assumption}[Component-specific bounded influence]
\label{ass:bounded-influence}
For integers $m_Z,m_W\geq0$, measurable functions $f_t$, fixed before learner randomization, give loss $f_t(z_{\max\{1,t-m_Z\}:t};w_{\max\{1,t-m_W\}:t})\in[0,1]$.
Older coordinates have no influence, including through hidden state, while interactions within the windows are unrestricted.
\end{assumption}

Put $m=m_{\max}=\max\{m_Z,m_W\}$, truncating horizons at $T-1$.
The diagonal $d_t(z,w)$ repeats each coordinate throughout its window.
Influence measures persistence after replacement.

\begin{assumption}[Diagonal counterfactual evaluation oracle]
\label{ass:diagonal-oracle}
After round $t$, every own-trajectory loss $d_t(a)$, $a\in\mathcal A$, is revealed.
\end{assumption}

\paragraph{Lazy Exponential Weights.}
Apply the EW update from Section~\ref{sec:selection} to $d_t$, starting with $q_1=\mu$ and $A_1\sim\mu$.
After round $t<T$, retain $A_t$ with probability $e^{-\eta d_t(A_t)}$ and otherwise refresh from $q_{t+1}$~\citep{geulen2010regret,kalai2005efficient}.
Refreshes preserve $A_t\sim q_t$ and occur at most $\eta(T-1)$ times in expectation.
Each refresh contaminates at most $m$ rounds.
This requires own-trajectory evaluation.

\begin{proposition}[Matched marginals isolate state effects]
\label{prop:matched-marginals}
For fixed exogenous losses, family, prior, and learning rate, independent and lazy draws with the same diagonal feedback and EW marginals have equal expected diagonal losses.
Hence,
\begin{equation}
    \mathbb E L_T^{\mathrm{dep,lazy}}-\mathbb E L_T^{\mathrm{dep,ind}}
    =\mathbb E[L_T^{\mathrm{dep,lazy}}-L_T^{\mathrm{diag,lazy}}]
    -\mathbb E[L_T^{\mathrm{dep,ind}}-L_T^{\mathrm{diag,ind}}].
    \label{eq:matched-marginals-state}
\end{equation}
\end{proposition}

Thus fewer switches can help or hurt.
The identity also holds conditional on shared exogenous approximate feedback.

\begin{theorem}[Lazy policy regret under bounded influence]
\label{thm:lazy-policy}
Under Assumptions~\ref{ass:bounded-influence}--\ref{ass:diagonal-oracle}, for every $\eta>0$ and $Q\ll\mu$ with finite $C_Q=\operatorname{KL}(Q\,\|\,\mu)$,
\begin{equation}
    R_T^{\mathrm{pol}}(Q)\leq C_Q/\eta
        +\eta\left[T/8+m_{\max}(T-1)\right].
    \label{eq:lazy-policy-regret}
\end{equation}
For $C_Q>0$, comparator-dependent tuning gives the following:
\begin{align}
    \eta^\star&=\sqrt{C_Q/(T/8+m_{\max}(T-1))},
    \label{eq:lazy-learning-rate}\\
    R_T^{\mathrm{pol}}(Q)&\!\leq\!2\sqrt{C_Q[T/8\!+\!m_{\max}(T\!-\!1)]}\!\leq\!2\sqrt{TC_Q(m_{\max}\!+\!1/8)}
    \!=\!O\!\left(\!\!\sqrt{(1\!+\!m_{\max})TC_Q}\right).
    \label{eq:lazy-policy-regret-optimized}
\end{align}
For $C_Q=0$, draw once from $\mu$ without refreshing; regret is zero.
\end{theorem}

The $m\eta(T-1)$ term controls state, while the remaining terms control selection.
These established low-switch methods~\citep{geulen2010regret,anava2015memory,zhang2022switching} connect prior complexity to influence.
Uniform priors give $C_Q=\log K$ for every point comparator.

\begin{proposition}[Mode-restricted lazy influence rates]
\label{prop:mode-lazy-rates}
Under Assumptions~\ref{ass:bounded-influence}--\ref{ass:diagonal-oracle}, run lazy EW with priors $\mu_Z=\pi\otimes\delta_{w_0}$ and $\mu_W=\delta_{z_0}\otimes\rho$, supported on $\mathcal A_Z$ and $\mathcal A_W$, respectively.
For fixed comparators $Q_Z\ll\pi$ and $Q_W\ll\rho$, let $C_Z=\operatorname{KL}(Q_Z\,\|\,\pi)$ and $C_W=\operatorname{KL}(Q_W\,\|\,\rho)$ be finite.
For every $\eta>0$, text-only and parameter-only selection satisfy
\begin{align}
    R_T^{\mathrm{pol}}(Q_Z\otimes\delta_{w_0})
    &\le C_Z/\eta+\eta[T/8+m_Z(T-1)],
    \label{eq:text-only-lazy-policy}\\
    R_T^{\mathrm{pol}}(\delta_{z_0}\otimes Q_W)
    &\le C_W/\eta+\eta[T/8+m_W(T-1)].
    \label{eq:parameter-only-lazy-policy}
\end{align}
\end{proposition}

Freezing one coordinate removes its switching contribution: the state charge depends on the influence horizon of the coordinate being changed.
For positive complexity, comparator-tuned rates are $O(\sqrt{(1+m_Z)TC_Z})$ and $O(\sqrt{(1+m_W)TC_W})$.
For the best fixed candidate within each finite family, add family restriction to compare with the joint optimum.
Different upper bounds alone do not establish which mode performs better.
Appendix~\ref{sec:bounded-influence} gives localization and coupling details, while Appendix~\ref{sec:episodic-resets} treats whole-policy selection across reset episodes.
Proofs are in Appendix~\ref{sec:app-proofs-state}.

\section{Feedback Requirements and Extensions}
\label{sec:feedback}

Counterfactual feedback evaluates each candidate on the history it would have produced if used throughout.
Deployment feedback reports only the loss actually incurred.
We now quantify what changes as feedback becomes less informative.
Fix a finite family $\mathcal C$ of $K$ candidates with positive prior $\mu$, and a comparator distribution $Q$ fixed independently of learner and probe randomness.
Write $m=m_{\max}$ and $C_Q=\operatorname{KL}(Q\,\|\,\mu)$ as in Section~\ref{sec:state}.

\paragraph{Approximate Full Counterfactual Feedback.}
Theorem~\ref{thm:lazy-policy} assumes exact diagonal losses.
An exogenous full estimate sequence in $[0,1]$ with uniform round-$t$ error $\epsilon_t$ adds at most $2\sum_t\epsilon_t$ to its bound.
Conditionally unbiased estimates in the same range preserve the expected bound;
refer to Appendix~\ref{sec:full-feedback-accuracy}.
The entire evaluator must be independent of deployment randomness.
Even a common estimation offset can alter retention and state effects while leaving EW marginals unchanged.

\paragraph{Sparse Independent Probes.}
Under the bounded-influence condition in Assumption~\ref{ass:bounded-influence}, an autonomous evaluator can instead probe one candidate $J_t\sim q_t$ after each round.
Its outcome $Z_t\in[0,1]$ must have conditional mean $d_t(J_t)$ given the exogenous response tree, earlier evaluation history, and $J_t$.
Conditional on that tree, the entire evaluator is independent of deployment randomness, consumes no deployed actions or outcomes, and is unchanged by deployment interventions.
Probing also leaves the tree unchanged;
see Assumption~\ref{ass:independent-diagonal-probe}.
Initialize $q_1=\mu$ and $A_1\sim\mu$, and update EW using $\widehat d_t(a)=\mathbf 1\{J_t=a\}Z_t/q_t(a)$.
Retain $A_t$ with probability $e^{-\eta\widehat d_t(A_t)}$ and otherwise refresh from $q_{t+1}$ using independent deployment randomness.

\begin{theorem}[Sparse-diagonal policy regret]
\label{thm:sparse-diagonal}
Under Assumption~\ref{ass:independent-diagonal-probe}, with probe law $r_t=q_t$, every fixed $Q$ and $\eta>0$ satisfy the following:
\begin{equation}
    R_T^{\mathrm{pol}}(Q)\leq C_Q/\eta
        +\eta\left[KT/2+m(T-1)\right].
    \label{eq:sparse-diagonal-uniform}
\end{equation}
For $C_Q>0$, $\eta=\sqrt{C_Q/[KT/2+m(T-1)]}$ gives
\begin{equation}
    R_T^{\mathrm{pol}}(Q)\leq2\sqrt{C_Q[KT/2+m(T-1)]}.
    \label{eq:sparse-diagonal-optimized}
\end{equation}
For $C_Q=0$, drawing once from $\mu$ gives zero expected regret.
\end{theorem}

One probe per round reduces $KT$ counterfactual evaluations to $T$ probes, replacing the full-feedback selection term $\eta T/8$ by $\eta KT/2$ while preserving the state term.
Simulation or cloning costs remain, and independence from only the current deployment draw is insufficient.
Appendix~\ref{sec:sparse-diagonal} gives the formal process and refined bounds.

\paragraph{Deployment-Only Feedback.}
Without a counterfactual evaluator, hold each chosen configuration for a block of $b$ rounds, where $1\le b\le T$ is an integer.
There are $J=\lfloor T/b\rfloor$ complete blocks and $r=T-bJ$ remaining rounds.
Block $j$ starts at $s_j=(j-1)b+1$ after the realized prefix $H_j=A_{1:s_j-1}$.
We write $U$ explicitly for the external random tape introduced in Section~\ref{sec:protocol}.

\begin{definition}[Block counterfactual]
\label{def:block-counterfactual}
The average loss from repeating $a$ after the actual prefix, rerunning environmental responses, is
\begin{equation}
    g_j(a)=(1/b)\sum_{u=1}^bF_{s_j+u-1}(H_j,a^u;U),
    \label{eq:block-counterfactual}
\end{equation}
where $a^u$ denotes $u$ repetitions of $a$.
\end{definition}

Starting from $p_1=\mu$, deploy a fresh private draw $B_j\sim p_j$ throughout block $j$ and update EW using $\widehat g_j(a)=\mathbf 1\{B_j=a\}Z_j/p_j(a)$, where $Z_j=g_j(B_j)$ is the observed block-average loss.
Retain the last configuration for the remaining $r$ rounds.
The signed \emph{causal reaction} compares repeating a comparator after the learner's prefixes with using it from the beginning, over the complete blocks:
\begin{equation}
    \Gamma_b(Q)=\mathbb E\sum_{a\in\mathcal C}Q(a)
    \left[b\sum_{j=1}^Jg_j(a)-\sum_{t=1}^{bJ}d_t(a)\right].
    \label{eq:causal-reaction}
\end{equation}
It includes changes in tasks, user responses, and operational state.

\begin{theorem}[Oracle-free causal policy-regret inequality]
\label{thm:causal-bandit}
For causal responses $F_t\in[0,1]$ driven by an external tape independent of private learner coins, the block algorithm satisfies, for every fixed $Q$ and $\eta>0$,
\begin{equation}
    R_T^{\mathrm{pol}}(Q) \leq bC_Q/\eta
        +\eta bKJ/2+\Gamma_b(Q)+r.
    \label{eq:causal-bandit}
\end{equation}
No oracle for $g_j$, $d_t$, or $\Gamma_b$ is required.
\end{theorem}

The first two terms bound selection from observed block losses, while $\Gamma_b(Q)$ accounts for the different histories and $r$ covers the unfinished block.
This retains the causal discrepancy in policy-regret mini-batching~\citep{arora2012policy}, and uncontrolled reaction can remain linear.
If the \emph{complete} causal system has influence horizon $m$, including task selection and evaluation, then $|\Gamma_b(Q)|\le\min\{m,b\}(J-1)$: only the first $m$ rounds of each later block can remember its prefix.
Optimizing $b$ and $\eta$ yields $O((mKC_Q)^{1/3}T^{2/3})$ for fixed positive $m,K,C_Q$ and sufficiently large $T$;
see Corollary~\ref{cor:bandit-finite-influence}.
Use $b=1$ for $m=0$, and a single prior draw for $C_Q=0$.
Forgetting in a cache alone does not establish this complete-system condition.

For random response trees, independence and unbiasedness must hold conditionally to compare with the tree's best fixed candidate.
Query-adaptive autonomous evaluation supports fixed or best expected-loss comparators, not comparators selected using probe noise.
Appendix~\ref{sec:app-theory-feedback} provides supporting lemmas and extensions.

\section{Empirical Analysis}
\label{sec:experiments}

\paragraph{Experimental Setup.}
\label{sec:exp-setup}

Controlled studies isolate selection, family restriction, and signed state effects on deployed loss.
SQL experiments use Qwen3.5-9B~\citep{qwen35modelcard} on Spider~\citep{YuT2018emnlp} and SParC~\citep{YuT2019acl}.
Excess loss is measured relative to each table's or benchmark's best fixed joint configuration unless a paired contrast is specified.
Summaries and error bars use means $\pm$ one standard error (SE).
Refer to Appendices~\ref{sec:app-experiments} and~\ref{sec:app-uncertainty} for detailed designs, numerical results, and uncertainty.

\paragraph{Joint and Coordinatewise Selection.}
\label{sec:exp-coordinate-selection}

\begin{figure}[t]
    \centering
    \begin{minipage}[t]{0.48\textwidth}
        \vspace{0pt}
        \centering
        \captionsetup{font=small}
        \begin{subfigure}[t]{0.48\linewidth}
            \centering
            \includegraphics[width=\linewidth]{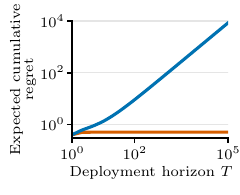}
            \caption{Fixed rate ($\eta=5$)}
            \label{fig:coordinate-selection-fixed}
        \end{subfigure}\hfill
        \begin{subfigure}[t]{0.48\linewidth}
            \centering
            \includegraphics[width=\linewidth]{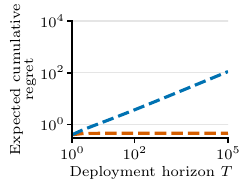}
            \caption{Slow schedule}
            \label{fig:coordinate-selection-slow}
        \end{subfigure}
        \par\smallskip
        \includegraphics[width=\linewidth,trim=40bp 0bp 124bp 0bp,clip]{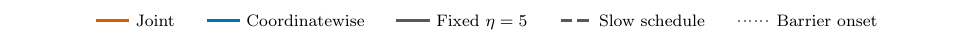}
        \vspace{-17pt}
        \caption{Exact expected regret for joint and coordinatewise selection with fixed and slow learning rates.}
        \label{fig:coordinate-selection-regret}
    \end{minipage}\hfill
    \begin{minipage}[t]{0.48\textwidth}
        \vspace{0pt}
        \centering
        \captionsetup{font=small}
        \begin{subfigure}[t]{0.48\linewidth}
            \centering
            \includegraphics[width=\linewidth]{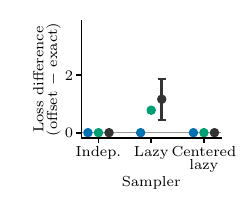}
            \caption{Common offset}
            \label{fig:exp2-offset-fixed-gap}
        \end{subfigure}\hfill
        \begin{subfigure}[t]{0.48\linewidth}
            \centering
            \includegraphics[width=\linewidth]{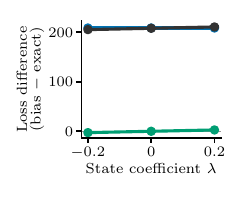}
            \caption{Ref.-opposing bias}
            \label{fig:exp2-bias-fixed-gap}
        \end{subfigure}
        \vspace{-3pt}
        \par\smallskip
        \includegraphics[width=\linewidth,trim=90bp 0bp 90bp 0bp,clip]{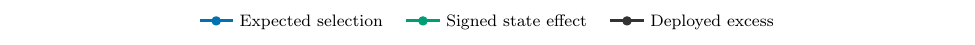}
        \vspace{-17pt}
        \caption{Approximate-minus-exact loss components on fixed-advantage tables: \subref{fig:exp2-offset-fixed-gap} $\lambda=0.2$; \subref{fig:exp2-bias-fixed-gap} lazy EW.}
        \label{fig:exp2-perturbations}
    \end{minipage}
    \vspace{-10pt}
\end{figure}

Joint EW and text-then-parameter sweeps share the uniform prior, preceding loss rows, and rate schedule on the additive and barrier tables of Section~\ref{sec:coordinate-selection}.
Family restriction, state effects, and the best fixed loss are zero, so cumulative loss equals regret.
The additive control gives equal regret, whereas one coordinatewise sweep incurs substantially more regret on the barrier under both schedules,
as in Figure~\ref{fig:coordinate-selection-regret}.
Under the slow schedule, high final-round optimal-selection probability coexists with this accumulated gap.
Extra sweeps reduce the gap at additional computation cost without new feedback;
see Figure~\ref{fig:coordinate-selection-sensitivity} and Appendix~\ref{sec:app-coordinate} for details.

\paragraph{Selection and Signed State Effects.}
\label{sec:exp-exact-persistent}

\begin{figure}[t]
    \centering
    \begin{subfigure}[t]{0.32\linewidth}
        \centering
        \includegraphics[width=\linewidth]{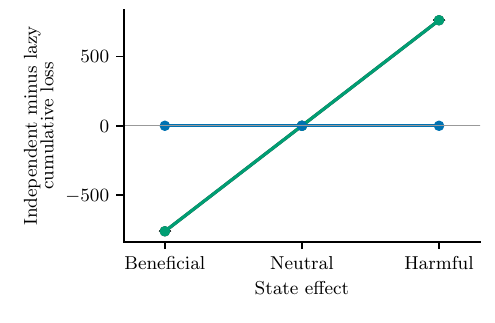}
        \caption{No selection signal}
        \label{fig:exp1-coupling-no-signal}
    \end{subfigure}
    \hfill
    \begin{subfigure}[t]{0.32\linewidth}
        \centering
        \includegraphics[width=\linewidth]{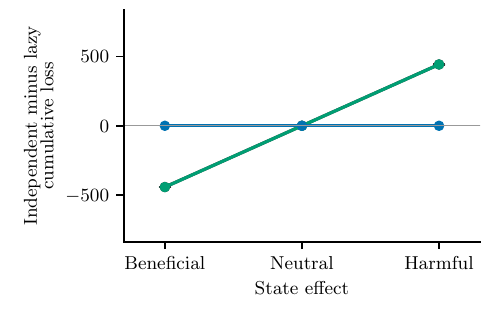}
        \caption{Fixed advantage}
        \label{fig:exp1-coupling-fixed-gap}
    \end{subfigure}
    \hfill
    \begin{subfigure}[t]{0.32\linewidth}
        \centering
        \includegraphics[width=\linewidth]{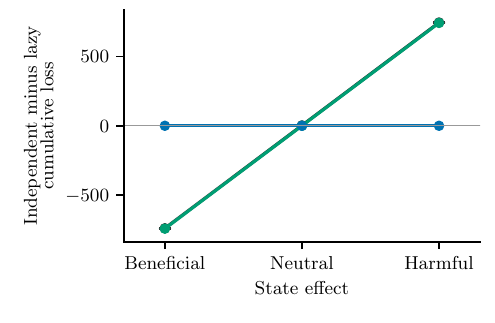}
        \caption{Changing relative quality}
        \label{fig:exp1-coupling-changing-quality}
    \end{subfigure}
    \par\smallskip
    \vspace{-5pt}
    \includegraphics[width=\linewidth]{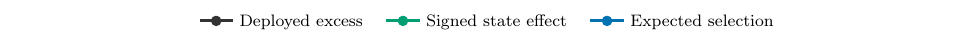}
    \vspace{-20pt}
    \caption{Independent-minus-lazy loss contrasts at $T=4{,}096$, with $\lambda\in\{-0.2, 0, 0.2\}$. Reversing the mechanism reverses the selector ordering while expected diagonal selection remains unchanged.}
    \label{fig:exp1-coupling}
    \vspace{-10pt}
\end{figure}

Fixed-advantage, changing-quality, and no-signal diagonal tables use $f_t=d_t(A_t)+\lambda I_t$,
where $I_t$ indicates mixed configurations in the relevant coordinate histories.
Negative, zero, and positive $\lambda$ give beneficial, neutral, and harmful state effects.
State and approximate-feedback figures use $(m_Z,m_W)=(4,0)$;
see Appendix~\ref{sec:app-persistent} for details.
Independent and lazy EW have identical expected diagonal losses,
as presented in Figure~\ref{fig:exp1-accounting},
but lazy EW has lower deployed loss under harmful state and higher loss under beneficial state;
see Figure~\ref{fig:exp1-coupling}.
Fewer switches can therefore help or hurt.
Without a selection signal, draw-once has zero excess while both EW selectors incur harmful state costs,
as shown in Figure~\ref{fig:exp1-horizons}.

\paragraph{Feedback Accuracy, Switching, and Evaluation Cost.}
\label{sec:exp-approximate-feedback}

We perturb diagonal feedback at $T=4{,}096$, fixing the true table, prior, rate, and learner random inputs.
A common offset preserves selection distributions but changes ordinary lazy EW's switching and state costs, while the trajectories of independent EW and centered lazy EW remain unchanged.
A reference-opposing bias changes both selection and state,
as shown in Figure~\ref{fig:exp2-perturbations};
refer to Appendix~\ref{sec:app-approximate} for experimental details and additional results.
Increasing the Bernoulli sampling budget reduces feedback error without monotone deployed-loss improvement in the displayed conditions;
see Figure~\ref{fig:exp2-budget}.

\paragraph{Restricted Feedback on Controlled Losses.}
\label{sec:exp-restricted-feedback}

\begin{figure}[t]
    \centering
    \begin{minipage}[t]{0.48\textwidth}
        \vspace{0pt}
        \centering
        \captionsetup{font=small}
        \begin{subfigure}[t]{0.48\linewidth}
            \centering
            \includegraphics[width=\linewidth]{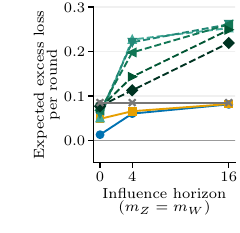}
            \caption{$\lambda=0.2$}
            \label{fig:feedback-controlled-positive}
        \end{subfigure}\hfill
        \begin{subfigure}[t]{0.48\linewidth}
            \centering
            \includegraphics[width=\linewidth]{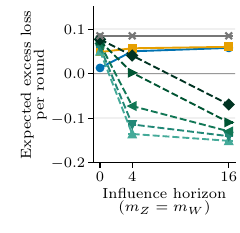}
            \caption{$\lambda=-0.2$}
            \label{fig:feedback-controlled-negative}
        \end{subfigure}
        \vspace{-2pt}
        \includegraphics[width=\linewidth]{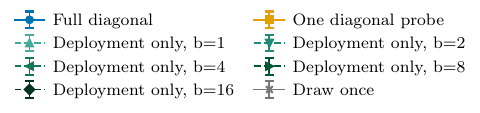}
        \vspace{-16pt}
        \caption{Restricted feedback on fixed-gap losses at $T=1{,}024$.
        Zero marks the best fixed comparator.}
        \label{fig:feedback-controlled}
    \end{minipage}\hfill
    \begin{minipage}[t]{0.48\textwidth}
        \vspace{0pt}
        \centering
        \captionsetup{font=small}
        \begin{subfigure}[t]{0.49\linewidth}
            \centering
            \includegraphics[width=\linewidth]{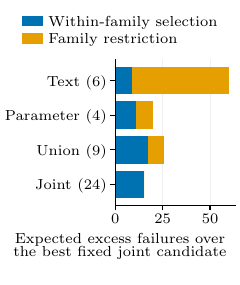}
            \vspace{-20pt}
            \caption{Full candidate families}
            \label{fig:sql-spider-accounting}
        \end{subfigure}\hfill
        \begin{subfigure}[t]{0.49\linewidth}
            \centering
            \includegraphics[width=\linewidth]{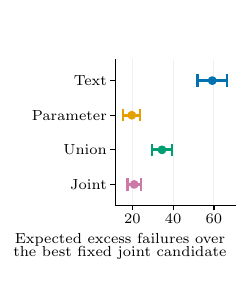}
            \vspace{-20pt}
            \caption{Equal count ($K=4$)}
            \label{fig:sql-spider-matched}
        \end{subfigure}
        \vspace{-5pt}
        \caption{Spider full-diagonal EW excess failures relative to each seed's joint optimum.
        Panel~\subref{fig:sql-spider-accounting} shows selection regret and family restriction, with family sizes in parentheses.
        Panel~\subref{fig:sql-spider-matched} shows all base-anchored $K=4$ subsets, averaged within each seed.}
        \label{fig:sql-spider-comparison}
    \end{minipage}
    \vspace{-10pt}
\end{figure}

On the four-candidate joint family, we compare full-diagonal and one-probe lazy EW, deployment-only EW with blocks $b\in\{1,2,4,8,16\}$, and draw-once.
All use uniform priors and regime-specific rates, so both information and learning rate vary;
refer to Appendix~\ref{sec:app-controlled-feedback}.
At $T=1{,}024$ and $m_Z=m_W=4$, increasing $b$ from 1 to 16 lowers excess under harmful state but raises it under beneficial state,
as shown in Figure~\ref{fig:feedback-controlled}.
These results illustrate how blocks sacrifice updates and suppress both kinds of state effects.
Negative excess means a changing history beats every fixed configuration under this synthetic law.

\paragraph{Selection and Family Restriction on Spider.}
\label{sec:exp-stateless-applied}

\begin{figure}[t]
    \centering
    \begin{subfigure}[t]{0.24\linewidth}
        \centering
        \includegraphics[width=\linewidth]{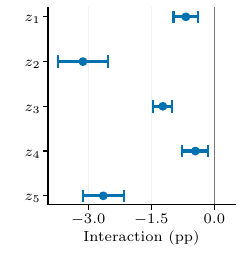}
        \caption{Spider interactions}
        \label{fig:sql-spider-interactions}
    \end{subfigure}\hfill
    \begin{subfigure}[t]{0.24\linewidth}
        \centering
        \includegraphics[width=\linewidth]{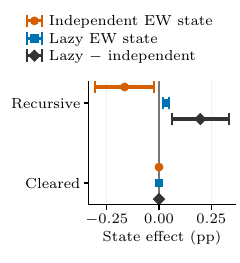}
        \caption{SParC state}
        \label{fig:sql-sparc-state}
    \end{subfigure}\hfill
    \begin{subfigure}[t]{0.24\linewidth}
        \centering
        \includegraphics[width=\linewidth]{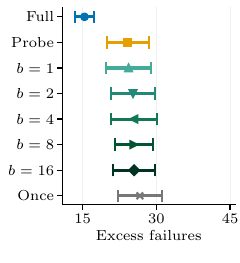}
        \caption{Spider feedback}
        \label{fig:sql-feedback-spider}
    \end{subfigure}\hfill
    \begin{subfigure}[t]{0.24\linewidth}
        \centering
        \includegraphics[width=\linewidth]{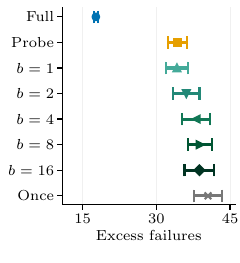}
        \caption{SParC feedback}
        \label{fig:sql-feedback-sparc}
    \end{subfigure}
    \caption{Diagnostics with prompt--adapter interactions at adapter learning rate $10^{-4}$ and SParC state effects, followed by restricted-feedback excess failures.
    Negative interactions indicate improvement beyond additive effects.
    Full and Probe denote full-diagonal and one-probe lazy EW, $b$ the deployment block length, and Once draw-once.}
    \label{fig:sql-diagnostics}
    \vspace{-10pt}
\end{figure}

Spider compares text, parameter, union, and joint families with 6, 4, 9, and 24 configurations against each seed's best fixed joint candidate.
Full-diagonal EW uses a fixed question order and a uniform prior.
Independent and lazy expectations coincide because state is absent;
refer to Appendices~\ref{sec:app-sql-protocol} and~\ref{sec:app-spider} for details.
Text EW has lower selection regret than joint EW, but its larger family restriction produces higher expected deployed loss,
as shown in Figure~\ref{fig:sql-spider-comparison}\subref{fig:sql-spider-accounting}.
Joint EW also has lower mean loss than parameter EW.
Averaging all base-anchored $K=4$ subsets gives similar parameter and joint excess losses,
as presented in Figure~\ref{fig:sql-spider-comparison}\subref{fig:sql-spider-matched}.
The full-pool ordering therefore need not survive equal candidate counts.
Demonstrations also show stronger mean improvement beyond additive prompt--adapter effects than verification,
as presented in Figure~\ref{fig:sql-diagnostics}\subref{fig:sql-spider-interactions};
refer to Appendix~\ref{sec:app-spider} for details.

\paragraph{State Effects on SParC.}
\label{sec:exp-stateful-applied}

SParC isolates history effects under matched configuration marginals.
Recursive history retains earlier generated SQL, while cleared history removes it, retaining the schema and raw utterances.
Dialogue resets bound influence by four rounds.
Both couplings share diagonal losses and an influence-dependent rate, which is retained in the cleared control.
Mixed-history outcomes measure state effects without updating selection;
see Appendix~\ref{sec:app-sparc}.
Independent EW has lower estimated recursive-history loss despite identical expected diagonal loss,
as shown in Figure~\ref{fig:sql-diagnostics}\subref{fig:sql-sparc-state}.
Lazy EW substantially reduces switching with only a small change in estimated expected loss;
see Table~\ref{tab:sql-sparc-deployment}.
Fewer mixed histories remove both beneficial and harmful effects.
These data do not establish that fewer switches worsen performance.
Clearing eliminates state effects and equalizes expected deployed losses but changes the diagonal table and comparator, so cross-condition totals cannot identify state effects.

\paragraph{Restricted Feedback on Precomputed SQL Losses.}
\label{sec:exp-sql-feedback}

We apply the controlled feedback regimes to precomputed Spider and cleared-SParC loss matrices, keeping outcomes, order, prior, and joint comparator fixed.
We use zero-influence learning rates;
see Appendix~\ref{sec:app-sql-feedback}.
Full feedback has the lowest mean excess on both tasks, whereas one-probe feedback queries fewer diagonal entries but incurs more excess,
as presented in Figures~\ref{fig:sql-diagnostics}\subref{fig:sql-feedback-spider} and~\ref{fig:sql-diagnostics}\subref{fig:sql-feedback-sparc}.
One-probe and deployment-only $b=1$ means are close;
refer to Appendix~\ref{sec:app-sql-feedback}.
The longest blocks reduce switches but raise mean excess relative to $b=1$ on both tasks; see Table~\ref{tab:sql-feedback}.

\section{Related Work}

\paragraph{Parameter Adaptation and Model Editing.}
Weight adaptation includes full tuning, adapters, low-rank updates~\citep{houlsby2019adapters,hu2022lora}, continuous prompt/prefix tuning~\citep{li2021prefix,lester2021prompt}, and targeted editing~\citep{meng2022rome}.
Updates can cause interference and forgetting~\citep{kirkpatrick2017overcoming}, with low-rank and full tuning differing even at similar performance~\citep{ShuttleworthR2025neurips}.
Comparisons with in-context learning report task-dependent accuracy, generalization, and cost tradeoffs~\citep{liu2022fewshot,mosbach2023fewshot,GhoshB2026acl,RazumovskaiaE2025tacl}, while theory characterizes when prompting can match weight tuning~\citep{genewein2025prompt}.
Joint construction includes BetterTogether~\citep{SoyluD2024emnlp}, MetaTuner~\citep{BoX2026iclr}, fast--slow updates~\citep{tiwari2026fastslow}, and WHALE's alternation of online rejection-sampling tuning with Meta-Harness search~\citep{KimH2026whale}.
Our conditional sampler studies selection within a fixed family, with candidate construction.

\paragraph{Expert Advice, Mixture Priors, and Limited Switching.}
Hedge and external regret compare fixed experts~\citep{hannan1957bayes,foster1999regret,littlestone1994weighted,freund1997decision,cesabianchi2006prediction}, and KL change-of-measure extends the analysis to measurable priors.
Bandit, tracking, and sleeping/growing-expert models change feedback, comparators, or availability~\citep{auer2002bandit,bubeck2012regret,herbster1998tracking,cortes2019sleeping,mourtada2017growing}.
Lazy and limited-switch methods control regret and action changes~\citep{kalai2005efficient,geulen2010regret,altschuler2018limited}.
Switching costs admit $\sqrt T$ full-information and $\widetilde\Theta(T^{2/3})$ adversarial-bandit rates; bounded-memory adversaries can force $T^{2/3}$ under ordinary full information~\citep{cesabianchi2013switching,dekel2014bandits}.
\citet{zhang2022switching} combine prior divergence and movement penalties in comparator-adaptive finite-expert bounds.
Under exogenous losses, coupling converts marginal movement into expected switches.

Appendix~\ref{sec:app-related-work} provides additional related work.

\section{Conclusion}

We decomposed deployed loss over the best fixed joint configuration into selection, family restriction, and signed state effects.
Low regret within a candidate family, eventual optimal selection, and infrequent switching can each occur alongside higher cumulative loss.
Under the stated influence and feedback assumptions, EW with infrequent switching bounds selection regret and state effects, while weaker feedback introduces additional costs and requires further assumptions.
Our controlled and SQL experiments illustrate how these distinctions arise within the evaluated candidate families.

\subsection*{AI use statement}

We used GPT-5.6 Sol~\citep{OpenAI2026gpt56} and GPT-6 Astra~\citep{OpenAI2026gpt6astra} as our primary generative AI models.
We used these models to refine research ideas, derive mathematical proofs, implement experiments, and draft and polish the manuscript.
The human authors take responsibility for the final content of this work, including all claims, proofs, experimental results, code, and text produced with AI assistance.

\subsection*{Ethics statement}

We do not identify direct ethical concerns arising from the theoretical analysis and benchmark experiments presented in this work.
However, methods for adapting language-model systems could be used in applications with harmful consequences.
We acknowledge this potential and emphasize the importance of responsible use and appropriate safeguards.

\subsubsection*{Acknowledgments}

This work used the resources provided by the University of Pittsburgh Center for Research Computing and Data, which are supported by National Institutes of Health (NIH) award S10OD028483 and National Science Foundation (NSF) award OAC-2117681.
In addition, it used resources through CIS251382 from the Advanced Cyberinfrastructure Coordination Ecosystem: Services \& Support (ACCESS) program, supported by NSF awards OAC-2138259, OAC-2138286, OAC-2138307, OAC-2137603, and OAC-2138296.
Jungtaek Kim acknowledges support from the start-up grant of the University of Arkansas and Emmanouil-Vasileios Vlatakis-Gkaragkounis acknowledges support from the start-up grant of the University of Wisconsin--Madison.

\bibliography{references}
\bibliographystyle{iclr2027_conference}

\clearpage
\appendix
\input{appendices/appendix_theory}
\clearpage
\input{appendices/appendix_proofs}
\clearpage
\input{appendices/appendix_experiments}
\clearpage
\input{appendices/appendix_costs}
\clearpage
\input{appendices/appendix_related_work}
\input{appendices/appendix_limitations}

\end{document}

%% file: figures/adaptation_families.tex
\begin{tikzpicture}[
    x=0.01\linewidth,
    y=0.012\linewidth,
    text=black,
    font=\fontsize{6.5}{7.5}\selectfont,
    every node/.style={align=center, inner sep=0pt},
    family title/.style={font=\fontsize{7}{8}\selectfont\bfseries},
    family math/.style={font=\fontsize{8}{9}\selectfont},
    family grid/.style={
        row sep=2.5pt,
        column sep={0.44\linewidth,between origins},
        nodes={text width=0.30\linewidth, anchor=base}
    }
]
    \useasboundingbox (0,0) rectangle (100,55);
    \draw[rounded corners=5pt, draw=black!65, fill=white,
          line width=0.65pt] (0.3,0.3) rectangle (99.7,54.7);

    \node[family math] at (50,52)
        {\textbf{Joint family}\quad
         $\mathcal A_{ZW}=\mathcal A\subseteq\mathcal Z\times\mathcal W$};
    \node at (50,47) {All admissible text--parameter pairs.};

    \fill[black!6] (33,27) ellipse [x radius=28, y radius=16.5];
    \fill[black!6] (67,27) ellipse [x radius=28, y radius=16.5];
    \begin{scope}
        \clip (33,27) ellipse [x radius=28, y radius=16.5];
        \fill[black!12] (67,27)
            ellipse [x radius=28, y radius=16.5];
    \end{scope}
    \draw[black!65, line width=0.55pt] (33,27)
        ellipse [x radius=28, y radius=16.5];
    \draw[black!65, line width=0.55pt] (67,27)
        ellipse [x radius=28, y radius=16.5];

    \matrix[family grid] at (50,27) {
        \node[family title] {Text-only\\family}; &
        \node[family title] {Parameter-only\\family}; \\
        \node[family math] {$\mathcal A_Z$}; &
        \node[family math] {$\mathcal A_W$}; \\
        \node[family math] {$(z,w_0)$}; &
        \node[family math] {$(z_0,w)$}; \\
        \node {Text varies;\\parameters fixed.}; &
        \node {Parameters vary;\\text fixed.}; \\
    };

    \node[family title] at (50,30.5) {Base};
    \fill[black] (50,27) circle [radius=1.6pt];
    \node[font=\fontsize{7}{8}\selectfont] at (50,23.5) {$(z_0,w_0)$};

    \node[family title] at (50,7.2)
        {$\bigstar$\enspace Both coordinates changed};
    \node[font=\fontsize{7}{8}\selectfont] at (50,3.4)
        {$(z,w),\quad z\ne z_0\ \text{and}\ w\ne w_0$};
\end{tikzpicture}

%% file: appendices/appendix_theory.tex
\section{Supplementary Theory}
\label{sec:app-theory}

This appendix supplies the detailed setup, exact constants, supporting statements, and extensions for Sections~\ref{sec:protocol}--\ref{sec:feedback}.
Proofs of the main and supplementary results appear in Appendix~\ref{sec:app-proofs}.

\subsection{Configurations, Comparators, and Loss Decomposition}
\label{sec:app-theory-accounting}

This section describes the configuration spaces, admissible families, and execution traces used to define the loss decomposition in~\eqref{eq:common-accounting}.

\subsubsection{Admissible Configurations and Execution}

Let $\theta_0$ denote the base model.
The admissible configuration families depend on how updates are produced and deployed.

\begin{definition}[Adaptation protocol and admissible families]
\label{def:protocol}
An \emph{adaptation protocol} $\Xi$ declares the base model, executor and
interfaces, available data, resource budgets, and permitted text- and
parameter-update procedures.  It induces standard Borel spaces
$(\mathcal{Z}_{\Xi},\mathcal{F}_{\mathcal Z})$ and
$(\mathcal{W}_{\Xi},\mathcal{F}_{\mathcal W})$ of admissible text and
parameter configurations, together with a measurable compatibility family
$\mathcal A_{\Xi}\subseteq\mathcal Z_{\Xi}\times\mathcal W_{\Xi}$ of
deployed pairs containing the base pair.  Cross-mode simulation is admissible
only if $\Xi$ permits and charges its interfaces, storage, and computation.
\end{definition}

Fix $\Xi$ and suppress its subscript.
We specify evaluator and feedback assumptions separately.
Standard-Borel spaces accommodate finite, countable, and usual continuous configurations.
The \emph{text configuration} $z$, with base object $z_0$, comprises reusable objects with canonical token serializations: prompts, retained memory, demonstrations, retrieval contents, and harness instructions.
Literal serializations are countable, while continuous relaxations require a $\Xi$-declared quantizer or serializer for deployment evaluation.

A parameter configuration $w\in\mathcal W$ specifies model parameters $\theta(w)$, including learned auxiliary tensors, with $\theta(w_0)=\theta_0$.
It may be continuous or atomic, including checkpoints, adapters, patches, and learned prompt/prefix tensors.
In particular, dense tensors belong to $w$.
The measurable executor $\operatorname{exec}$ combines $(z,\theta(w))$, task input, and possibly persistent operational state $s$: environment state, tool sessions, caches, and runtime objects not selected as reusable configuration.

For a current configuration $a=(z,w)$, define the admissible one-coordinate update sets:
\begin{align}
    \mathcal{U}_Z(a)&=\mathcal A\cap(\mathcal{Z}\times\{w\}),\label{eq:text-update-set}\\
    \mathcal{U}_W(a)&=\mathcal A\cap(\{z\}\times\mathcal{W}).\label{eq:parameter-update-set}
\end{align}
These updates hold the other coordinate at its current value, whereas joint updates select from $\mathcal A$.
The comparison families in Section~\ref{sec:protocol} instead fix coordinates at the base pair $a_0=(z_0,w_0)$.
For non-Cartesian compatibility families, priors must be supported on $\mathcal A$.
The prior-based results remain valid, but product KL identities need not hold.
Joint configurations may change both base coordinates regardless of installation order.

Given task input $x_t$ at round $t$, execution produces a trace, loss, and next state:
\begin{align}
    \tau_t &\sim
    \operatorname{exec}(M_{\theta(w_t)},z_t,s_t,x_t),\label{eq:execution-trace}\\
    \ell_t &= \ell(x_t,\tau_t)\in[0,1],\label{eq:execution-loss}\\
    s_{t+1} &\sim
    P_t(\cdot\mid s_t,z_t,w_t,x_t,\tau_t).\label{eq:state-transition}
\end{align}
Here $\tau_t$ records outputs and intermediate actions, and $P_t$ governs state transitions.
The normalized loss may combine quality and per-use resources.
Appendix~\ref{sec:deployment} treats installation and path-dependent transition costs separately.

\subsubsection{Fixed Comparators and Loss Decomposition}
\label{sec:app-accounting-conventions}

We use the candidate families, causal losses, and cumulative comparator values defined in Section~\ref{sec:protocol}.
Exogenous tasks are shared across counterfactual histories, while deployment-responsive tasks, users, and environments are rerun under each history.
The accounting identity~\eqref{eq:common-accounting} conditions on this response tree.
We refine its restriction and averaging conventions below.
For text-only or parameter-only selection,
restriction can be further split at the union family into
$V_{T,\mathcal C}^{e}-V_{T,\mathcal C}^{\cup}$ and
$V_{T,\mathcal C}^{\cup}-V_{T,\mathcal C}^{ZW}$.
Neither difference alone measures a nonadditive text--parameter interaction.

For fixed diagonal tables and action marginals $q_t$, $\mathbb E L_T^{\mathrm{diag}}=\sum_t\int d_t\,\mathrm dq_t$.
Substituting this expectation into a realized trajectory's decomposition requires the \emph{sampling residual}:
\begin{equation}
    L_T^{\mathrm{diag}}-\mathbb E L_T^{\mathrm{diag}}.
    \label{eq:diagonal-sampling-residual}
\end{equation}
Experiments retain this residual and distinguish averaged hindsight optima from expected-loss optima: generally, $\mathbb E\min_a D_T(a)\neq\min_a\mathbb E D_T(a)$.

\subsubsection{Feasibility and Configuration Reachability}

Feasibility asks whether a requested operational effect is attainable, separately from its loss.

\begin{definition}[Protocol-relative update reachability]
\label{def:reachability}
Fix an operating condition $c=(t,s,x)$.
For configuration $b=(z',w')$, let $\Gamma_c(b)$ be the law of the execution trace and next state $(\tau,s^+)$ obtained by executing $M_{\theta(w')}$ with $z'$ from $(s,x)$ and applying $P_t$.
From current configuration $a$, the reachable sets are
\begin{align}
    \mathfrak{R}_Z(c;a)
    &=
    \{\Gamma_c(b):b\in\mathcal{U}_Z(a)\},\\
    \mathfrak{R}_W(c;a)
    &=
    \{\Gamma_c(b):b\in\mathcal{U}_W(a)\}.
\end{align}
A target effect is feasible for a mode when its law belongs to that mode's
reachable set, while a task-specific discrepancy defines approximate feasibility.
\end{definition}

Two updates have the same effect at $c$ when their trace/state laws agree, while approximate feasibility uses a declared discrepancy tolerance.
Neither reachable set need contain the other, and both depend on $\Xi$ and potentially $c$.
Finite search failure does not prove nonreachability, and budget-violating effects are inadmissible.
For multi-round targets, replace $\Gamma_c$ by the declared policy's rollout law.

These definitions imply neither disjoint neural effects nor persistence determined by artifact type.

\subsection{Selection and Family Restriction}
\label{sec:app-theory-selection}

We bound stateless full-information selection in~\eqref{eq:common-accounting}, then distinguish restriction, coordinatewise selection, and candidate discovery.

\subsubsection{Finite Candidate Selection}
\label{sec:app-finite-selection}

Section~\ref{sec:finite-selection} states the finite-family update~\eqref{eq:finite-ew} and selection-plus-restriction bound~\eqref{eq:finite-common-bound}.
The proof is in Appendix~\ref{sec:app-proofs-selection}.
With a uniform prior over $N_e>1$ candidates, choosing $\eta=\sqrt{8\log N_e/T}$ bounds selection regret by $\sqrt{T\log N_e/2}$ without knowing the best candidate.
Restriction can remain linear, although expanding a family can reduce it while increasing selection complexity, independently of the mode label.
A singleton family incurs no diagonal selection regret.

\subsubsection{Measurable Priors and Comparator Complexity}
\label{sec:measurable-selection}

To include continuous parameters, let $(\mathcal A,\mathcal F_{\mathcal A})$ be a standard-Borel space, not necessarily a product, with prior $\mu$ fixed before the losses.
Deployment of $a=(z,w)$ has measurable expected loss
\begin{equation}
    L_t(a)=L_t(z,w)
    =
    \mathbb{E}_{\tau\sim\operatorname{exec}(M_{\theta(w)},z,x_t)}
    [\ell(x_t,\tau)]
    \in[0,1].
    \label{eq:stateless-expected-loss}
\end{equation}
This evaluator averages model, tool, and harness randomness in~\eqref{eq:execution-loss}, whereas Section~\ref{sec:protocol} fixes exogenous randomness.
The bounds apply to either table under its comparator averaging convention.

Loss $L_t$ may depend on earlier history but is fixed before $a_t\sim p_t$.
The learner incurs $L_t(a_t)$ and observes all of $L_t$.
Given pre-draw information $\mathcal H_{t-1}$, its conditional mean loss is
\begin{equation}
    \overline L_t
    =
    \mathbb{E}[L_t(a_t)\mid\mathcal{H}_{t-1}]
    =
    \int_{\mathcal A}L_t(a)\,\mathrm{d}p_t(a).
    \label{eq:conditional-mean-loss}
\end{equation}
Averaging over the learner's draw defines conditional pseudo-regret, and an outer expectation gives expected pseudo-regret.

\begin{definition}[Point and distributional external regret]
\label{def:external-regret}
For a fixed reference configuration $a^\star\in\mathcal{A}$, define
\begin{equation}
    R_T(a^\star)
    =
    \sum_{t=1}^T\overline L_t
    -\sum_{t=1}^T L_t(a^\star).
\end{equation}
For a fixed probability measure $Q$ on $\mathcal A$, define
\begin{equation}
    R_T(Q)
    =
    \sum_{t=1}^T\overline L_t
    -\sum_{t=1}^T\int_{\mathcal A}L_t(a)\,\mathrm{d}Q(a).
    \label{eq:distributional-external-regret}
\end{equation}
The comparator holds one draw from fixed $Q$ throughout the sequence, with $R_T(a^\star)=R_T(\delta_{a^\star})$.
Overall point regret is
\begin{equation}
    R_T^{\mathrm{ext}}=\sup_{a^\star\in\mathcal{A}}R_T(a^\star)=
    \sum_{t=1}^T\overline L_t-\inf_{a\in\mathcal{A}}\sum_{t=1}^T L_t(a).
\end{equation}
The infimum need not be attained.
\end{definition}

Definition~\ref{def:external-regret} excludes action-dependent state trajectories.
For probability measures $Q,P$, define
\begin{equation}
    \operatorname{KL}(Q\,\|\,P)
    =
    \int\log\!\left(\frac{\mathrm{d}Q}{\mathrm{d}P}\right)\,\mathrm{d}Q,
    \label{eq:kl-divergence}
\end{equation}
when $Q\ll P$, and $+\infty$ otherwise.
Here $Q\ll P$ denotes absolute continuity and $\mathrm dQ/\mathrm dP$ the relative density.
For $P=\mu$, KL measures prior-relative comparator complexity.

Exponential weights initialized by $\mu$ uses the normalizer:
\begin{equation}
    Z_t
    =
    \int_{\mathcal A}
    \exp\!\left(-\eta\sum_{r=1}^{t-1}L_r(a)\right)\,\mathrm{d}\mu(a),
\end{equation}
and sampling measure $p_t$ with density
\begin{equation}
    \frac{\mathrm{d}p_t}{\mathrm{d}\mu}(a)
    =
    \frac{\exp(-\eta\sum_{r=1}^{t-1}L_r(a))}{Z_t}.
    \label{eq:measurable-exponential-weights}
\end{equation}
Here $Z_1=1$ and $p_1=\mu$.
This update is measurable-space Hedge~\citep{littlestone1994weighted,freund1997decision,cesabianchi2006prediction}.

Theorem~\ref{thm:prior-complexity} gives the arbitrary-rate and optimized bounds.
The proofs of the selection results below appear in Appendix~\ref{sec:app-proofs-selection}.

\begin{corollary}[Product-prior complexity decomposition]
\label{cor:product-complexity}
For $\mathcal A=\mathcal Z\times\mathcal W$, $\mu=\pi\otimes\rho$, and finite-KL $Q$ with marginals $Q_Z,Q_W$,
\begin{equation}
    \operatorname{KL}(Q\,\|\,\pi\otimes\rho)=\operatorname{KL}(Q_Z\,\|\,\pi)+\operatorname{KL}(Q_W\,\|\,\rho)+I_Q(Z;W).
    \label{eq:joint-kl-decomposition}
\end{equation}
Here $I_Q=\operatorname{KL}(Q\,\|\, Q_Z\otimes Q_W)$ measures comparator dependence, not loss interaction.
\end{corollary}

For a product comparator $Q=Q_Z\otimes Q_W$, $I_Q(Z;W)=0$.
A frozen language model can induce the text prior, while the parameter prior may encode rank, precision, sparsity, or provenance.

\begin{corollary}[Prior-atom description-length regret]
\label{cor:prior-atom-regret}
Under the product prior $\mu=\pi\otimes\rho$, for positive-mass atoms $z^\star,w^\star$, put $C_\pi=-\log\pi(\{z^\star\})$ and $C_\rho=-\log\rho(\{w^\star\})$.
Then $R_T(z^\star,w^\star)\le(C_\pi+C_\rho)/\eta+\eta T/8$, optimized to $\sqrt{T(C_\pi+C_\rho)/2}$ when the complexity is positive.
\end{corollary}

For an autoregressive language-model prior with almost-sure termination or bounded serialization length, $C_\pi$ is a text description length.
Zero-mass points have infinite pointwise KL, so they require a nearby positive-mass comparator and a regularity condition.

\begin{corollary}[Local point regret on continuous spaces]
\label{cor:local-point-regret}
Let $\mathcal A=\mathcal Z\times\mathcal W$, $\mu=\pi\otimes\rho$, and let $d_Z,d_W$ be metrics with measurable balls.
Suppose every loss satisfies
\begin{equation}
    |L_t(z,w)-L_t(z',w')|\le G_Zd_Z(z,z')+G_Wd_W(w,w'),
    \label{eq:coordinate-lipschitz-loss}
\end{equation}
for nonnegative $G_Z,G_W$ uniform over rounds.
Fix $a^\star=(z^\star,w^\star)$ and radii $\delta_Z,\delta_W>0$ whose balls $\mathbb B_Z=B(z^\star,\delta_Z)$ and $\mathbb B_W=B(w^\star,\delta_W)$ have positive respective prior mass.
Set $C_{\delta_Z,\delta_W}(a^\star)=-\log\pi(\mathbb B_Z)-\log\rho(\mathbb B_W)$.
For every $\eta>0$,
\begin{equation}
    R_T(a^\star)\le \frac{C_{\delta_Z,\delta_W}(a^\star)}{\eta}
    +\frac{\eta T}{8}+T(G_Z\delta_Z+G_W\delta_W).
    \label{eq:local-point-regret}
\end{equation}
\end{corollary}

Shrinking the balls reduces localization loss but may increase complexity.
For positive local complexity, optimizing $\eta$ gives
\begin{equation}
    R_T(a^\star)\le\sqrt{\frac{TC_{\delta_Z,\delta_W}(a^\star)}{2}}
    +T(G_Z\delta_Z+G_W\delta_W).
\end{equation}
For mixed spaces, replace one ball by an atom.
The bound is sublinear if $C_{\delta_Z,\delta_W}(a^\star)=o(T)$ and $G_Z\delta_Z+G_W\delta_W=o(1)$.
Without regularity, however, the distributional theorem does not control zero-mass points.

\paragraph{Learning-Rate and Computational Interpretation.}
The arbitrary-$\eta$ result holds simultaneously over finite-KL comparators, whereas optimized rates are comparator-tuned.
Grids, doubling, or parameter-free methods remove the need for advance knowledge at an additional regret cost~\citep{orabona2016coin}, although preserving low-refresh guarantees is nontrivial.
The guarantee is statistical and does not require Gibbs normalization, sampling, or candidate proposal to be computationally efficient.

\subsubsection{Mode Restriction and the Value of Joint Configurations}
\label{sec:app-mode-restriction}

The all-admissible family values and gaps in Proposition~\ref{prop:selection-restriction} depend on $\Xi$, the evaluator, and $T$, independently of hard-target feasibility; see Definition~\ref{def:reachability}.
Joint access need not measure interaction: $L(z,w)=1-0.4z-0.4w$ on $\{0,1\}^2$ with base $(0,0)$ has a positive joint-effect gap but zero factorial interaction.
These gaps also differ from finite-candidate restriction in the \hyperref[sec:exp-stateless-applied]{Spider comparison} and per-use target discrepancy $\varepsilon_e$; see Appendix~\ref{sec:deployment} for the latter.

For mode-restricted selection under the Cartesian specialization, use slice priors:
\begin{align}
    \mu_Z&=\pi\otimes\delta_{w_0}
    \quad\text{on }\mathcal A_Z,\label{eq:text-slice-prior}\\
    \mu_W&=\delta_{z_0}\otimes\rho
    \quad\text{on }\mathcal A_W.
    \label{eq:mode-specific-priors}
\end{align}
These priors avoid conditioning on a possibly zero-mass product slice.
Proposition~\ref{prop:selection-restriction} gives the exact selection--approximation--restriction identity and its EW bound.
Appendix~\ref{sec:candidate-discovery} further splits comparator approximation into localization and discovery.

\begin{corollary}[Mixture-prior mode selection]
\label{cor:mixture-mode-selection}
Let $\mathcal E\subseteq\{Z,W,ZW\}$ be nonempty, with priors $\mu_e$ supported on $\mathcal A_e$ and positive weights $\alpha_e$ summing to one.
Extend each mode prior by zero to the union and define
\begin{equation}
    \mu_{\mathrm{mix}}=\sum_{e\in\mathcal E}\alpha_e\mu_e.
    \label{eq:mixture-mode-prior}
\end{equation}
EW with this mixture prior satisfies, for fixed finite-KL $Q_e\ll\mu_e$,
\begin{equation}
    R_T(Q_e)\le(\operatorname{KL}(Q_e\,\|\,\mu_e)+\log(1/\alpha_e))/\eta+\eta T/8.
    \label{eq:main-mixture-mode-regret}
\end{equation}
For $e\in\{Z,W\}$, add the proposition's approximation and restriction gaps to compare with the joint optimum.
\end{corollary}

Overlapping modes contribute additively to the mixture.
Choosing $\mu_{ZW}=\pi\otimes\rho$ preserves~\eqref{eq:joint-kl-decomposition}.

To keep mode identity observable, use the tagged lift:
\begin{align}
    \widetilde{\mathcal A}_{\mathcal E}
    &=\bigsqcup_{e\in\mathcal E}(\{e\}\times\mathcal A_e),\\
    \widetilde\mu_{\mathrm{mix}}
    &=\sum_{e\in\mathcal E}\alpha_e(\delta_e\otimes\mu_e),
\end{align}
with projected loss $\widetilde L_t(e,a)=L_t(a)$.
EW samples $(e,a)$ and deploys $a$, retaining mode identity where families overlap.
The lifted comparator has exact KL
\begin{equation}
    \operatorname{KL}(\delta_e\otimes Q_e\,\|\,
    \widetilde\mu_{\mathrm{mix}})
    =\operatorname{KL}(Q_e\,\|\,\mu_e)+\log(1/\alpha_e).
\end{equation}
Without tags, overlap can reduce KL, so this certificate need not identify a posterior mode cost.

Equal weights charge at most $\log 2$ for two modes or $\log 3$ with the joint mode, while approximation or restriction can remain $\Theta(T)$.
Full-information integration neither proposes candidates nor certifies feasibility or deployment cost.
Bandit variants additionally introduce action-set dependence~\citep{auer2002bandit,bubeck2012regret}.

\subsubsection{Joint and Coordinatewise Selection}
\label{sec:app-coordinate-selection}

Use the finite Cartesian family, product prior, exogenous loss table, full feedback, and common initialization of Section~\ref{sec:coordinate-selection}.
For a positive deterministic learning-rate schedule $\eta_t$, define the common target distribution:
\begin{equation}
    q_t(z,w)=
    \frac{\pi(z)\rho(w)\exp\!\left(-\eta_t\sum_{r<t}d_r(z,w)\right)}
    {\sum_{(u,v)\in\mathcal C}\pi(u)\rho(v)
    \exp\!\left(-\eta_t\sum_{r<t}d_r(u,v)\right)}.
    \label{eq:coordinate-ew-target}
\end{equation}
The constant choice $\eta_t=\eta$ recovers~\eqref{eq:finite-ew}, whereas a varying $\eta_t$ rescales the entire cumulative loss.
A sweep preserves $q_t$ when initialized from it, but the preceding configuration need not have that law.

Additivity and a product prior factorize the EW target; see Proposition~\ref{prop:coordinate-additive-equivalence}.
Bounded multiplicative losses $d_t(z,w)=g_t(z)h_t(w)$ generally do not factorize it, but nonadditivity alone need not separate regret.

\paragraph{Exact Constants for the Fixed-Rate Separation.}
For the barrier loss $f$ of Proposition~\ref{prop:coordinate-fixed-rate-separation}, the exact constants are as follows.
For every fixed $\eta_t=\eta>0$ and every $T\geq1$, the selectors above satisfy
\begin{align}
    R_T^{\mathrm{joint}}
    &\leq \frac{3\Delta+2b}{4}
    +\frac{\Delta}{e^{\eta\Delta}-1}
    +\frac{2(\Delta+b)}{e^{\eta(\Delta+b)}-1}
    \leq 1+\frac3\eta,
    \label{eq:joint-fixed-rate-separation}\\
    R_T^{\mathrm{coord}}
    &\geq\frac{\Delta T}{4}
    \exp\!\left(-\frac{2}{e^{\eta b}-1}\right).
    \label{eq:coordinate-fixed-rate-separation}
\end{align}
The linear lower bound is not uniform as $\eta$ vanishes.
A slower schedule separates regret even when both selectors eventually reach the optimum.

\paragraph{Exact Constants for the Convergent Separation.}
Under Corollary~\ref{cor:coordinate-convergent-separation}, the value of $\eta_1>0$ does not affect uniform initialization.
For $\gamma=\Delta/b>1$ and $T\geq2$,
\begin{align}
    R_T^{\mathrm{joint}}
    &\leq\frac{3\Delta+2b}{4}
    +\Delta 4^{-\gamma}\left(1+\frac1{\gamma-1}\right)
    +2(\Delta+b)4^{-(\gamma+1)}\left(1+\frac1\gamma\right),
    \label{eq:joint-convergent-separation}\\
    R_T^{\mathrm{coord}}
    &\geq\frac{\Delta e^{-1/2}}4
    \sum_{n=1}^{T-1}n^{-1/2}
    =\Omega(\sqrt T).
    \label{eq:coordinate-convergent-separation}
\end{align}
These separations concern the specified conditional sampler, not arbitrary alternating optimization.
More sweeps, global refreshes, or deliberate transitions through higher-loss configurations can change its behavior.
Exact joint sampling may also be costly, and these results establish no computational advantage.
The \hyperref[sec:exp-coordinate-selection]{controlled selection study} in Section~\ref{sec:experiments} tests interaction, sweep count, and learning rate by exact probability propagation.

\subsubsection{Candidate Proposal, Discovery, and Selection}
\label{sec:candidate-discovery}

Candidate construction uses scores, traces, gradients, or critiques~\citep{zhou2023ape,deng2022rlprompt,pryzant2023protegi,yang2024opro,guo2024evoprompt,fernando2024promptbreeder,yuksekgonul2025textgrad,feedbackdescent,metaharness}.
We separate discovery from selection on a fixed family.
Changing families instead require tracking, sleeping-expert, or growing-expert guarantees~\citep{herbster1998tracking,cortes2019sleeping,mourtada2017growing}.

Let $\widehat L_{\mathrm{train}}$ and $\widehat L_{\mathrm{valid}}$ denote empirical losses on the declared training and validation data.
A bilevel joint proposal fits, for each proposed harness $z$,
\begin{equation}
    \widehat w(z)
    \approx
    \mathop{\arg\min}_{w:(z,w)\in\mathcal A}
    \widehat L_{\mathrm{train}}(z,w),
\end{equation}
then chooses
\begin{align}
    \widehat z
    &\approx
    \mathop{\arg\min}_{z:\,\exists w\ (z,w)\in\mathcal A}
    \widehat L_{\mathrm{valid}}\bigl(z,\widehat w(z)\bigr),\\
    \widehat a&=\bigl(\widehat z,\widehat w(\widehat z)\bigr).
    \label{eq:bilevel-joint-proposal}
\end{align}
Outer validation search may be discrete and inner fitting gradient-based, but these approximate minimizations do not establish finite-budget convergence.
Alternating procedures use~\eqref{eq:text-update-set}--\eqref{eq:parameter-update-set}: BetterTogether alternates prompt optimization and fine-tuning~\citep{SoyluD2024emnlp}, while WHALE alternates model updates and executable-harness search~\citep{KimH2026whale}.
They motivate, but differ from, the fixed-family conditional sampler in Section~\ref{sec:coordinate-selection}.
Our guarantees apply after the family and prior are fixed.

For nonempty measurable $\mathcal C_e=\mathcal C\cap\mathcal A_e$, $e\in\{Z,W\}$, define
\begin{align}
    V_{T,\mathcal C}^e
    & = \inf_{a\in\mathcal C_e}\sum_{t=1}^T L_t(a),\\
    \mathsf{Disc}_T^e(\mathcal C_e)
    & = V_{T,\mathcal C}^e-V_T^e\geq0,\label{eq:family-discovery-gap}\\
    \Lambda_{T,\mathcal C}^e(Q_e)
    & =
    \sum_{t=1}^T\int_{\mathcal C_e}L_t(a)\,\mathrm dQ_e(a)
    -V_{T,\mathcal C}^e\geq0,
    \label{eq:family-localization-gap}
\end{align}
where $Q_e$ is supported on $\mathcal C_e$.
Discovery charges the candidate family's gap to the all-admissible optimum $V_T^e$, while localization charges $Q_e$'s gap to the best candidate in the family.
Substituting $\mathsf{Apx}_T^e(Q_e)=\Lambda_{T,\mathcal C}^e(Q_e)+\mathsf{Disc}_T^e(\mathcal C_e)$ into~\eqref{eq:main-selection-restriction} gives
\begin{align}
    \sum_{t=1}^T\overline L_t-V_T^{ZW}
    ={}&\underbrace{\left[
        \sum_{t=1}^T\overline L_t
        -\sum_{t=1}^T\int_{\mathcal C_e}L_t(a)\,\mathrm dQ_e(a)
    \right]}_{\text{selection}}
    +\underbrace{\Lambda_{T,\mathcal C}^e(Q_e)}_{\text{localization}}\nonumber\\
    &+\underbrace{\mathsf{Disc}_T^e(\mathcal C_e)}_{\text{discovery}}
    +\underbrace{\Delta_e^{\mathrm{mode}}(T)
    +\Delta^{\mathrm{joint}}(T)}_{\text{mode and joint restriction}}.
    \label{eq:four-way-family-accounting}
\end{align}
A family prior controls only selection.
Localization vanishes for a point mass on the best atomic candidate, while discovery remains relative to the proposals and budget.
Non-Cartesian families require a joint prior and KL.

\subsection{State Effects and Policy Regret}
\label{sec:app-theory-state}

Bounded influence and full diagonal feedback control the state term in~\eqref{eq:common-accounting}, and resets permit whole-episode policy selection.
Section~\ref{sec:feedback} weakens the feedback assumptions.

\subsubsection{Persistent State and Policy Regret}
\label{sec:persistent-policy-regret}

Section~\ref{sec:state} defines policy regret using each fixed comparator's own trajectory.
Propositions~\ref{prop:external-policy-separation} and~\ref{prop:irreversible-state} show why evaluating all configurations on the learner's state can conceal irreversible damage and why unrestricted persistent state precludes sublinear guarantees.
The constructions and proofs are in Appendix~\ref{sec:app-proofs-state}.
Relabeling gives the parameter-only obstruction when that mode admits the same transitions, although actual text and parameter updates need not be symmetric.

\subsubsection{Policy Regret under Bounded Influence}
\label{sec:bounded-influence}

Bound each coordinate's influence while allowing interactions within its history window~\citep{arora2012policy,anava2015memory}.
Fix a prior $\mu$ on $\mathcal A=\mathcal Z\times\mathcal W$ or a measurable compatibility subset supporting every refresh.

The functions in Assumption~\ref{ass:bounded-influence} have domains
\begin{equation}
    f_t:\mathcal{Z}^{\min\{m_Z+1,t\}}
    \times\mathcal{W}^{\min\{m_W+1,t\}}\rightarrow[0,1].
\end{equation}
Windows include the current coordinate, so zero horizon excludes carryover.
Bundles use their largest horizon.
For $e\in\{Z,W\}$, write $m_e$ for its coordinate horizon, truncated at $T-1$ without changing the windows.

For $a=(z,w)\in\mathcal{A}$, define the diagonal loss:
\begin{equation}
    d_t(a)
    =
    f_t\!\left(
        \underbrace{z,\ldots,z}_{\min\{m_Z+1,t\}};
        \underbrace{w,\ldots,w}_{\min\{m_W+1,t\}}
    \right).
    \label{eq:bounded-diagonal-loss}
\end{equation}
This specializes~\eqref{eq:common-diagonal-loss}.
Assumption~\ref{ass:diagonal-oracle} requires valid simulation or cloning of each constant history.
Ordinary full information on the learner's realized history does not reveal these losses.

\paragraph{Lazy Exponential Weights.}
Apply exponential weights~\eqref{eq:measurable-exponential-weights} to the diagonal oracle, coupling draws by measurable Shrinking Dartboard to control state~\citep{geulen2010regret,kalai2005efficient}.
For $\eta>0$, set $L_t=d_t$ in~\eqref{eq:measurable-exponential-weights} to define the target marginal $q_t$, and draw $A_1=(z_1,w_1)\sim\mu$.
After deploying $A_t=(z_t,w_t)$ and observing $d_t$, update
\begin{align}
    \zeta_t
    &=
    \int_{\mathcal A}e^{-\eta d_t(a)}\,\mathrm{d}q_t(a),\\
    q_{t+1}(\mathrm{d}a)
    &=
    \frac{e^{-\eta d_t(a)}}{\zeta_t}q_t(\mathrm{d}a).
    \label{eq:lazy-ew-update}
\end{align}
For $t<T$, retain $A_t$ with probability $e^{-\eta d_t(A_t)}$ and otherwise \emph{refresh} from $q_{t+1}$, with indicator $Y_t$.
Refreshes may redraw $A_t$, so the number of switches is at most $\sum_{t<T}Y_t$.
Under $A_t\sim q_t$, $\zeta_t$ is the mean retention probability.
At $\eta=0$, draw once from $\mu$ and never refresh.

\begin{lemma}[Lazy exponential-weights coupling]
\label{lem:lazy-marginals}
The lazy algorithm has marginals $A_t\sim q_t$.
For $t=1,\ldots,T-1$,
\begin{align}
    \mathbb E[Y_t]
    &\leq
    \min\left\{
        \eta\,\mathbb E[d_t(A_t)],
        1-e^{-\eta}
    \right\}
    \leq\eta,\\
    \mathbb E\left[\sum_{t=1}^{T-1}Y_t\right]
    &\leq\eta(T-1).
    \label{eq:lazy-refresh-bound}
\end{align}
\end{lemma}

To control the signed state effect isolated by Proposition~\ref{prop:matched-marginals}, count departures from constant histories.

\begin{lemma}[Contamination accounting]
\label{lem:contamination}
Under Assumption~\ref{ass:bounded-influence}, for any realized sequence $A_t=(z_t,w_t)$, put
\begin{equation}
    \chi_s
    =
    \max\left\{
        m_Z\mathbf 1\{z_s\neq z_{s-1}\},
        m_W\mathbf 1\{w_s\neq w_{s-1}\}
    \right\},
    \qquad s=2,\ldots,T.
\end{equation}
The charge $\chi_s$ uses the larger horizon for simultaneous changes.
Let $N_T^{\mathrm{cont}}$ count rounds with a nonconstant coordinate history.
Pathwise,
\begin{align}
    \left|L_T^{\mathrm{dep}}-L_T^{\mathrm{diag}}\right|
    &\leq N_T^{\mathrm{cont}}\nonumber\\
    &\leq\sum_{s=2}^T\chi_s.
    \label{eq:pathwise-switch-contamination}
\end{align}
Overlap can make the summed charges loose.
For the lazy algorithm,
\begin{equation}
    \sum_{s=2}^T\chi_s
    \leq
    m_{\max}\sum_{t=1}^{T-1}Y_t.
    \label{eq:refresh-contamination}
\end{equation}
\end{lemma}

Theorem~\ref{thm:lazy-policy} combines these selection and contamination bounds, with the exact learning rate in~\eqref{eq:lazy-learning-rate}.

For mixture prior~\eqref{eq:mixture-mode-prior}, with each $\mu_e$ supported on its slice $\mathcal A_e$ and $Q_e\ll\mu_e$, replace $C_Q$ by
\begin{equation}
    \overline C_e^{\mathrm{mix}}
    =
    \operatorname{KL}(Q_e\,\|\,\mu_e)+\log(1/\alpha_e),
\end{equation}
which is exact on the tagged lift and may exceed untagged KL.
Whole-pair refreshes still charge $m_{\max}$, even when comparator complexity concentrates on one mode.

\begin{corollary}[Lazy local point policy regret]
\label{cor:lazy-local-point-regret}
Under Theorem~\ref{thm:lazy-policy}, suppose $\mu=\pi\otimes\rho$ and the diagonal losses satisfy the Lipschitz condition of Corollary~\ref{cor:local-point-regret}.
For its metrics, positive-mass balls, and local complexity, every $\eta>0$ gives
\begin{equation}
    R_T^{\mathrm{pol}}(a^\star)\le C_{\delta_Z,\delta_W}(a^\star)/\eta
    +\eta\left[T/8+m_{\max}(T-1)\right]+T(G_Z\delta_Z+G_W\delta_W).
    \label{eq:lazy-local-policy-regret}
\end{equation}
\end{corollary}

For positive local complexity, optimizing $\eta$ gives
\begin{equation}
    R_T^{\mathrm{pol}}(a^\star)\le
    2\sqrt{C_{\delta_Z,\delta_W}(a^\star)[T/8+m_{\max}(T-1)]}
    +T(G_Z\delta_Z+G_W\delta_W).
\end{equation}
At zero local complexity, one conditioned-prior draw leaves only localization loss, whereas a product-prior atom needs no localization.

Proposition~\ref{prop:mode-lazy-rates} in Section~\ref{sec:state} gives the mode-restricted bounds for the supported slice priors~\eqref{eq:text-slice-prior}--\eqref{eq:mode-specific-priors}.
Zero KL permits a fixed prior draw, atoms give description lengths, and balls add localization.
Proofs are in Appendix~\ref{sec:app-proofs-state}.

\subsubsection{Episodic Optimization with Resets}
\label{sec:episodic-resets}

Consider $K$ episodes of $H$ rounds, each in a fresh sandbox with common resets
\begin{align}
    s_{k,1}&=s_0,\\
    a_{k,0}&=a_0=(z_0,w_0).
\end{align}
A policy $\varphi$ maps within-episode history to $a_{k,h}=(z_{k,h},w_{k,h})$ through admissible updates, whose modes may differ in effect; see Definition~\ref{def:reachability}.
Fix a standard-Borel policy family $(\mathcal P,\mathcal F_{\mathcal P})$ with prior $\Pi_0$.

For expected evaluation loss $\ell_{k,h}(s,a)\in[0,1]$, define episode loss
\begin{equation}
    J_k(\varphi)
    = \mathbb{E}_{\varphi}
    \left[
        \sum_{h=1}^H
        \ell_{k,h}(s_{k,h},a_{k,h})
    \right].
    \label{eq:episodic-policy-loss}
\end{equation}
The expectation includes execution, transitions, and policy randomness from the common reset.
Assume measurable $J_k:\mathcal P\to[0,H]$ are fixed before learner randomization, with exogenous tasks.
Given pre-selection history $\mathcal H_{k-1}$, including completed-episode feedback, define
\begin{equation}
    \overline J_k
    =
    \mathbb{E}[J_k(\varphi_k)\mid\mathcal{H}_{k-1}].
\end{equation}
This expectation also averages over policy selection, as in~\eqref{eq:conditional-mean-loss}.
For a fixed policy $\varphi^\star$ or comparator probability measure $Q_{\mathcal P}$ on $\mathcal P$, define
\begin{align}
    R_K^{\mathrm{epi}}(\varphi^\star)
    & = \sum_{k=1}^K \overline J_k
    - \sum_{k=1}^K J_k(\varphi^\star),\\
    R_K^{\mathrm{epi}}(Q_{\mathcal P})
    & = \sum_{k=1}^K \overline J_k
    - \sum_{k=1}^K\int_{\mathcal P}J_k(\varphi)\,\mathrm{d}Q_{\mathcal P}(\varphi).
\end{align}
The comparator holds one policy draw across episodes, while allowing within-episode configuration changes.
Thus~\eqref{eq:episodic-policy-loss} extends the own-trajectory comparison to policies.

Specializing~\eqref{eq:measurable-exponential-weights} to $\mathcal P$, sample the episode-$k$ policy from
\begin{equation}
    \frac{\mathrm{d}q_k}{\mathrm{d}\Pi_0}(\varphi)
    =
    \frac{\exp(-\eta\sum_{i=1}^{k-1}J_i(\varphi))}
    {\int_{\mathcal P}
     \exp(-\eta\sum_{i=1}^{k-1}J_i(\psi))\,\mathrm{d}\Pi_0(\psi)}.
    \label{eq:episodic-measurable-weights}
\end{equation}
Execute it for the episode, then reveal $J_k$ for every policy, including unexecuted ones.

\begin{theorem}[Episodic policy prior-complexity regret]
\label{thm:episodic-policy-regret}
Under the preceding reset, exogeneity, measurability, and full-feedback assumptions, for every $\eta>0$ and fixed $Q_{\mathcal P}\ll\Pi_0$ with finite $C_{\mathcal P}=\operatorname{KL}(Q_{\mathcal P}\|\Pi_0)$, episode-level EW satisfies
\begin{equation}
    R_K^{\mathrm{epi}}(Q_{\mathcal P})
    \le \frac{C_{\mathcal P}}{\eta}+\frac{\eta KH^2}{8}.
    \label{eq:episodic-kl-regret}
\end{equation}
For $C_{\mathcal P}>0$, $\eta=\sqrt{8C_{\mathcal P}/(KH^2)}$ gives $R_K^{\mathrm{epi}}\le H\sqrt{KC_{\mathcal P}/2}$.
\end{theorem}

The proof is in Appendix~\ref{sec:app-proofs-state}.
Predictably reactive episode losses instead yield realized-sequence external pseudo-regret.

As in Corollary~\ref{cor:prior-atom-regret}, atoms give description length $-\log\Pi_0(\{\varphi^\star\})$.
For measurable metric balls and episode losses uniformly $G_{\mathcal P}$-Lipschitz in policy distance, conditioning on the radius-$\delta$ ball $B(\varphi^\star,\delta)$ gives complexity $-\log\Pi_0(B(\varphi^\star,\delta))$ and localization loss $KG_{\mathcal P}\delta$, as in Corollary~\ref{cor:local-point-regret}.

Restricted policy families may still incur the gaps in Section~\ref{sec:mode-restriction}.

\subsection{Feedback Requirements and Extensions}
\label{sec:app-theory-feedback}

The feedback interface determines control of selection and state in~\eqref{eq:common-accounting}.
Full diagonal feedback evaluates each candidate's own trajectory, sparse probing reduces evaluations, and deployment-only learning requires control of causal reaction.
Appendix~\ref{sec:feedback-comparators} specifies when the guarantees retain the fixed joint-family comparator.
These guarantees neither construct candidates nor rank unrestricted families.

Sections~\ref{sec:state}--\ref{sec:feedback} present the main guarantees for full counterfactual feedback, sparse probes, and deployment-only feedback.
Here $K$ counts candidates, not episodes as in Appendix~\ref{sec:episodic-resets}.
We use $C_Q=\operatorname{KL}(Q\Vert\mu)$ and $m=m_{\max}$ for comparator complexity and the maximum influence horizon, respectively.
The comparator-tuned rates concern expectations, while the exact bounds cover zero complexity, zero influence, and finite-horizon boundaries.

Exponential weights and importance weighting~\citep{auer2002bandit}, low-switch couplings~\citep{geulen2010regret,anava2015memory}, and policy-regret mini-batching~\citep{arora2012policy} supply the algorithmic ingredients.
Decoupled exploration and deployment~\citep{AvnerO2012icml} and limited-observation Shrinking Dartboard~\citep{ChengD2023icml} provide further precedents.

\subsubsection{Accuracy of Full Diagonal Evaluation}
\label{sec:full-feedback-accuracy}

These results relax Assumption~\ref{ass:diagonal-oracle} conditional on an exogenous response tree.
For random trees, the relevant independence, mean, or error assumptions must hold conditionally.
Unconditional unbiasedness does not justify a tree-dependent hindsight comparator; see Appendix~\ref{sec:feedback-comparators}.

\begin{corollary}[Approximate diagonal evaluation]
\label{cor:approximate-diagonal-evaluation}
Under Assumption~\ref{ass:bounded-influence}, let the entire estimate sequence be independent of all learner randomness, with $\widehat d_t\in[0,1]$ and $\sup_a|\widehat d_t(a)-d_t(a)|\le\epsilon_t$ almost surely for fixed $\epsilon_t\ge0$.
Lazy EW using these estimates satisfies, for every $\eta>0$ and $Q\ll\mu$ with finite $C_Q$,
\begin{equation}
    R_T^{\mathrm{pol}}(Q)\le \frac{C_Q}{\eta}+\eta\left[\frac{T}{8}+m_{\max}(T-1)\right]
    +2\sum_{t=1}^T\epsilon_t.
    \label{eq:approximate-diagonal-policy-regret}
\end{equation}
\end{corollary}

Independence from only the current draw is insufficient.
Off-policy estimates require additional coverage, variance, and validity assumptions~\citep{dudik2011doubly,jiang2016doubly}.
Proofs of the feedback results are in Appendix~\ref{sec:app-proofs-feedback}.

\begin{corollary}[Finite-family sampling cost for diagonal estimation]
\label{cor:diagonal-sampling-cost}
Let $\mu$ be supported on a $K$-candidate family $\mathcal C\subseteq\mathcal A$, restricting the learner and comparators $Q\ll\mu$ to it.
Fix $\delta\in(0,1)$.
Estimate each $d_t(a)$ by the mean of $n\geq1$ independent $[0,1]$ samples with mean $d_t(a)$, using a sampling mechanism exogenous to all learner randomness.
Define
\begin{equation}
    \epsilon_n=\sqrt{\frac{\log(2KT/\delta)}{2n}}.
\end{equation}
With probability at least $1-\delta$, all $KT$ estimates satisfy
$|\widehat d_t(a)-d_t(a)|\leq\epsilon_n$ simultaneously.
Under the remaining assumptions of
Corollary~\ref{cor:approximate-diagonal-evaluation}, policy regret averaged over
estimation and learner randomness satisfies
\begin{equation}
    R_T^{\mathrm{pol}}(Q)
    \leq\frac{C_Q}{\eta}+\frac{\eta T}{8}
         +m_{\max}\eta(T-1)+2T\epsilon_n+\delta T.
    \label{eq:sampled-diagonal-regret}
\end{equation}
The estimator uses $KTn$ oracle samples.
\end{corollary}

Uniform accuracy $\epsilon$ is guaranteed by $n\geq\log(2KT/\delta)/(2\epsilon^2)$, and the envelope can be capped at one.
Oracle costs must include simulation or cloning, while logged language-model counterfactuals need separate unbiasedness justification.
With fixed samples per pair, the envelope need not vanish, so this certificate alone does not imply sublinear regret.

\begin{corollary}[Fresh unbiased diagonal feedback]
\label{cor:unbiased-diagonal}
Fix a finite compatible $K$-candidate family with positive prior $\mu$.
Under Assumption~\ref{ass:bounded-influence}, let the entire estimate sequence be independent of all learner randomness, with $\widehat d_t\in[0,1]$ and
\begin{equation}
    \mathbb E[\widehat d_t(a)\mid\widehat d_{1:t-1}]=d_t(a).
\end{equation}
For lazy EW using these estimates, every $\eta>0$ and fixed $Q$ satisfy, averaging over estimation and learner randomness,
\begin{equation}
    R_T^{\mathrm{pol}}(Q)\le \frac{C_Q}{\eta}+\eta\left[\frac{T}{8}+m_{\max}(T-1)\right].
    \label{eq:unbiased-diagonal-regret}
\end{equation}
\end{corollary}

Fresh independent oracle samples satisfy Corollary~\ref{cor:unbiased-diagonal} at $n=1$.
Larger $n$ improves uniform accuracy but is unnecessary for this coarse bound.
Biased or trajectory-dependent estimates do not inherit it.
The $2\sum_t\epsilon_t$ error bridge compares true and estimated learner and comparator losses for the \emph{same} policy, not learners trained on different feedback.
An offset $\widehat d_t(a)=d_t(a)+c_t\in[0,1]$ preserves EW marginals but multiplies retention~\eqref{eq:lazy-ew-update} by $e^{-\eta c_t}$, changing state effects; see the \hyperref[sec:exp-approximate-feedback]{feedback accuracy study} in Section~\ref{sec:experiments}.

\subsubsection{Causal Responses and Comparator Conventions}
\label{sec:causal-feedback}

Fix a nonempty finite compatible family $\mathcal C\subseteq\mathcal A$, $K=|\mathcal C|$, and prior $\mu(a)>0$.
Neither a Cartesian family nor a factorized prior is required.

\begin{definition}[Causal response and expected policy regret]
\label{def:causal-response}
Let $U$ be an external random tape independent of the learner's private randomization, and let $F_t:\mathcal C^t\times\mathcal U\to[0,1]$ be the measurable causal response functions from Section~\ref{sec:protocol}.
Realized loss is $F_t(A_{1:t};U)$.
For a constant history $a^t=(a,\ldots,a)$, write $d_t(a;U)=F_t(a^t;U)$, making the tape explicit in~\eqref{eq:common-diagonal-loss}.
Policy regret is as defined in Section~\ref{sec:state}, averaging over $U$ and learner randomness, with $Q$ fixed independently of learner observations and randomization.
\end{definition}

Each intervention reruns task selection, user responses, and environment state through $F_t$.
This defines counterfactuals but supplies no evaluation oracle.
The environment may observe actions and outcomes, not future private coins.
For observed announcements or messages, their rules and intervention-dependent values must also be specified.

A deterministic tree's best fixed candidate is admissible.
Random environments distinguish best expected loss from realized hindsight:
\begin{equation}
    \min_a\mathbb E D_T(a)\neq\mathbb E\min_aD_T(a).
    \label{eq:expected-versus-hindsight-optimum}
\end{equation}
Conditioning on an exogenous tree can justify hindsight comparison; see Appendix~\ref{sec:feedback-comparators}.
Conditioning on noisy probes, however, cannot justify a probe-dependent comparator in an unbiasedness argument.

\subsubsection{Predictability Does Not Justify Conditional Lazy Marginals}
\label{sec:adaptive-lazy-pitfall}
Predictability and the bounded influence specified in Assumption~\ref{ass:bounded-influence} do not supply conditional lazy marginals: a retained action can correlate with losses chosen from its past despite matching unconditional marginals.
\begin{proposition}[Two-round counterexample]
\label{prop:adaptive-lazy-failure}
A predictable second-round loss vector can satisfy
$\mathbb E\ell_2(A_2)\ne\mathbb E\langle q_2, \ell_2\rangle$ under lazy sampling,
even when $A_2$ has unconditional marginal $q_2$.
\end{proposition}
For $F_2(a_1,a_2)=\mathbf 1\{a_1=a_2\}$, both diagonal losses are $d_2(a)=1$.
The counterexample invalidates substituting current-action losses on the realized prefix, not Theorem~\ref{thm:lazy-policy}'s diagonal oracle.

\begin{proposition}[External regret versus few switches]
\label{prop:external-switch-tradeoff}
Let $\ell_1\equiv0$ and $\ell_t(a)=\mathbf 1\{a=A_{t-1}\}$ for
$t\ge2$, and let
\begin{equation}
 S_T=\sum_{t=2}^T\mathbf 1\{A_t\ne A_{t-1}\}.
\end{equation}
Every learner satisfies
\begin{equation}
 \mathbb E\left[\sum_t\ell_t(A_t)-\min_{a\in\{0,1\}}\sum_t\ell_t(a)\right]
 \ge \frac{T-1}{2}-\mathbb E S_T.
\end{equation}
\end{proposition}
Constant-trajectory comparators lose $T-1$, excluding a policy-regret lower bound but motivating fresh block draws under reactive feedback; see Appendix~\ref{sec:deployment-only-feedback}.

\subsubsection{Exponential Weights with Nonnegative Estimates}
\label{sec:importance-weighted-ew}
Importance-weighted estimates may exceed one, invalidating the bounded-feedback term $\eta T/8$.
A standard second-order potential argument instead controls selection by their second moments~\citep{auer2002bandit,ChengD2023icml}.
\begin{lemma}[Prior-EW with nonnegative estimates]
\label{lem:nonnegative-ew}
Let $x_t(a)\ge0$ be finite, let $\eta>0$, and define
\begin{equation}
 q_t(a)=\frac{\mu(a)\exp(-\eta\sum_{s<t}x_s(a))}
 {\sum_c\mu(c)\exp(-\eta\sum_{s<t}x_s(c))}.
\end{equation}
For every distribution $Q$ on $\mathcal C$, pathwise,
\begin{equation}
 \sum_{t=1}^T\langle q_t-Q, x_t\rangle
 \le \frac{C_Q}{\eta}
       +\frac{\eta}{2}\sum_{t=1}^T\langle q_t, x_t^2\rangle.
 \label{eq:nonnegative-ew}
\end{equation}
The sequence $x_t$ may be adaptively generated.
\end{lemma}
Actual-loss bounds additionally require regime-specific sampling identities, unbiasedness, and comparator validity.

\subsubsection{Sparse Independent Diagonal Probes}
\label{sec:sparse-diagonal}

Let $A_t$ denote deployment, $J_t$ the probe, $Z_t$ its outcome, and $R_t\in\{0,1\}$ the lazy refresh indicator.

\begin{assumption}[Independent probes of an exogenous diagonal table]
\label{ass:independent-diagonal-probe}
Conditional on an exogenous response tree $U$, the following requirements hold.
Assumption~\ref{ass:bounded-influence} holds.
An autonomous evaluator uses randomness independent of all deployment coins and consumes no deployed actions or outcomes.
Its pre-probe sigma-field $\mathcal G_{t-1}$ includes $U$ and earlier probes, and may include evaluation context hidden from the learner.
The selector $q_t$ uses only previous estimates, and positive $r_t$ is $\mathcal G_{t-1}$-measurable.
Drawing $J_t\sim r_t$, the evaluator returns $Z_t\in[0,1]$ with
\begin{equation}
    \mathbb E[Z_t\mid\mathcal G_{t-1},J_t=a]=d_t(a).
    \label{eq:probe-unbiased}
\end{equation}
Probing leaves the response tree unchanged, and deployment intervention leaves the entire evaluation process unchanged.
\end{assumption}

Independent simulation noise on an exogenous table qualifies, whereas deployment-reactive evaluation does not.
Proposition~\ref{prop:query-adaptive-probes} permits probe-driven adaptation with narrower comparators.

Under Assumption~\ref{ass:independent-diagonal-probe}, initialize the selector with $q_1=\mu$ and $A_1\sim\mu$.
At round $t$, execute $A_t$,
obtain the independent probe $(J_t,Z_t)$, and form
\begin{equation}
 \widehat d_t(a)=\frac{\mathbf 1\{J_t=a\}Z_t}{r_t(a)}.
 \label{eq:sparse-diagonal-estimator}
\end{equation}
Update
\begin{equation}
 q_{t+1}(a)=\frac{q_t(a)e^{-\eta\widehat d_t(a)}}
 {\sum_cq_t(c)e^{-\eta\widehat d_t(c)}}.
 \label{eq:sparse-diagonal-update}
\end{equation}
For $t<T$, retain $A_{t+1}=A_t$ with probability
$e^{-\eta\widehat d_t(A_t)}$ and otherwise draw afresh from $q_{t+1}$.
Draws use the independent deployment tape and return no deployment-loss feedback.
A refresh need not switch configurations.

\begin{lemma}[Probe mean and second moment]
\label{lem:probe-moments}
Under Assumption~\ref{ass:independent-diagonal-probe},
\begin{align}
 \mathbb E[\widehat d_t(a)\mid\mathcal G_{t-1}]&=d_t(a),\\
 \mathbb E[\langle q_t, \widehat d_t^2\rangle\mid\mathcal G_{t-1}]&=V_t,
\end{align}
where
\begin{equation}
 V_t=\sum_a\frac{q_t(a)}{r_t(a)}
       \mathbb E[Z_t^2\mid\mathcal G_{t-1},J_t=a].
 \label{eq:probe-second-moment}
\end{equation}
In particular, $r_t=q_t$ gives $V_t\le K$.
\end{lemma}

\begin{lemma}[Conditional lazy marginals and refreshes]
\label{lem:sparse-lazy-marginals}
Under Assumption~\ref{ass:independent-diagonal-probe}, conditional on the full auxiliary tape and the fixed response tree, $A_t$ has law $q_t$.
Moreover,
\begin{equation}
 \mathbb E R_t\le\eta\mathbb E\langle q_t, d_t\rangle\le\eta,
 \qquad 1\le t<T.
 \label{eq:sparse-refresh}
\end{equation}
\end{lemma}
Conditioning requires the entire autonomous process, not current-coin independence; see Proposition~\ref{prop:adaptive-lazy-failure}.

\paragraph{Refined Bound for Arbitrary Probe Distributions.}
Theorem~\ref{thm:sparse-diagonal} in Section~\ref{sec:feedback} specializes a more detailed inequality.
Under Assumption~\ref{ass:independent-diagonal-probe}, for $m=m_{\max}$, every fixed $Q$ and $\eta>0$ satisfy
\begin{equation}
 R_T^{\mathrm{pol}}(Q)\le\frac{C_Q}{\eta}
 +\frac{\eta}{2}\sum_{t=1}^T\mathbb E V_t
 +m\eta\sum_{t=1}^{T-1}\mathbb E\langle q_t, d_t\rangle.
 \label{eq:sparse-diagonal-refined}
\end{equation}
An infinite expected second moment makes this bound vacuous.
For $r_t=q_t$, all moments are finite, and~\eqref{eq:sparse-diagonal-uniform}--\eqref{eq:sparse-diagonal-optimized} follow.
Text-only and parameter-only families use $m_Z$ and $m_W$, and non-Cartesian families use their supported joint prior.
Tuning is comparator-dependent, but uniform priors give $C_Q=\log K$ for every point comparator.

Cloning, replay, constant-history initialization, and simulation costs remain.
Finite influence supplies no simulator, and the expected bound guarantees neither low variance nor high-probability performance.

\begin{proposition}[Autonomous query-adaptive evaluation]
\label{prop:query-adaptive-probes}
The sparse-probe bounds hold for fixed $Q$ if the auxiliary process chooses response functions from earlier probes and $q_t$, independently of deployment coins and without consuming deployed actions or outcomes.
The current diagonal vector and $r_t$ must be $\mathcal G_{t-1}$-measurable and satisfy~\eqref{eq:probe-unbiased}.
Conditional on the complete auxiliary process, responses must have diagonal $d_t$ and the stated coordinate influence bounds.
Deployment intervention must leave that entire process unchanged.
\end{proposition}

This allows probe-driven curricula, not deployment reactions or realized query-dependent comparators.
The best expected-loss candidate is fixed before probes and remains admissible under~\eqref{eq:expected-versus-hindsight-optimum}.
The budget and bias extensions below inherit this fixed-$Q$ restriction for query-adaptive trees.

\subsubsection{Probe Budgets and Biased Evaluation}
\label{sec:probe-budgets}

\begin{corollary}[Multiple probes per round]
\label{cor:multiple-probes}
Hold the current evaluator context and $q_t$ fixed while drawing $k\ge1$
conditionally independent probes with $r_t=q_t$. If $\widehat d_t$ is the
average of their importance-weighted vectors, then
\begin{equation}
 \mathbb E[\langle q_t, \widehat d_t^2\rangle\mid\mathcal G_{t-1}]
 \le 1+\frac{K-1}{k}.
 \label{eq:multiple-probes}
\end{equation}
Consequently $K$ in \eqref{eq:sparse-diagonal-uniform} can be replaced by
$1+(K-1)/k$, while the expected refresh bound is unchanged.
\end{corollary}

Sampling is with replacement, whereas querying every candidate once has a different estimator law.

\begin{corollary}[Expected query budget]
\label{cor:query-budget}
Use an independent Bernoulli gate $G_t$ with probability $p\in(0,1]$ and,
when it opens, one probe drawn from $q_t$. Set the estimate to zero when
the gate is closed. The resulting estimator,
\begin{equation}
 \widehat d_t(a)=\frac{G_t\mathbf 1\{J_t=a\}Z_t}{p q_t(a)},
\end{equation}
is unbiased, with weighted second moment at most $K/p$. Writing
$B_{\mathrm{qry}}=pT$ for the \emph{expected} number of probes, the optimized
bound for positive $C_Q$ is
\begin{equation}
 R_T^{\mathrm{pol}}(Q)\le2\sqrt{C_Q\left(
      \frac{KT^2}{2B_{\mathrm{qry}}}+m(T-1)\right)}.
 \label{eq:query-budget}
\end{equation}
\end{corollary}
A hard cap makes later inclusion probabilities zero, invalidating Corollary~\ref{cor:query-budget}'s argument.
Fixed-budget designs therefore need separate analysis~\citep{ChengD2023icml}.

Simulator bias separates the true selection target from the mean driving retention.

\begin{proposition}[Biased-probe bound]
\label{prop:biased-probes}
Retain probe autonomy, but suppose
\begin{align}
 \mathbb E[Z_t\mid\mathcal G_{t-1},J_t=a]&=\bar d_t(a),\\
 \sup_a|\bar d_t(a)-d_t(a)|&\le\epsilon_t,
\end{align}
where $\epsilon_t\ge0$ is predictable, $Z_t\in[0,1]$, and $V_t$ is the
actual weighted second moment of the estimator. Then
\begin{equation}
 R_T^{\mathrm{pol}}(Q)\le{}\frac{C_Q}{\eta}
   +\frac{\eta}{2}\sum_{t=1}^T\mathbb E V_t
   +2\sum_{t=1}^T\mathbb E\epsilon_t+m\eta\sum_{t=1}^{T-1}\mathbb E\langle q_t, \bar d_t\rangle.
 \label{eq:biased-probes-refined}
\end{equation}
In particular, for $r_t=q_t$,
\begin{equation}
 R_T^{\mathrm{pol}}(Q)\le\frac{C_Q}{\eta}
   +\frac{\eta KT}{2}+m\eta(T-1)
   +2\sum_{t=1}^T\mathbb E\epsilon_t.
 \label{eq:biased-probes-uniform}
\end{equation}
\end{proposition}

Replacing $\bar d_t$ by $d_t$ in the refresh term requires an additional $m\eta\sum_{t<T}\mathbb E\epsilon_t$: even when $d_t\equiv0$, positive simulator means can trigger refreshes.

\subsubsection{Deployment-Only Feedback and Causal Reaction}
\label{sec:deployment-only-feedback}

Under Definition~\ref{def:causal-response}, the learner observes only actual deployment losses.
Use the block notation and counterfactual losses of Definition~\ref{def:block-counterfactual} in Section~\ref{sec:feedback}.
Conditional on $U$ and the past, $g_j$ precedes the draw and includes within-block reactions, but only its executed coordinate is observed.

\paragraph{Block Exponential Weights.}
Initialize $p_1=\mu$.
For each block $j$, deploy a fresh private draw $B_j\sim p_j$ for $b$ rounds, observe $Z_j=g_j(B_j)$, and set
\begin{align}
 \widehat g_j(a)&=\frac{\mathbf 1\{B_j=a\}Z_j}{p_j(a)},
 \label{eq:block-bandit-estimator}\\
 p_{j+1}(a)&=\frac{p_j(a)e^{-\eta\widehat g_j(a)}}
 {\sum_cp_j(c)e^{-\eta\widehat g_j(c)}}.
 \label{eq:block-bandit-update}
\end{align}
Retain the last configuration for the $r$ remaining rounds, giving at most $J-1$ switches.
Block draws are fresh, not lazy.
Exact probabilities stay positive, but numerical underflow remedies involving clipping or exploration require corresponding bias or exploration terms.

Fresh draws justify sampling identities against adaptive prefixes.
\begin{lemma}[Bandit selection on block losses]
\label{lem:block-bandit-selection}
For every fixed $Q$ and $\eta>0$,
\begin{equation}
 \mathbb E\sum_{j=1}^J\big[g_j(B_j)-\langle Q, g_j\rangle\big]
 \le\frac{C_Q}{\eta}+\frac{\eta KJ}{2}.
 \label{eq:block-bandit-selection}
\end{equation}
\end{lemma}
Adding the causal reaction~\eqref{eq:causal-reaction} and the remainder charge to Lemma~\ref{lem:block-bandit-selection} gives Theorem~\ref{thm:causal-bandit} in Section~\ref{sec:feedback}.

\subsubsection{Conditions Controlling Causal Reaction}
\label{sec:reaction-control}

\begin{assumption}[Total component-specific finite influence]
\label{ass:total-causal-influence}
For every fixed $U$, the complete causal response $F_t$ depends only on the coordinate windows in Assumption~\ref{ass:bounded-influence}, including all task-selection and evaluator rules, from a common initial condition.
\end{assumption}
Forgetting in a cache is insufficient if task selection remembers the first action indefinitely, because that dependence belongs in $F_t$.

\begin{lemma}[Prefix charge under finite influence]
\label{lem:finite-influence-prefix}
Under Assumption~\ref{ass:total-causal-influence}, with
$m=m_{\max}$,
\begin{equation}
 |\Gamma_b(Q)|\le\min\{m,b\}(J-1).
 \label{eq:finite-influence-prefix}
\end{equation}
\end{lemma}

\begin{corollary}[Deployment-only bounded-influence rate]
\label{cor:bandit-finite-influence}
Under Assumption~\ref{ass:total-causal-influence}, for every $\eta>0$,
\begin{equation}
    R_T^{\mathrm{pol}}(Q)\le \frac{bC_Q}{\eta}+\frac{\eta bKJ}{2}
    +\min\{m,b\}(J-1)+r.
    \label{eq:bandit-finite-influence}
\end{equation}
For $C_Q>0$, choosing $\eta=\sqrt{2C_Q/(KJ)}$ gives
\begin{align}
    R_T^{\mathrm{pol}}(Q)&\le b\sqrt{2KJC_Q}+\min\{m,b\}(J-1)+r\nonumber\\
    &\le\sqrt{2KC_QTb}+\frac{mT}{b}+b.
    \label{eq:bandit-finite-simple}
\end{align}
For fixed positive $m,K,C_Q$ and sufficiently large $T$, integer $b\asymp(m^2T/(KC_Q))^{1/3}$, truncated to $[1,T]$, gives
\begin{equation}
    R_T^{\mathrm{pol}}(Q)=O((mKC_Q)^{1/3}T^{2/3}).
    \label{eq:bandit-finite-rate}
\end{equation}
For $m=0$, use $b=1$, and for $C_Q=0$, draw once from $\mu$.
\end{corollary}

Use~\eqref{eq:bandit-finite-influence} when parameters grow with $T$.
The zero-influence rate is $\sqrt{2KTC_Q}$, and a singleton family has zero regret.
For zero KL, fresh block resampling need not have zero regret, unlike a single prior draw.
Text-only, parameter-only, and joint or cross-mode blocks use $m_Z$, $m_W$, and $m$, respectively.

Equation~\eqref{eq:bandit-finite-influence} already includes prefix effects through $b\sum_jg_j(B_j)$, so another contamination term would double-count.
The established $T^{2/3}$ rate~\citep{arora2012policy} cannot universally be improved to $\sqrt T$ under deployment-only feedback~\citep{dekel2014bandits}.
Appendix~\ref{sec:app-proofs-feedback} gives the reduction to bandits with switching costs.
This leaves optimal $m$ and prior dependence in~\eqref{eq:bandit-finite-rate} open.
Sparse probes provide stronger information.

To separate initial warm-up from persistent reaction, fix an integer $h_0\ge0$ and define
\begin{equation}
 \mathcal A_b(Q)=\mathbb E\sum_{j=2}^J\sum_aQ(a)\sum_{u=h_0+1}^b
 \Big|F_{s_j+u-1}(H_j,a^u;U)-F_{s_j+u-1}(a^{s_j+u-1};U)\Big|,
 \label{eq:residual-reaction}
\end{equation}
with an empty sum if $b\le h_0$.
Without assuming total finite influence,
\begin{equation}
 R_T^{\mathrm{pol}}(Q)\le\frac{bC_Q}{\eta}+\frac{\eta bKJ}{2}
 +\min\{h_0,b\}(J-1)+\mathcal A_b(Q)+r.
 \label{eq:residual-reaction-bound}
\end{equation}
Sublinear regret requires $\mathcal A_b(Q)=o(T)$ and appropriate $b,\eta$, neither of which the definition supplies.

A suffix-sensitivity bound can control reaction without exact forgetting.
\begin{proposition}[A suffix-sensitivity envelope]
\label{prop:suffix-sensitivity}
Suppose deterministic numbers $\beta_u\in[0,1]$ satisfy, for every admissible
prefix $h$ of length $s-1$, every $a,U$, and every relevant $u\ge1$,
\begin{equation}
 \big|F_{s+u-1}(h,a^u;U)-F_{s+u-1}(a^{s+u-1};U)\big|\le\beta_u.
\end{equation}
With $H_b=\sum_{u=1}^b\beta_u$, the block algorithm satisfies
\begin{equation}
 R_T^{\mathrm{pol}}(Q)\le\frac{bC_Q}{\eta}+\frac{\eta bKJ}{2}+H_b(J-1)+r.
 \label{eq:suffix-sensitivity}
\end{equation}
If $H=\sum_{u\ge1}\beta_u<\infty$, one may replace $H_b$ by $H$.
\end{proposition}

The envelope must hold for all protocol prefixes, not merely observed paths.
Contractive state with Lipschitz losses supplies one such envelope.

\begin{assumption}[Contractive environment state]
\label{ass:contractive-environment}
An environment state $e_t$ lies in a metric space of diameter at most $D$, with
\begin{align}
 e_{t+1}&=\Psi_t(e_t,A_t;U_t),\\
 \operatorname{dist}(\Psi_t(e,a;u),\Psi_t(e',a;u))
 &\le\rho\operatorname{dist}(e,e'),\quad 0\le\rho<1.
\end{align}
The $[0,1]$ loss depends on $e_t$, coordinate windows of lengths $m_Z+1,m_W+1$, and exogenous randomness.
For identical windows and randomness, it is $G$-Lipschitz in $e_t$.
Learner and comparator share the initial state, maps, and admissible external noise tape.
\end{assumption}

\begin{corollary}[Contractive reaction without a counterfactual oracle]
\label{cor:contractive-reaction}
Under Assumption~\ref{ass:contractive-environment}, let $m=m_{\max}$ and
\begin{align}
 H_b&=\min\{m,b\}+\sum_{u=m+1}^b\min\{1,GD\rho^{u-1}\},
 \label{eq:finite-contraction-charge}\\
 H&=m+\frac{GD\rho^m}{1-\rho}.
 \label{eq:contraction-charge}
\end{align}
Then \eqref{eq:suffix-sensitivity} holds, with $H_b\le\min\{b,H\}$.
For fixed positive $H,K,C_Q$, the corresponding tuned order is
$O((HKC_Q)^{1/3}T^{2/3})$.
\end{corollary}

For $\rho=0,m=0$, set $\rho^0=1$: initial state mismatch can affect loss before contraction.
For $H=0$, use $b=1$, not the cube-root formula.

For example, $e_{t+1}=\rho e_t+(1-\rho)v(A_t)$ on a bounded convex set models a deployment-dependent task mixture with Lipschitz expected loss.
Its discrete sampled-task losses, however, need not be pathwise Lipschitz.
Summable reaction replaces finite influence, relating to online control under mixing~\citep{evendar2009online}, although Markov mixing alone does not imply pathwise contraction.

\subsubsection{Identification from Logged Deployment Data}
\label{sec:logged-feedback}

An ordinary one-step importance-weighted outcome,
\begin{equation}
 \frac{\mathbf 1\{A_t=a\}Z_t}{\beta_t(a\mid H_{t-1})},
\end{equation}
intervenes only on the current action, so it need not estimate own-trajectory loss~\eqref{eq:common-diagonal-loss}.
For $F_2(a_1,a_2)=\mathbf 1\{a_1\ne a_2\}$ after $a_1=1$, choosing $a_2=0$ loses one, whereas $d_2(0)=0$.
Sequential evaluation requires sequential intervention~\citep{jiang2016doubly}.

\begin{lemma}[Window importance sampling in a sufficient model]
\label{lem:window-importance}
Consider a fixed deterministic response table with total joint influence horizon $m$ (for coordinate influence horizons, $m=m_{\max}$ is sufficient),
$t\ge m+1$, and a behavior policy with known positive propensities
$\beta_s(a\mid H_{s-1})\ge\epsilon>0$. Let the logged outcome $Z_t\in[0,1]$
equal the realized loss, or be conditionally unbiased for it given the full
history through the current action. Put
\begin{align}
 W_t^a&=\prod_{s=t-m}^t\frac{\mathbf 1\{A_s=a\}}{\beta_s(a\mid H_{s-1})},\\
 \widetilde d_t(a)&=W_t^a Z_t.
\end{align}
Let $\mathcal H_{t-m-1}$ denote the logged history before the window.
Then
\begin{align}
 \mathbb E[\widetilde d_t(a)\mid\mathcal H_{t-m-1}]&=d_t(a),\\
 \mathbb E[\widetilde d_t(a)^2\mid\mathcal H_{t-m-1}]&\le\epsilon^{-(m+1)}.
 \label{eq:window-importance}
\end{align}
\end{lemma}

Uniform logging gives a second-moment bound $K^{m+1}$ for this estimator, not a minimax lower bound.
Stochastic systems also require consistency, sequential ignorability, and window identification of the full own trajectory.
Overlapping windows need not be unbiased conditional on preceding estimates, and lazy logging violates Assumption~\ref{ass:independent-diagonal-probe}.
These issues require independent episodes, valid sample splitting, or delayed/blockwise analysis.
Without intervention support, neither models nor doubly robust estimation alone identify diagonal values.

\subsubsection{Transferring Bounds to the Fixed Joint Comparator}
\label{sec:feedback-comparators}

For a deterministic response tree, let $a_e^\star$ minimize the fixed-candidate loss $D_T$ from Section~\ref{sec:protocol} over $\mathcal C_e$, attaining $V_{T,\mathcal C}^e$.
Adding family restriction to its policy regret gives
\begin{equation}
    \mathbb E L_T^{\mathrm{dep}}-V_{T,\mathcal C}^{ZW}
    =R_T^{\mathrm{pol}}(a_e^\star)
      +V_{T,\mathcal C}^e-V_{T,\mathcal C}^{ZW}.
    \label{eq:feedback-joint-comparator}
\end{equation}
This groups selection and state in~\eqref{eq:common-accounting}.
Distributional comparators additionally incur their approximation gap to the mode optimum.
The identity does not certify candidate discovery or reachability.

Conditioning on the exogenous tree fixes its best candidate before probing, making it admissible in Theorem~\ref{thm:sparse-diagonal} under Assumption~\ref{ass:independent-diagonal-probe}.
Theorem~\ref{thm:causal-bandit} likewise permits $Q(U)$ fixed conditional on $U$, and arbitrary-$\eta$ bounds average conditional complexity and reaction over $U$.
Unobserved realized complexity may prevent implementable tuning, but uniform priors give $\log K$ for every point comparator without requiring knowledge of its identity.

Proposition~\ref{prop:query-adaptive-probes} permits fixed $Q$ or a best expected-loss candidate, not a query-dependent comparator correlated with probe noise.
Conditioning on the full transcript establishes deployment marginals, not conditional unbiasedness of realized estimates, and the optima in~\eqref{eq:expected-versus-hindsight-optimum} remain distinct.
Sparse probing therefore needs a valid counterfactual evaluator, and sublinear deployment-only regret needs controlled reaction~\eqref{eq:causal-reaction}.
Implementation must also account for numerical stabilization and evaluation costs.

%% file: appendices/appendix_proofs.tex
\section{Proofs}
\label{sec:app-proofs}

This appendix proves the results stated in the main text and Appendix~\ref{sec:app-theory}.
The proofs follow the same topic order: configurations and loss decomposition, selection and family restriction, state effects, and feedback requirements.
The final subsection proves the cost result in Appendix~\ref{sec:deployment}.

\subsection{Configurations, Comparators, and Loss Decomposition}
\label{sec:app-proofs-accounting}

\begin{proof}[Proof of Proposition~\ref{prop:common-accounting}]
\phantomsection\label{proof:common-accounting}
Add and subtract $L_T^{\mathrm{diag}}$ and $V_{T,\mathcal C}^{e}$.
Family inclusion implies $V_{T,\mathcal C}^{e}\geq V_{T,\mathcal C}^{ZW}$.
A changing selection can outperform every fixed candidate, and a mixed history can improve or worsen its diagonal loss.
Thus, neither remaining term has a fixed sign.
\end{proof}

\subsection{Selection and Family Restriction}
\label{sec:app-proofs-selection}

\begin{proof}[Proof of~\eqref{eq:finite-common-bound}]
\phantomsection\label{proof:finite-common-bound}
Apply Theorem~\ref{thm:prior-complexity} on $\mathcal C_e$ with $L_t=d_t$, prior $\mu_e$, and $Q=\delta_{a_e^\star}$.
Its KL term is $\log(1/\mu_e(a_e^\star))$, and its regret is $\mathbb E L_T^{\mathrm{diag}}-V_{T,\mathcal C}^e$.
Statelessness makes the state term in~\eqref{eq:common-accounting} zero, so adding family restriction gives~\eqref{eq:finite-common-bound}.
\end{proof}

\begin{proof}[Proof of Theorem~\ref{thm:prior-complexity}]
\phantomsection\label{proof:prior-complexity}
Hoeffding's lemma~\citep{hoeffding1963probability} and telescoping give
\begin{equation}
    \log Z_{T+1}
    \leq
    -\eta\sum_{t=1}^T\overline L_t
    +\frac{\eta^2T}{8}.
    \label{eq:kl-potential-upper}
\end{equation}
For $q=\mathrm dQ/\mathrm d\mu$, Jensen's inequality gives
\begin{equation}
    \log Z_{T+1}
    \geq
    -\eta\int_{\mathcal A}\sum_{t=1}^T L_t(a)\,\mathrm{d}Q(a)
    -\operatorname{KL}(Q\,\|\,\mu).
    \label{eq:kl-potential-lower}
\end{equation}
Combining the bounds gives~\eqref{eq:general-kl-regret}.
Optimizing this bound for positive KL gives~\eqref{eq:general-kl-regret-optimized}.
If $\operatorname{KL}(Q\,\|\,\mu)=0$, then $Q=\mu$, and the $\eta=0$ extension $p_t=\mu$ has zero distributional pseudo-regret.
\end{proof}

\begin{proof}[Proof of Corollary~\ref{cor:product-complexity}]
\phantomsection\label{proof:product-complexity}
Apply Radon--Nikodym factorization and the KL chain rule through $Q_Z\otimes Q_W$.
\end{proof}

\begin{proof}[Proof of Corollary~\ref{cor:prior-atom-regret}]
\phantomsection\label{proof:prior-atom-regret}
Apply Theorem~\ref{thm:prior-complexity} to $Q=\delta_{(z^\star,w^\star)}$.
For this comparator,~\eqref{eq:joint-kl-decomposition} makes its KL the sum of the negative log atom masses.
Zero complexity means the prior is concentrated on the reference.
\end{proof}

\begin{proof}[Proof of Corollary~\ref{cor:local-point-regret}]
\phantomsection\label{proof:local-point-regret}
Use $Q=\pi(\cdot\mid\mathbb B_Z)\otimes\rho(\cdot\mid\mathbb B_W)$, whose KL is $C_{\delta_Z,\delta_W}(a^\star)$.
Equation~\eqref{eq:coordinate-lipschitz-loss} adds at most $G_Z\delta_Z+G_W\delta_W$ per round.
Applying Theorem~\ref{thm:prior-complexity} and optimizing then gives the stated bound.
\end{proof}

\begin{proof}[Proof of Proposition~\ref{prop:selection-restriction}]
\phantomsection\label{proof:selection-restriction}
Add and subtract the family values and the $Q_e$-average cumulative loss defined in Section~\ref{sec:mode-restriction}.
The definitions of $\Delta_e^{\mathrm{mode}}$, $\Delta^{\mathrm{joint}}$, and $\mathsf{Apx}_T^e(Q_e)$ identify the resulting gaps, and Theorem~\ref{thm:prior-complexity} bounds the selection term on $\mathcal A_e$.
\end{proof}

\begin{proof}[Proof of Corollary~\ref{cor:mixture-mode-selection}]
\phantomsection\label{proof:mixture-mode-selection}
Since $\mu_{\mathrm{mix}}\geq\alpha_e\mu_e$ as measures, $Q_e\ll\mu_{\mathrm{mix}}$ and integration gives
\begin{equation}
    \operatorname{KL}(Q_e\,\|\,\mu_{\mathrm{mix}})
    \leq
    \operatorname{KL}(Q_e\,\|\,\mu_e)+\log(1/\alpha_e).
\end{equation}
Apply Theorem~\ref{thm:prior-complexity} and~\eqref{eq:main-selection-restriction}.
\end{proof}

\begin{proof}[Proof of Proposition~\ref{prop:coordinate-additive-equivalence}]
\phantomsection\label{proof:coordinate-additive-equivalence}
The cumulative loss and product prior yield the following factorization of~\eqref{eq:coordinate-ew-target}:
\begin{equation}
    q_t(z,w)=q_t^Z(z)q_t^W(w).
\end{equation}
The first conditional draw therefore has law $q_t^Z$ independently of the preceding $w$, and the second has law $q_t^W$ independently of the new $z$.
The resulting joint law is $q_t$, including the common initialization at $t=1$.
Stateless losses then give equal expected cumulative loss.
\end{proof}

\begin{proof}[Proof of Proposition~\ref{prop:coordinate-fixed-rate-separation}]
\phantomsection\label{proof:coordinate-fixed-rate-separation}
The best fixed configuration is $(1,1)$, with zero loss.
For $n=t-1\geq1$, the joint normalizer includes its unit unnormalized weight, giving
\begin{equation}
    \mathbb E f(A_t^{\mathrm{joint}})
    \leq \Delta e^{-\eta n\Delta}
    +2(\Delta+b)e^{-\eta n(\Delta+b)}.
\end{equation}
The uniform initialization has expected loss $(3\Delta+2b)/4$.
Summing the geometric series proves the first bound, with its final inequality following from $e^x-1\geq x$.

The coordinatewise selector starts at $(0,0)$ with probability $1/4$.
Conditional on remaining there, each coordinate draw at round $t=n+1$ retains zero with probability $(1+e^{-\eta nb})^{-1}$.
The event $E_T$ of starting at $(0,0)$ and retaining zero at every subsequent coordinate draw therefore has probability:
\begin{align}
    \Pr(E_T)
    &=\frac14\prod_{n=1}^{T-1}(1+e^{-\eta nb})^{-2}\nonumber\\
    &\geq\frac14\exp\!\left(-2\sum_{n=1}^{T-1}e^{-\eta nb}\right)\nonumber\\
    &\geq\frac14\exp\!\left(-\frac{2}{e^{\eta b}-1}\right).
\end{align}
Here $\log(1+x)\leq x$ gives the first inequality.
Loss equals $\Delta$ on all $T$ rounds of $E_T$ and is nonnegative otherwise, proving~\eqref{eq:coordinate-fixed-rate-separation}.
\end{proof}

\begin{proof}[Proof of Corollary~\ref{cor:coordinate-convergent-separation}]
\phantomsection\label{proof:coordinate-convergent-separation}
Writing $n=t-1$, the three suboptimal unnormalized joint weights are $(4n)^{-\gamma}$ and two copies of $(4n)^{-(\gamma+1)}$.
Bound the normalizer below by one and sum the resulting expected losses, using $\sum_{n\geq1}n^{-p}\leq1+1/(p-1)$ for $p>1$.
These weights also bound the probabilities of nonoptimal joint draws by a summable sequence, so the first Borel--Cantelli lemma gives eventual optimal selection.

For the coordinatewise selector, the event $E_t$ from the proof of Proposition~\ref{prop:coordinate-fixed-rate-separation} satisfies the following:
\begin{align}
    \Pr(E_t)
    &=\frac14\prod_{k=1}^{n}\left(1+\frac1{4k}\right)^{-2}\nonumber\\
    &\geq\frac14\exp\!\left(-\frac12\sum_{k=1}^{n}\frac1k\right)\nonumber\\
    &\geq\frac{e^{-1/2}}{4\sqrt n}.
\end{align}
Multiply by $\Delta$ and sum over $t=2,\ldots,T$ to obtain~\eqref{eq:coordinate-convergent-separation}.

To establish eventual optimal selection, put $s_n=(4n)^{-(\gamma+1)}$.
From any starting pair, the next complete sweep ends at $(1,1)$ with conditional probability at least
\begin{equation}
    \frac1{(4n+1)(1+s_n)}\geq\frac1{10n}.
\end{equation}
The product of the corresponding avoidance bounds over any infinite tail is zero, so $(1,1)$ is visited infinitely often almost surely.
A sweep starting at $(1,1)$ departs from it with probability at most $2s_n$.
These departure probabilities are summable, so Borel--Cantelli gives only finitely many departures.
Infinitely many visits and finitely many departures imply that the coordinatewise selector eventually remains at $(1,1)$.
\end{proof}

\subsection{State Effects and Policy Regret}
\label{sec:app-proofs-state}

\begin{proof}[Proof of Proposition~\ref{prop:external-policy-separation}]
\phantomsection\label{proof:external-policy-separation}
Take text configurations $\mathcal Z=\{0,1\}$ with initial, good, and bad states, zero round-one loss, and
\begin{align}
    z_1=0 &\Rightarrow s_2=s_{\mathrm{good}},\\
    z_1=1 &\Rightarrow s_2=s_{\mathrm{bad}}.
\end{align}
The good and bad states are absorbing, with losses for $t\geq2$ given by
\begin{equation}
    \ell_t(s,z)
    =
    \begin{cases}
        0, & s=s_{\mathrm{good}},\\
        1, & s=s_{\mathrm{bad}},
    \end{cases}
    \label{eq:absorbing-state-loss}
\end{equation}
independent of $z$.
Choosing $z_1=1$ loses $T-1$, as does either fixed action on that bad-state path, giving external regret:
\begin{equation}
    \sum_{t=1}^T\ell_t(s_t,z_t)
    -\min_{z\in\{0,1\}}
     \sum_{t=1}^T\ell_t(s_t,z)
    =(T-1)-(T-1)=0.
\end{equation}
The constant counterfactual $z=0$ enters the good state and loses zero, so
\begin{equation}
    R_T^{\mathrm{pol}}(0)=(T-1)-0=T-1.
\end{equation}
This completes the proof.
\end{proof}

\begin{proof}[Proof of Proposition~\ref{prop:irreversible-state}]
\phantomsection\label{proof:irreversible-state}
Fix an optimizer $\mathsf{Alg}$ and text-only environments $E_\xi$, $\xi\in\{0,1\}$, sharing their initial state and pre-round history.
Their indistinguishability makes $p=\Pr_{\mathsf{Alg}}(z_1=1)$ common to both.
Reuse the absorbing states and losses from the proof of Proposition~\ref{prop:external-policy-separation}, with first transition
\begin{equation}
    s_2=
    \begin{cases}
        s_{\mathrm{good}}, & z_1=\xi,\\
        s_{\mathrm{bad}}, & z_1\neq\xi.
    \end{cases}
\end{equation}
By~\eqref{eq:absorbing-state-loss}, the matching fixed action $z=\xi$ loses zero, and the optimizer's regrets are
\begin{equation}
    R_T^{\mathrm{pol}}(0;E_0)
    =p(T-1).
\end{equation}
\begin{equation}
    R_T^{\mathrm{pol}}(1;E_1)
    =(1-p)(T-1).
\end{equation}

Since $\max\{p,1-p\}\geq1/2$,
\begin{equation}
    \max\left\{
        R_T^{\mathrm{pol}}(0;E_0),
        R_T^{\mathrm{pol}}(1;E_1)
    \right\}
    \geq \frac{T-1}{2}.
\end{equation}
The hard environment is chosen after fixing the optimizer but before realizing its random seed.
Revealing $\xi$ after the irreversible transition cannot help.
\end{proof}

\begin{proof}[Proof of Lemma~\ref{lem:lazy-marginals}]
\phantomsection\label{proof:lazy-marginals}
For measurable $D\subseteq\mathcal A$, the conditional transition kernel is
\begin{equation}
    K_t(a,D)
    =
    e^{-\eta d_t(a)}\mathbf 1_D(a)
    +\left(1-e^{-\eta d_t(a)}\right)q_{t+1}(D).
    \label{eq:lazy-transition-kernel}
\end{equation}
Integrating against $q_t$ gives $\zeta_t q_{t+1}(D)+(1-\zeta_t)q_{t+1}(D)=q_{t+1}(D)$, proving the marginals by induction.
Use $\mathbb E[Y_t]=\int(1-e^{-\eta d_t})\,\mathrm dq_t$, $1-e^{-x}\leq x$, and $d_t\leq1$, then sum.
\end{proof}

\begin{proof}[Proof of Proposition~\ref{prop:matched-marginals}]
\phantomsection\label{proof:matched-marginals}
Both actions have law $q_t$ by Lemma~\ref{lem:lazy-marginals}, hence expected diagonal loss $\int d_t\,\mathrm dq_t$.
The restriction terms also agree, so subtracting~\eqref{eq:common-accounting} in expectation proves the identity.
\end{proof}

\begin{proof}[Proof of Lemma~\ref{lem:contamination}]
\phantomsection\label{proof:contamination}
A coordinate switch at $s$ can contaminate only $s,\ldots,\min\{T,s+m_e-1\}$, at most $m_e$ rounds.
Simultaneous intervals share a start, so their union has length at most the larger horizon.
Deployed and diagonal losses coincide outside these intervals and differ by at most one inside, proving~\eqref{eq:pathwise-switch-contamination} by the triangle inequality and a union bound.
Without a refresh, $A_s=A_{s-1}$, so $\chi_s\leq m_{\max}Y_{s-1}$ gives~\eqref{eq:refresh-contamination}.
\end{proof}

\begin{proof}[Proof of Theorem~\ref{thm:lazy-policy}]
\phantomsection\label{proof:lazy-policy}
Lemma~\ref{lem:lazy-marginals} and Theorem~\ref{thm:prior-complexity} applied to~\eqref{eq:bounded-diagonal-loss} give
\begin{equation}
    \sum_{t=1}^T\mathbb E[d_t(A_t)]
    -\int_{\mathcal A}\sum_{t=1}^T d_t(a)\,\mathrm{d}Q(a)
    \leq
    \frac{C_Q}{\eta}+\frac{\eta T}{8}.
\end{equation}
Adding the signed state term as in~\eqref{eq:common-accounting}, then applying~\eqref{eq:pathwise-switch-contamination}, \eqref{eq:refresh-contamination}, and~\eqref{eq:lazy-refresh-bound}, gives the sharper chain
\begin{align}
    R_T^{\mathrm{pol}}(Q)
    &\leq
    \frac{C_Q}{\eta}
    +\frac{\eta T}{8}
    +\mathbb E\left[\sum_{s=2}^T\chi_s\right]\nonumber\\
    &\leq
    \frac{C_Q}{\eta}
    +\frac{\eta T}{8}
    +m_{\max}\mathbb E\left[\sum_{t=1}^{T-1}Y_t\right]\nonumber\\
    &\leq
    \frac{C_Q}{\eta}
    +\frac{\eta T}{8}
    +m_{\max}(T-1)(1-e^{-\eta})\nonumber\\
    &\leq
    \frac{C_Q}{\eta}
    +\frac{\eta T}{8}
    +m_{\max}\eta(T-1).
\end{align}
Optimizing gives~\eqref{eq:lazy-learning-rate} and~\eqref{eq:lazy-policy-regret-optimized} for $C_Q>0$.
At $C_Q=0$, equality in KL gives $Q=\mu$, and a fixed prior draw incurs $\sum_t d_t(A_1)$, matching the comparator in expectation.
\end{proof}

\begin{proof}[Proof of Corollary~\ref{cor:lazy-local-point-regret}]
\phantomsection\label{proof:lazy-local-point-regret}
The conditioned prior in the proof of Corollary~\ref{cor:local-point-regret} has KL $C_{\delta_Z,\delta_W}(a^\star)$ and diagonal loss at most $T(G_Z\delta_Z+G_W\delta_W)$ above $a^\star$ by Lipschitzness.
Apply Theorem~\ref{thm:lazy-policy} and optimize.
\end{proof}

\begin{proof}[Proof of Proposition~\ref{prop:mode-lazy-rates}]
\phantomsection\label{proof:mode-lazy-rates}
Constant parameter history contributes no contamination in Lemma~\ref{lem:contamination}, so Theorem~\ref{thm:lazy-policy} applies on $\mathcal Z$ with prior $\pi$ and horizon $m_Z$.
The parameter-only proof is symmetric.
\end{proof}

\begin{proof}[Proof of Theorem~\ref{thm:episodic-policy-regret}]
\phantomsection\label{proof:episodic-policy-regret}
Apply Theorem~\ref{thm:prior-complexity} on $\mathcal P$ to losses $J_k/H\in[0,1]$ with learning rate $\eta H$, which gives the weights in~\eqref{eq:episodic-measurable-weights}.
Multiplying its regret bound by $H$ proves~\eqref{eq:episodic-kl-regret}, and optimizing over $\eta$ gives the stated rate.
\end{proof}

\subsection{Feedback Requirements and Extensions}
\label{sec:app-proofs-feedback}

\begin{proof}[Proof of Corollary~\ref{cor:approximate-diagonal-evaluation}]
\phantomsection\label{proof:approximate-diagonal-evaluation}
Apply Lemma~\ref{lem:lazy-marginals} and Theorem~\ref{thm:prior-complexity} with the fixed estimates $\widehat d_t$ as losses.
Bounded estimates give an expected refresh count at most $\eta(T-1)$, so Lemma~\ref{lem:contamination} bounds the expected absolute state term by $m_{\max}\eta(T-1)$.
For every $t$,
\begin{equation}
    \mathbb E[d_t(A_t)]-\int d_t\,\mathrm dQ
    \leq
    \mathbb E[\widehat d_t(A_t)]-\int \widehat d_t\,\mathrm dQ
    +2\epsilon_t.
\end{equation}
Sum and apply the KL bound to $\widehat d_{1:T}$.
\end{proof}

\begin{proof}[Proof of Corollary~\ref{cor:diagonal-sampling-cost}]
\phantomsection\label{proof:diagonal-sampling-cost}
Hoeffding's inequality bounds each failure probability by
$2\exp(-2n\epsilon_n^2)$, so a union bound over $KT$ pairs gives $\delta$.
For any estimate table in the success event, apply
Corollary~\ref{cor:approximate-diagonal-evaluation} with $\epsilon_t=\epsilon_n$.
Otherwise, true policy regret is at most $T$, since both cumulative losses lie
in $[0,T]$.  Taking expectations gives~\eqref{eq:sampled-diagonal-regret}.
\end{proof}

\begin{proof}[Proof of Corollary~\ref{cor:unbiased-diagonal}]
\phantomsection\label{proof:unbiased-diagonal}
For a fixed exogenous estimate table, Lemma~\ref{lem:lazy-marginals} and Theorem~\ref{thm:prior-complexity}
apply to $\widehat d_t$.  Since $q_t$ depends only on $\widehat d_{1:t-1}$,
conditional unbiasedness gives
\begin{equation}
    \mathbb E\!\left[\sum_a(q_t(a)-Q(a))\widehat d_t(a)\right]
    =\mathbb E\!\left[\sum_a(q_t(a)-Q(a))d_t(a)\right].
\end{equation}
Taking expectations of the estimated-loss bound controls true diagonal
selection by $C_Q/\eta+\eta T/8$.  Bounded estimates give at most
$\eta(T-1)$ expected refreshes, allowing Lemma~\ref{lem:contamination} to bound the state term.
\end{proof}

\begin{proof}[Proof of Proposition~\ref{prop:adaptive-lazy-failure}]
\phantomsection\label{proof:adaptive-lazy-failure}
Take $\mathcal C=\{0,1\}$, $q_1=(1/2,1/2)$, and
$\ell_1\equiv0$. Lazy sampling retains $A_2=A_1$ surely, while
$q_2=(1/2,1/2)$. After observing $A_1$, but before round two, let
\begin{equation}
 \ell_2(a)=\mathbf 1\{a=A_1\}.
\end{equation}
Then $\ell_2(A_2)=1$, whereas $\langle q_2, \ell_2\rangle=1/2$.
Conditional on the realized past, the actual law is
$\mathcal L(A_2\mid A_1)=\delta_{A_1}$, not $q_2$.
\end{proof}

\begin{proof}[Proof of Proposition~\ref{prop:external-switch-tradeoff}]
\phantomsection\label{proof:external-switch-tradeoff}
The learner's loss is $T-1-S_T$. Since $\ell_t(0)+\ell_t(1)=1$,
the smaller fixed-action cumulative loss on this realized sequence is at most
$(T-1)/2$. Subtract and take expectations.
\end{proof}

\begin{proof}[Proof of Lemma~\ref{lem:nonnegative-ew}]
\phantomsection\label{proof:nonnegative-ew}
Let $Z_t=\sum_a\mu(a)e^{-\eta\sum_{s<t}x_s(a)}$.
For $u\ge0$, $e^{-u}\le1-u+u^2/2$. Therefore
\begin{equation}
 \log\frac{Z_{t+1}}{Z_t}
 \le -\eta\langle q_t, x_t\rangle
       +\frac{\eta^2}{2}\langle q_t, x_t^2\rangle.
\end{equation}
Here $1-u+u^2/2>0$, so using $\log(1+v)\le v$ is valid.
The Jensen lower bound~\eqref{eq:kl-potential-lower} applies with $x_t$ replacing $L_t$ because that step does not require losses bounded by one.
Sum the upper bound, use $Z_1=1$, combine with the lower bound, and divide by $\eta$.
\end{proof}

\begin{proof}[Proof of Lemma~\ref{lem:probe-moments}]
\phantomsection\label{proof:probe-moments}
For the mean, the probability $r_t(a)$ cancels the denominator in
\eqref{eq:sparse-diagonal-estimator}. For the second moment it cancels only
one of its two squared factors. Sum with weights $q_t(a)$ and use
$Z_t^2\le1$.
\end{proof}

\begin{proof}[Proof of Lemma~\ref{lem:sparse-lazy-marginals}]
\phantomsection\label{proof:sparse-lazy-marginals}
Conditioning fixes all estimates and leaves deployment coins independent.
The kernel in~\eqref{eq:lazy-transition-kernel}, with $\widehat d_t$ replacing $d_t$, therefore preserves $q_{t+1}$ by the calculation in the proof of Lemma~\ref{lem:lazy-marginals}.
That calculation needs only nonnegative finite losses, which keep retention probabilities in $[0,1]$.
Induction gives the marginals, and
\begin{align}
 \mathbb E[R_t\mid\text{auxiliary transcript}]
 &=\sum_aq_t(a)(1-e^{-\eta\widehat d_t(a)})\nonumber\\
 &\le\eta\langle q_t, \widehat d_t\rangle.
\end{align}
Average using Lemma~\ref{lem:probe-moments}: although estimates can exceed one, their conditional means lie in $[0,1]$.
\end{proof}

\begin{proof}[Proof of Theorem~\ref{thm:sparse-diagonal}]
\phantomsection\label{proof:sparse-diagonal}
Apply Lemma~\ref{lem:nonnegative-ew} to $x_t=\widehat d_t$.
Since $q_t$ and fixed $Q$ are measurable before the probe, conditional unbiasedness gives the following:
\begin{equation}
 \mathbb E\sum_t\langle q_t-Q, d_t\rangle
 \le\frac{C_Q}{\eta}+\frac{\eta}{2}\sum_t\mathbb E V_t.
\end{equation}
Lemma~\ref{lem:sparse-lazy-marginals} gives $\mathbb E[d_t(A_t)\mid\text{auxiliary transcript}]=\langle q_t,d_t\rangle$, identifying expected true diagonal selection regret.

On this response tree, Lemma~\ref{lem:contamination} gives, with $S_T=\sum_{t=2}^T\mathbf 1\{A_t\neq A_{t-1}\}$,
\begin{equation}
 \left|\sum_tF_t(A_{1:t})-\sum_td_t(A_t)\right|
 \le m S_T\le m\sum_{t<T}R_t.
\end{equation}
Add this state charge once and apply \eqref{eq:sparse-refresh}. Finally use
$V_t\le K$, $\langle q_t, d_t\rangle\le1$, and optimize $C_Q/\eta+\eta B$.
\end{proof}

\begin{proof}[Proof of Proposition~\ref{prop:query-adaptive-probes}]
\phantomsection\label{proof:query-adaptive-probes}
Conditioning on the autonomous auxiliary process fixes all estimates without conditioning on deployment coins, so Lemma~\ref{lem:sparse-lazy-marginals} and the pathwise contamination bound still apply.
For the selection term, condition only on $\mathcal G_{t-1}$ before each probe.
Both $q_t$ and fixed $Q$ are measurable there, which justifies the unbiasedness step in the proof of Theorem~\ref{thm:sparse-diagonal}.
The second-moment and refresh calculations are unchanged.
\end{proof}

\begin{proof}[Proof of Corollary~\ref{cor:multiple-probes}]
\phantomsection\label{proof:multiple-probes}
For a single probe vector $X_t$, conditional independence gives
\begin{align}
 \mathbb E[\langle q_t, \widehat d_t^2\rangle\mid\mathcal G_{t-1}]
 &=(1-1/k)\langle q_t, d_t^2\rangle
 +(1/k)\mathbb E[\langle q_t, X_t^2\rangle\mid\mathcal G_{t-1}]\nonumber\\
 &\le1-1/k+K/k.
\end{align}
The nonnegative unbiased average satisfies Theorem~\ref{thm:sparse-diagonal} with this second moment.
\end{proof}

\begin{proof}[Proof of Corollary~\ref{cor:query-budget}]
\phantomsection\label{proof:query-budget}
Condition on the pre-gate context. Multiplying the inclusion probability
$pq_t(a)$ by the estimator cancels the denominator, whereas one factor
remains for its square. Thus the mean is $d_t(a)$ and the moment is at most $K/p$.
Apply Theorem~\ref{thm:sparse-diagonal} and optimize.
\end{proof}

\begin{proof}[Proof of Proposition~\ref{prop:biased-probes}]
\phantomsection\label{proof:biased-probes}
The second-order argument bounds selection relative to $\bar d_t$.
The difference from true diagonal selection is
\begin{equation}
 \langle q_t-Q, d_t-\bar d_t\rangle
 \le\|q_t-Q\|_1\epsilon_t\le2\epsilon_t.
\end{equation}
The lazy refresh calculation gives
$\mathbb E R_t\le\eta\mathbb E\langle q_t, \bar d_t\rangle$, not the same expression
with $d_t$. Add the state charge and use $\bar d_t\in[0,1]$.
\end{proof}

\begin{proof}[Proof of Lemma~\ref{lem:block-bandit-selection}]
\phantomsection\label{proof:block-bandit-selection}
Condition on $U$ and the learner's history before the fresh block draw.
Then $g_j,p_j,Q$ are fixed, and $B_j\sim p_j$. Hence
\begin{align}
 \mathbb E[\widehat g_j(a)\mid U,\text{past}]&=g_j(a),\\
 \mathbb E[\langle p_j, \widehat g_j^2\rangle\mid U,\text{past}]
 &=\sum_a g_j(a)^2\le K.
\end{align}
Apply Lemma~\ref{lem:nonnegative-ew}, use $\mathbb E[g_j(B_j)\mid U,\text{past}]=\langle p_j,g_j\rangle$, and average over the past and $U$.
\end{proof}

\begin{proof}[Proof of Theorem~\ref{thm:causal-bandit}]
\phantomsection\label{proof:causal-bandit}
Actual loss on complete blocks is exactly $b\sum_jg_j(B_j)$. Adding and
subtracting $b\sum_j\langle Q, g_j\rangle$ gives
\begin{equation}
 \mathbb E\sum_{t=1}^{bJ}F_t(A_{1:t};U)
    -\mathbb E\sum_aQ(a)\sum_{t=1}^{bJ}d_t(a;U)=b\mathbb E\sum_{j=1}^J
        [g_j(B_j)-\langle Q, g_j\rangle]+\Gamma_b(Q).
\end{equation}
Apply Lemma~\ref{lem:block-bandit-selection}. The difference on the remaining
$r$ rounds is at most $r$, since both losses lie in $[0,1]$.
\end{proof}

\begin{proof}[Proof of Lemma~\ref{lem:finite-influence-prefix}]
\phantomsection\label{proof:finite-influence-prefix}
The first block's common empty prefix gives zero discrepancy.
In later blocks, $u\ge m+1$ repetitions make both coordinate windows constant, equating the losses in~\eqref{eq:causal-reaction}.
At most $\min\{m,b\}$ earlier differences remain, each bounded by one.
Summing their absolute values and averaging over $Q$ and randomness gives the bound.
\end{proof}

\begin{proof}[Proof of Corollary~\ref{cor:bandit-finite-influence}]
\phantomsection\label{proof:bandit-finite-influence}
Combine Theorem~\ref{thm:causal-bandit} with Lemma~\ref{lem:finite-influence-prefix},
optimize the first two terms, and use $J\le T/b$ and $r<b$.
\end{proof}

\begin{proof}[Deployment-only lower bound]
\phantomsection\label{proof:deployment-only-lower-bound}
The memory-one subclass
\begin{align}
 F_t(a_{1:t})&=\frac{1}{2}\big(\ell_t(a_t)+\mathbf 1\{a_t\ne a_{t-1}\}\big),\quad t\ge2,\\
 F_1(a_1)&=\frac{1}{2}\ell_1(a_1),
\end{align}
contains adversarial bandits with switching cost, up to a fixed scaling.
The lower bound of~\citet{dekel2014bandits} therefore gives a $\widetilde\Omega(T^{2/3})$ obstruction for fixed action count, without establishing optimal $m$ or prior dependence in~\eqref{eq:bandit-finite-rate}.
\end{proof}

\begin{proof}[Proof of~\eqref{eq:residual-reaction-bound}]
\phantomsection\label{proof:residual-reaction-bound}
The first block has the same empty prefix under deployment and the comparator, so its reaction is zero.
In the remaining blocks, split the first $\min\{h_0,b\}$ rounds from the remainder in~\eqref{eq:causal-reaction}.
Bound each early absolute discrepancy by one and the remainder by~\eqref{eq:residual-reaction}, then apply Theorem~\ref{thm:causal-bandit}.
\end{proof}

\begin{proof}[Proof of Proposition~\ref{prop:suffix-sensitivity}]
\phantomsection\label{proof:suffix-sensitivity}
The first block still has identical prefixes. In every other block, sum the
$\beta_u$ bounds in \eqref{eq:causal-reaction}, then apply
Theorem~\ref{thm:causal-bandit}.
\end{proof}

\begin{proof}[Proof of Corollary~\ref{cor:contractive-reaction}]
\phantomsection\label{proof:contractive-reaction}
After $u-1$ common actions, contraction reduces any initial distance at most $D$ to $D\rho^{u-1}$.
For $u>m$, coordinate windows agree, bounding loss discrepancy by $\min\{1,GD\rho^{u-1}\}$.
For the remaining rounds $u\le m$, use the bound of one.
Apply Proposition~\ref{prop:suffix-sensitivity} and sum the geometric tail, using that each summand is also at most one, to obtain~\eqref{eq:contraction-charge}.
Optimize as in~\eqref{eq:bandit-finite-simple} with $H$ replacing $m$.
\end{proof}

\begin{proof}[Proof of Lemma~\ref{lem:window-importance}]
\phantomsection\label{proof:window-importance}
The product of propensity ratios implements the intervention fixing every
action in the window to $a$. On that intervention, total finite memory
identifies current loss with $d_t(a)$ independently of the earlier prefix.
Sequential conditioning gives the mean identity and $\mathbb E W_t^a=1$.
Since $0\le W_t^a\le\epsilon^{-(m+1)}$ and $0\le Z_t\le1$,
\begin{equation}
 (W_t^aZ_t)^2\le\epsilon^{-(m+1)}W_t^a.
\end{equation}
Take the same conditional expectation.
\end{proof}

\subsection{Adaptation Costs and Amortization}
\label{sec:app-proofs-costs}

\begin{proof}[Proof of Proposition~\ref{prop:deployment-break-even}]
\phantomsection\label{proof:deployment-break-even}
Subtract~\eqref{eq:deployment-cost-text} from~\eqref{eq:deployment-cost-parameter} and divide by the per-use difference, which is positive by~\eqref{eq:deployment-per-use-advantage}.
The same comparison immediately gives the nonpositive-threshold case.
\end{proof}

%% file: appendices/appendix_experiments.tex
\section{Experimental Details and Numerical Results}
\label{sec:app-experiments}

This appendix provides the experimental specifications and numerical results supporting Section~\ref{sec:experiments}.
Summaries and error bars use means $\pm$ one SE.

\subsection{Replication and Uncertainty Calculations}
\label{sec:app-uncertainty}

For independent replication means $x_1,\ldots,x_N$ with average $\bar x$, we calculate
\begin{equation}
    \operatorname{SE}(\bar x)=\sqrt{\frac{\sum_{i=1}^{N}(x_i-\bar x)^2}{N(N-1)}}.
\end{equation}
Paired contrasts are formed within each replication before this calculation.
The persistent-state and approximate-feedback studies have ten independent environment tables per loss family, with 64 paired learner repetitions per table and condition.
Their uncertainty includes variation across tables after averaging learner repetitions.
The joint-versus-coordinatewise study propagates distributions exactly and has no sampling uncertainty.

The controlled restricted-feedback study uses four fixed environments and 64 selector repetitions per environment and condition.
SQL restricted-feedback replay uses 256 selector repetitions in each of three training banks, with independent selector streams across banks.
For $E$ equally weighted fixed environments or banks, let $s_e^2$ be the sample variance across $R_e$ selector repetitions in stratum $e$.
The conditional Monte Carlo standard error (MCSE) of the selector mean combines their independent mean variances as
\begin{equation}
    \operatorname{MCSE}(\bar x)=\frac{1}{E}\sqrt{\sum_{e=1}^{E}\frac{s_e^2}{R_e}}.
\end{equation}
Dispersion between the fixed environments or banks is not added to this conditional variance.

SQL training-seed summaries use the three separately trained candidate banks as replications on the same fixed benchmark.
They assume independent training randomization conditional on the shared construction data, benchmark, and base candidates.
The SEs themselves, however, are imprecisely estimated from only three replications.
Matched candidate subsets are averaged within a seed before calculating uncertainty because overlapping subsets are not additional replications.
For Spider, the six base-model prompt configurations share fixed outcomes across seeds.
Interaction SEs do not quantify uncertainty in selected family minima or variation over new databases.
Full-diagonal Spider expectations and full-feedback and draw-once replay expectations are exact within each bank, but may vary between training banks.

The original SParC deployment study shares selector randomness across training banks.
For selector repetition $r$, we first average the bank-specific quantity $x_{br}$ over the three banks,
\begin{equation}
    x_r=\frac{1}{3}\sum_{b=1}^{3}x_{br},
\end{equation}
then compute the SE over the 16 paired repetition averages.
Coupling differences are taken within each paired repetition before averaging.
This preserves cross-bank dependence rather than treating the banks as independent selector runs.
Training-seed uncertainty conditions on the benchmark, while conditional MCSEs additionally condition on the evaluated banks or environments.
These SE summaries do not provide confidence levels or significance tests.

\subsection{Joint and Coordinatewise Selection}
\label{sec:app-coordinate}

The coordinate-selection study uses the common family $\{0,1\}^2$, a uniform product prior, and exact full-information feedback.
The additive control has losses $(0.4,0.2,0.2,0)$ in configuration order $((0,0),(0,1),(1,0),(1,1))$.
The barrier condition has losses $(0.4,0.6,0.6,0)$, corresponding to $\Delta=0.4$ and $b=0.2$ in Proposition~\ref{prop:coordinate-fixed-rate-separation}.
Both tables are constant across rounds and share the base loss and optimal configuration, but their intermediate losses differ.
Exact joint EW and the coordinatewise sampler share the prior, loss table, and learning-rate schedule.
Both selectors initialize uniformly, and the coordinatewise sampler deploys only after a complete text-then-parameter sweep.
The best fixed comparator $(1,1)$ has zero loss, so expected cumulative loss equals regret.
Family restriction and state effects are also zero.

For the four-state sweep kernel $K_t$ and coordinatewise distribution $p_t$, we compute
\begin{align}
    p_1&=\mu,\\
    p_t&=p_{t-1}K_t^s,\qquad t\geq2,
\end{align}
where $s$ is the number of sweeps before deployment and every sweep in round $t$ uses the same target $q_t$.
Joint loss is computed directly from $q_t$ in~\eqref{eq:coordinate-ew-target}, and both targets use only preceding loss rows.
Expectations are evaluated exactly in floating-point arithmetic, without selector Monte Carlo error.
Conditional probabilities are normalized within each coordinate slice, avoiding division by globally underflowed target probabilities.
Independent tests enumerate short conditional-draw trajectories and check additive equivalence, target stationarity, and the stated bounds.
The 21 distinct conditions propagate 1,522,200 rounds in total.
Across all conditions, probability normalization and agreement of the two methods' regrets on additive losses held within floating-point tolerance.

The primary comparisons run through $T=100{,}000$ using either fixed $\eta=5$ or the slow schedule of Corollary~\ref{cor:coordinate-convergent-separation} with the known synthetic barrier $b=0.2$.
The additive control gives equal expected regret under both schedules.
For the barrier loss with fixed $\eta=5$, joint regret is $0.504$ at $T=100{,}000$, whereas one-sweep coordinatewise regret is $8{,}454.705$; see Figure~\ref{fig:coordinate-selection-regret}.
Under the slow schedule, joint regret is $0.460$, while coordinatewise regret grows from $35.264$ at $T=10{,}000$ to $111.154$ at $T=100{,}000$.
Over those same horizons, the coordinatewise probability of selecting $(1,1)$ rises from $99.56\%$ to $99.86\%$.
The finite-horizon curves illustrate the separations.
The almost-sure eventual fixation and regret lower bounds, however, follow from Section~\ref{sec:coordinate-selection}, rather than from these curves.

\begin{figure}[t]
    \centering
    \begin{subfigure}[t]{0.32\linewidth}
        \centering
        \includegraphics[width=\linewidth]{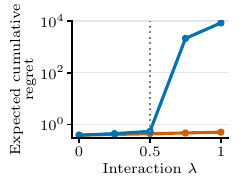}
        \caption{Interaction strength}
        \label{fig:coordinate-selection-interaction}
    \end{subfigure}\hspace{0.04\linewidth}
    \begin{subfigure}[t]{0.32\linewidth}
        \centering
        \includegraphics[width=\linewidth]{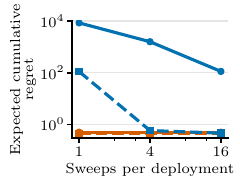}
        \caption{Sweep count}
        \label{fig:coordinate-selection-sweeps}
    \end{subfigure}
    \par\smallskip
    \includegraphics[width=\linewidth]{figures/coordinate_selection_legend.pdf}
    \caption{Sensitivity of exact expected regret to interaction strength and coordinatewise sweep count at $T=100{,}000$, with a shared logarithmic scale.
    The dotted line in Panel~\subref{fig:coordinate-selection-interaction} marks the onset of a strict coordinatewise local minimum.}
    \label{fig:coordinate-selection-sensitivity}
\end{figure}

The interaction sweep fixes $T=100{,}000$, $\eta=5$, and one sweep per deployment.
We retain $f(0,0)=0.4$ and $f(1,1)=0$ and set
\begin{equation}
    f(0,1)=f(1,0)=0.2+0.4\lambda,
    \qquad \lambda\in\{0,0.25,0.5,0.75,1\}.
\end{equation}
The table is additive at $\lambda=0$ and has a strict barrier at $(0,0)$ only for $\lambda>0.5$.
At $T=100{,}000$, coordinatewise regret is $0.443$ and $0.541$ for $\lambda=0.25$ and $0.5$, but rises to $2{,}112.357$ for $\lambda=0.75$.
The tested nonadditive tables therefore do not all exhibit a large regret gap; see Figure~\ref{fig:coordinate-selection-sensitivity}, Panel~\subref{fig:coordinate-selection-interaction}.
We also compare $s\in\{1,4,16\}$ sweeps under the fixed and slow schedules; see Figure~\ref{fig:coordinate-selection-sensitivity}, Panel~\subref{fig:coordinate-selection-sweeps}.
All intermediate draws remain undeployed, and extra sweeps receive no new feedback.
At the same horizon, four and sixteen sweeps reduce fixed-rate coordinatewise regret to $1{,}581.021$ and $112.628$, respectively.
Under the slow schedule, the corresponding regrets fall to $0.582$ and $0.462$, close to the joint value $0.460$.
These finite-horizon reductions do not imply that a fixed number of sweeps eliminates the fixed-rate trapping mechanism.
Additional checks use fixed rates $\eta\in\{1,2,5\}$ and the horizon-tuned rate $\eta=\sqrt{8\log(4)/T}$ at $T\in\{100,1{,}000,10{,}000,100{,}000\}$, restarting each horizon-tuned run from the common prior.
The fixed-rate coordinatewise regrets at $T=100{,}000$ are $906.794$, $2{,}854.616$, and $8{,}454.705$, respectively.
With the horizon-tuned rate, joint and coordinatewise regrets are instead $131.838$ and $132.180$, a gap of $0.342$.
However, the fixed-rate linear separation cannot be transferred uniformly to this horizon-dependent schedule.
These comparisons use the same number of deployment rounds but do not equalize sampling computation.
Their scope is selection among four fixed configurations, rather than prompt search, parameter training, or arbitrary alternating optimization.

\subsection{Selection and Signed State Effects}
\label{sec:app-persistent}

The persistent-state study uses configurations $\{0,1\}\times\{0,1\}$.
Ten independent environment seeds each generate a master table of length $4{,}096$ for each of three loss-table families, supplying prefixes at $T\in\{256,1{,}024,4{,}096\}$.
The first has identical diagonal losses across configurations.
The second has fixed means $(0.62,0.56,0.54,0.46)$ in configuration order $((0,0),(0,1),(1,0),(1,1))$, with bounded exogenous variation and a learning signal that does not shrink with the horizon.
The third changes the favored configuration over fixed 128-round phases.
All tables and schedules precede learner randomization.

Let $I_t$ indicate whether any relevant coordinate window contains different values, so that it is zero when both histories are constant.
A zero-memory coordinate contains only its current value and cannot trigger $I_t$.
The deployed loss follows the \hyperref[sec:exp-exact-persistent]{synthetic loss law in Section~\ref*{sec:experiments}}, with $\lambda\in\{-0.2,0,0.2\}$.
Interior diagonal losses keep $f_t\in[0,1]$.
Negative, zero, and positive $\lambda$ make mixed histories beneficial, neutral, and harmful, respectively.
This synthetic mechanism makes no claim about switching actual prompts or adapters.

The joint configuration family is evaluated across all loss-table families, horizons, $(m_Z,m_W)\in\{(0,0),(4,0),(0,4)\}$, and state coefficients.
A restriction panel evaluates text-only and parameter-only families on the fixed-advantage and changing-quality tables at $T=1{,}024$ under both positive-memory conditions.
Every panel compares independent and lazy EW with draw-once, which retains one initial sample, and with best fixed configurations.
Both EW selectors use the uniform prior and the rate from~\eqref{eq:lazy-learning-rate},
\begin{equation}
    \eta=\sqrt{\frac{\log|\mathcal C_e|}
                         {T/8+m_{\max}(T-1)}}.
    \label{eq:synthetic-declared-rate}
\end{equation}
This uses the family size, declared horizon, and influence bound, without the best candidate's identity.

Across the three loss-table families, each environment seed has 315 algorithm conditions and 64 learner tapes (sequences of random inputs) per condition, totaling 201,600 trajectories over ten seeds.
Shared tapes pair algorithm contrasts, which are averaged over tapes before forming the ten independent table means per loss-table family.
Each trajectory records excess deployed loss over the common joint comparator, expected diagonal selection regret, family restriction, signed state effect, and the sampling residual in~\eqref{eq:diagonal-sampling-residual}.

Draw-once removes learning and switches, while zero memory removes state effects.
When there is no selection signal, the coupling alone determines differences.
These controls test the accounting interpretation, while theorem-inequality checks verify implementation.

\begin{figure}[t]
    \centering
    \includegraphics[width=\linewidth]{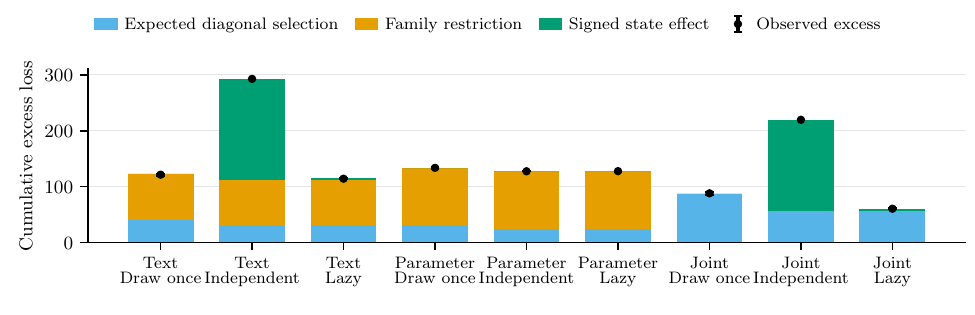}
    \caption{Decomposition of cumulative excess loss on fixed-advantage tables at $T=1{,}024$ and $\lambda=0.2$.
    Black marks show observed totals, including the diagonal sampling residual.}
    \label{fig:exp1-accounting}
\end{figure}

All trajectories satisfied the sampled decomposition to floating-point tolerance (maximum residual $2.8\times10^{-11}$), and matched EW selectors had exactly equal computed expected diagonal selection.
At $T=4{,}096$, $(m_Z,m_W)=(4,0)$, and $\lambda=0.2$, both had expected diagonal selection regret $144.797$ on the fixed-advantage tables.
Their mean state effects were $445.631$ and $4.038$, respectively, and sampled excess deployed losses were $590.328\pm0.994$ and $148.543\pm1.675$.
The sampling residual in~\eqref{eq:diagonal-sampling-residual} reconciles empirical totals with exact expected diagonal selection.
Figure~\ref{fig:exp1-accounting} illustrates the decomposition on fixed-advantage tables at $T=1{,}024$.

Reversing the state coefficient reversed the selector ordering: on fixed-advantage tables at $T=4{,}096$ and $(m_Z,m_W)=(4,0)$, the paired independent-minus-lazy excess-loss contrast was $441.785\pm1.404$ for $\lambda=0.2$, $-441.401\pm2.264$ for $\lambda=-0.2$, and $0.192\pm1.646$ for $\lambda=0$.

Relative to the common fixed joint-family optimum, mean text-only and parameter-only family restrictions on fixed-advantage tables at $T=1{,}024$ were $82.039$ and $102.701$.
Under independent selection with harmful state effects, changing influence from $(4,0)$ to $(0,4)$ changed text-versus-parameter mean excess loss from $292.850$ versus $127.504$ to $112.627$ versus $312.816$.
Candidate losses and expected diagonal selection were unchanged: these conditional reversals reflect the assigned influence mechanism, not intrinsic memory lengths of text or parameter artifacts.

\begin{figure}[t]
    \centering
    \begin{subfigure}[t]{0.32\linewidth}
        \centering
        \includegraphics[width=\linewidth]{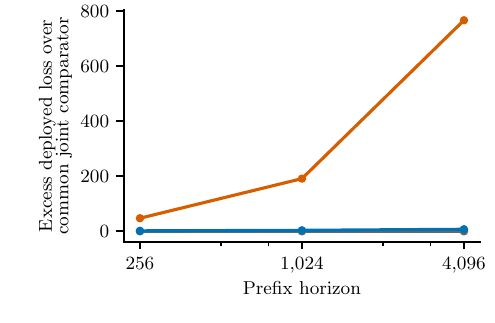}
        \caption{No selection signal}
        \label{fig:exp1-horizons-no-signal}
    \end{subfigure}
    \hfill
    \begin{subfigure}[t]{0.32\linewidth}
        \centering
        \includegraphics[width=\linewidth]{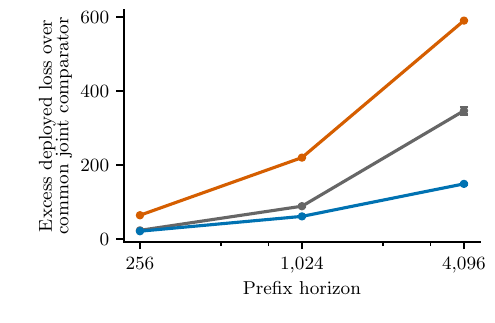}
        \caption{Fixed advantage}
        \label{fig:exp1-horizons-fixed-gap}
    \end{subfigure}
    \hfill
    \begin{subfigure}[t]{0.32\linewidth}
        \centering
        \includegraphics[width=\linewidth]{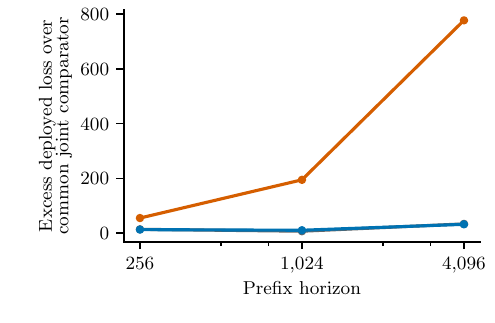}
        \caption{Changing relative quality}
        \label{fig:exp1-horizons-changing-quality}
    \end{subfigure}
    \includegraphics[width=\linewidth]{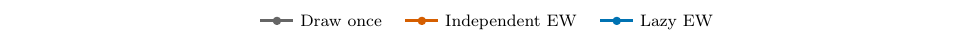}
    \caption{Horizon dependence and the draw-once control under harmful state with $\lambda=0.2$.}
    \label{fig:exp1-horizons}
\end{figure}

Learning was not uniformly beneficial: without a selection signal in the same harmful-state condition at $T=4{,}096$, draw-once had zero excess loss, versus $767.456$ for independent EW and $7.470$ for lazy EW; see Figure~\ref{fig:exp1-horizons}.
For changing quality, lazy EW's $32.835\pm1.959$ and draw-once's $34.386\pm0.996$ differed by less than the SE of the lazy estimate.
These mechanisms establish no unconditional advantage for learning, a coupling, or an adaptation mode.

\subsection{Feedback Accuracy, Switching, and Evaluation Cost}
\label{sec:app-approximate}

The approximate-feedback study reuses the fixed-advantage and changing-quality tables from Section~\ref{sec:app-persistent}, with $T\in\{1{,}024,4{,}096\}$, $(m_Z,m_W)\in\{(0,0),(4,0)\}$, all three state coefficients, and the same four-configuration joint family.

The eight feedback arms comprise exact losses, a common additive offset of $0.1$, two reference-opposing biases of magnitude $0.04$ and $0.08$, and Monte Carlo estimates with $n\in\{1,16,64,256\}$ samples per candidate-round.
The biases raise the best fixed reference's loss and lower its alternatives.
These synthetic errors are fixed from the true table before learning and make no assumptions about operational estimators.
The offset arm remains in $[0,1]$ without clipping.
Each estimate $\widehat d_t(a)$ averages $n$ independent Bernoulli samples of mean $d_t(a)$, independently of learner tapes.
We charge $4Tn$ samples per table and compare each table's maximum candidate--round error $|\widehat d_t(a)-d_t(a)|$ with Corollary~\ref{cor:diagonal-sampling-cost}'s envelope.
The Hoeffding envelope uses $\delta=0.05$ separately for each feedback table and is capped at one.
This envelope bounds within-table accuracy separately from empirical error bars and is not a simultaneous confidence statement over the experimental grid.

Each arm uses independent EW, lazy EW, and a centered lazy control with the same prior and rate~\eqref{eq:synthetic-declared-rate}.
The centered control diagnoses state effects by applying retention to $\widehat d_t(a)-\min_b\widehat d_t(b)$, preserving EW marginals while removing sensitivity to action-independent offsets.
Across both loss-table families, the grid has 576 algorithm conditions per environment seed and 64 paired learner tapes per condition, totaling 368,640 trajectories over ten seeds.

Approximate-minus-exact contrasts separate changes in true diagonal selection and signed state, while the shared configuration family fixes restriction.
Same-policy comparisons under true and estimated losses check the $2T\epsilon$ bridge in Corollary~\ref{cor:approximate-diagonal-evaluation}, which does not bound contrasts between differently trained learners.
Fresh Monte Carlo estimates satisfy Corollary~\ref{cor:unbiased-diagonal} even at $n=1$: budget affects accuracy and observed performance, but its coarse expected-regret guarantee holds at every budget.
As in Section~\ref{sec:app-persistent}, contrasts are paired at the table level.
The plotted offset and budget comparisons use $T=4{,}096$, $(m_Z,m_W)=(4,0)$, and $\lambda=0.2$.
The plotted reference-opposing bias comparison retains this horizon and influence pair, uses lazy EW with bias $0.08$, and varies $\lambda$ over the three state coefficients.

\begin{figure}[t]
    \centering
    \begin{subfigure}[t]{0.32\linewidth}
        \centering
        \includegraphics[width=\linewidth]{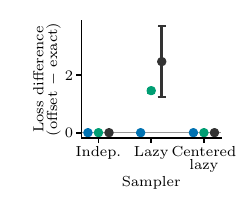}
        \caption{Common offset}
        \label{fig:exp2-offset-changing-quality}
    \end{subfigure}\hspace{0.04\linewidth}
    \begin{subfigure}[t]{0.32\linewidth}
        \centering
        \includegraphics[width=\linewidth]{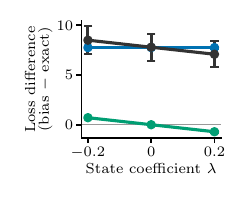}
        \caption{Reference-opposing bias}
        \label{fig:exp2-bias-changing-quality}
    \end{subfigure}
    \par\smallskip
    \includegraphics[width=\linewidth]{figures/approximate_feedback_perturbation_legend.pdf}
    \caption{Approximate-minus-exact loss components on changing-quality tables under a common offset in Panel~\subref{fig:exp2-offset-changing-quality} and reference-opposing bias for lazy EW in Panel~\subref{fig:exp2-bias-changing-quality}.
    The offset panel fixes $\lambda=0.2$, and the panels use different vertical scales.}
    \label{fig:exp2-perturbations-changing-quality}
\end{figure}

All trajectories passed the decomposition and same-policy bridge checks to floating-point tolerance (maximum accounting residual $2.6\times10^{-11}$).
The common offset preserved EW marginals and the trajectories of independent EW and centered lazy EW under shared tapes, but changed ordinary lazy EW's switches and state effects.
At $T=4{,}096$, $(m_Z,m_W)=(4,0)$, and $\lambda=0.2$, ordinary lazy EW's mean state effect increased by $0.779$ for fixed advantage and $1.458$ for changing quality.
The total-loss contrasts were $1.160\pm0.712$ and $2.464\pm1.223$; see Figure~\ref{fig:exp2-perturbations}, Panel~\subref{fig:exp2-offset-fixed-gap}, and Figure~\ref{fig:exp2-perturbations-changing-quality}, Panel~\subref{fig:exp2-offset-changing-quality}.
Their means are small relative to table-level variability, despite changed deployed behavior under identical selection distributions.

With reference-opposing bias $0.08$ in the displayed fixed-advantage condition, expected true diagonal selection increased by $208.607$ and mean harmful state cost by $2.690$.
For changing quality, selection increased by $7.738$ while the state term decreased by $0.707$, giving an observed total-loss increase of $7.069\pm1.291$; see Figure~\ref{fig:exp2-perturbations-changing-quality}, Panel~\subref{fig:exp2-bias-changing-quality}.

\begin{figure}[t]
    \centering
    \begin{subfigure}[t]{0.24\linewidth}
        \centering
        \includegraphics[width=\linewidth]{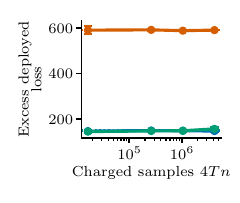}
        \caption{Fixed advantage}
        \label{fig:exp2-budget-loss-fixed-gap}
    \end{subfigure}\hfill
    \begin{subfigure}[t]{0.24\linewidth}
        \centering
        \includegraphics[width=\linewidth]{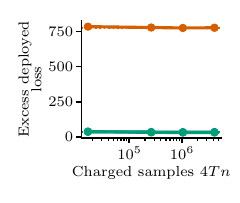}
        \caption{Changing quality}
        \label{fig:exp2-budget-loss-changing-quality}
    \end{subfigure}\hfill
    \begin{subfigure}[t]{0.24\linewidth}
        \centering
        \includegraphics[width=\linewidth]{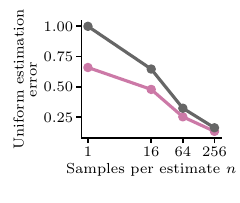}
        \caption{Fixed advantage}
        \label{fig:exp2-budget-error-fixed-gap}
    \end{subfigure}\hfill
    \begin{subfigure}[t]{0.24\linewidth}
        \centering
        \includegraphics[width=\linewidth]{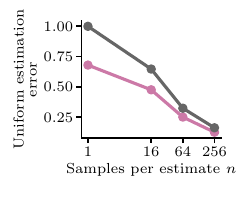}
        \caption{Changing quality}
        \label{fig:exp2-budget-error-changing-quality}
    \end{subfigure}
    \par\smallskip
    \includegraphics[width=\linewidth]{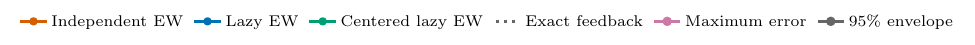}
    \caption{Feedback accuracy and excess deployed loss as the sampling budget grows.
    Dotted lines mark exact feedback, and the gray curve is the capped Hoeffding envelope.
    The loss panels use different vertical scales.}
    \label{fig:exp2-budget}
\end{figure}

At $\lambda=0.2$, increasing Bernoulli samples per candidate-round from $1$ to $256$ improved uniform accuracy without a monotone deployed-loss gain in the displayed conditions; see Table~\ref{tab:exp2-budget} and Figure~\ref{fig:exp2-budget}.
This is consistent with Corollary~\ref{cor:unbiased-diagonal}'s common coarse guarantee, without showing that additional samples never help.

For comparison, exact feedback gave excess loss $148.543\pm1.675$.
The reported sample counts represent Bernoulli observations rather than measured language-model calls.

\begin{table}[t]
    \centering
    \small
    \caption{Monte Carlo feedback for lazy EW on fixed-advantage tables at $T=4{,}096$, $(m_Z,m_W)=(4,0)$, and $\lambda=0.2$.
    Max. error is the table mean of the maximum candidate--round error, and observations count represented Bernoulli samples.}
    \label{tab:exp2-budget}
    \begin{tabular}{rrrr}
        \toprule
        \textbf{Samples $\boldsymbol{n}$} & \textbf{Observations/table} & \textbf{Max. error} & \textbf{Excess loss $\pm$ SE} \\
        \midrule
        1   & 16,384    & 0.660 & $147.835\pm4.906$ \\
        16  & 262,144   & 0.479 & $150.527\pm1.305$ \\
        64  & 1,048,576 & 0.252 & $149.555\pm2.290$ \\
        256 & 4,194,304 & 0.132 & $148.311\pm1.913$ \\
        \bottomrule
    \end{tabular}

\end{table}

A binomial draw divided by $n$ exactly implements the Bernoulli sample average.
The study represents 138,035,200 observations, charging each feedback table once despite reuse across learner and state conditions.
This known synthetic law tests estimation, without measuring external-simulator runtime or establishing off-policy identification from logged language-model interactions.

\subsection{Restricted Feedback on Controlled Losses}
\label{sec:app-controlled-feedback}

We compare the algorithms in Section~\ref{sec:feedback} on the same four-configuration \hyperref[sec:exp-exact-persistent]{synthetic loss law}.
The design crosses three diagonal-table families, horizons $T\in\{256,1{,}024\}$, equal coordinate influence windows $m_Z=m_W\in\{0,4,16\}$, and state coefficients $\lambda\in\{-0.2,0,0.2\}$.
Four independent environment seeds generate length-$1{,}024$ master tables, whose prefixes supply both horizons.
For each condition, we use 64 selector repetitions per environment.
Every selector uses the full joint family and the uniform prior, so restriction is zero relative to the best fixed configuration in that environment.
The zero-state control separates learning from history effects because changing $\lambda$ at fixed influence leaves the diagonal table and prescribed learning rates unchanged.

Full diagonal EW observes all $K$ current diagonal losses after deployment.
Sparse diagonal EW, called one-probe lazy EW in the main text, observes one independently and uniformly sampled candidate, using the unbiased importance-weighted vector in~\eqref{eq:sparse-diagonal-estimator}.
Neither learner updates from its deployed loss, and probe randomness is independent of deployment randomness.
Both use the lazy coupling, while independent deployment with the same diagonal updates provides a coupling control.
Deployment-only EW receives only the actual loss of its selected configuration, holds that configuration for blocks of length $b\in\{1,2,4,8,16\}$, and updates from the importance-weighted complete-block average in~\eqref{eq:block-bandit-estimator}.
An incomplete tail retains the last complete-block action without another draw or update.
Draw-once retains a uniform initial configuration throughout and supplies an exact expected-loss baseline without learning feedback.

Writing $m=\max\{m_Z,m_W\}$, the rates are
\begin{align}
    \eta_{\mathrm{full}}&=\sqrt{\frac{\log K}{T/8+m(T-1)}},\\
    \eta_{\mathrm{sparse}}&=\sqrt{\frac{\log K}{KT/2+m(T-1)}},\\
    \eta_{\mathrm{block}}&=\sqrt{\frac{2\log K}{K\lfloor T/b\rfloor}}.
\end{align}
Comparisons therefore evaluate each feedback regime with its prescribed rate, without holding learning rate fixed while changing information.
The $\lambda=0$ control also reveals how the conservative influence-dependent rates change learning when state itself is neutral.

The evaluation layer records true diagonal losses for accounting, without revealing them to learners that lack diagonal feedback.
For full and sparse diagonal algorithms, estimated expected deployed loss adds a sampled signed state effect to $\sum_t\langle q_t,d_t\rangle$, integrating the diagonal term over deployment randomness conditional on autonomous probes.
Deployment-only estimates average actual trajectory losses because their update histories depend on deployment.
Conditional selector MCSEs follow Appendix~\ref{sec:app-uncertainty} and exclude uncertainty over new environments.
This experiment evaluates information access and causal synthetic state.
Its measurements do not cover model-generation cost or reactive user behavior.

The fixed-advantage condition exposes the tradeoff between learning and state suppression.
With harmful state, deployment-only blocks of length 16 reduce excess relative to $b=1$, but remain above full-diagonal and sparse lazy EW; see Table~\ref{tab:feedback-controlled}.
The block comparison decomposes into selection plus state as $53.519+178.738$ versus $78.924+36.600$: less harmful state offsets worse diagonal selection.
The ordering reverses under beneficial state, where batching removes the advantage of changing histories over fixed configurations.

\begin{table}[t]
    \centering
    \small
    \caption{Controlled fixed-advantage excess loss at $T=1{,}024$ and $m_Z=m_W=4$, using 64 selector repetitions in each of four fixed environments.
    Draw-once expectations are exact for each table.}
    \label{tab:feedback-controlled}
    \begin{tabular}{lrrr}
        \toprule
        \textbf{Feedback and Selection} & $\boldsymbol{\lambda=-0.2}$ & $\boldsymbol{\lambda=0}$ & $\boldsymbol{\lambda=0.2}$ \\
        \midrule
        Full diagonal, lazy & $51.574\pm0.126$ & $56.707\pm0.000$ & $61.840\pm0.126$ \\
        One probe, lazy & $58.574\pm0.495$ & $62.983\pm0.503$ & $67.392\pm0.534$ \\
        Deployment only, $b=1$ & $-138.617\pm0.232$ & $52.135\pm0.565$ & $232.257\pm2.160$ \\
        Deployment only, $b=16$ & $41.820\pm0.436$ & $78.855\pm0.459$ & $115.524\pm0.565$ \\
        Draw once & $87.241\pm0.000$ & $87.241\pm0.000$ & $87.241\pm0.000$ \\
        \bottomrule
    \end{tabular}
\end{table}

Figure~\ref{fig:feedback-controlled} adds the intermediate block lengths and influence sweep, using different vertical ranges for harmful and beneficial state.

The independent-coupling diagnostics isolate state at fixed diagonal marginals.
At $T=1{,}024$, $m_Z=m_W=4$, and $\lambda=0.2$, sparse independent and sparse lazy EW have the same diagonal loss, but signed state effects $197.545$ and $4.409$, giving excess losses $260.527$ and $67.392$.
With no diagonal signal, selection regret is zero and excess is entirely a state effect.
Under changing quality at the same harmful-state setting, draw-once excess $8.680$ is below full lazy $14.303$ and sparse lazy $12.904$.
Even with $\lambda=0$, the conservative rates can change the ranking: fixed-advantage excess loss at $T=1{,}024$ and $m_Z=m_W=16$ is $71.190$ for full diagonal EW and $52.050$ for deployment-only $b=1$.
Thus the theoretical guarantees do not imply a universal finite-horizon empirical ranking across these loss laws.

\subsection{Candidate Construction and SQL Evaluation}
\label{sec:app-sql-protocol}

The SQL experiments use Qwen3.5-9B~\citep{qwen35modelcard} on Spider~\citep{YuT2018emnlp} and SParC~\citep{YuT2019acl}.
Construction, evaluation, and replay protocols are fixed before comparison.
Selection constructs no new prompts or adapters.
Spider and SParC were jointly partitioned by database, keeping identical schemas together: their official development databases formed the final evaluation set, twenty training-side databases supplied development, and 120 remained for construction.
Construction contained 5,876 Spider questions and 7,592 SParC turns, while development contained 1,124 Spider questions and 1,433 SParC turns.
Neither construction nor development used final evaluation databases, although checkpoint pretraining exposure to these public benchmarks is unknown.
Development scored execution on each original database, whereas final evaluation used released test suites~\citep{ZhongR2020emnlp}, covering the final evaluation databases but not the training databases.
These metrics differ: an accidentally correct result on one database can fail elsewhere in its suite.

The six text candidates used different instructions.
Candidate $z_0$ used direct SQL instructions, while $z_1$ used verification of schema references, joins, filters, aggregation, ordering, and limits.
Candidate $z_2$ used two construction-only SQL demonstrations, $z_3$ used explicit reference resolution and schema linking, and $z_4$ used decomposition into SQL clauses.
Finally, $z_5$ combined demonstrations with verification.
Each was paired with the base weights $w_0$ or one of three low-rank adaptation (LoRA) adapters~\citep{hu2022lora} trained at learning rates $5\times10^{-5}$, $10^{-4}$, and $2\times10^{-4}$.
We repeated this construction for seeds 20260910, 20260911, and 20260912, giving nine trained adapters and a separate $6\times4$ joint family per seed.
The three recipes are candidate choices within a replication, while the training seeds are replications that all share the six base-model prompt configurations.
Each adapter used rank 16, scaling 32, query and value projections, one epoch, microbatch size one, and eight-example gradient accumulation.
AdamW used $(\beta_1,\beta_2)=(0.9,0.999)$, $\epsilon=10^{-8}$, zero weight decay, unit gradient clipping, and fixed unshuffled examples.
Completion-only supervision trained 2,228,224 parameters on 11,752 construction examples, equally split between Spider and SParC and assigned cyclically to $z_0,z_1,z_2$.
The other three prompt templates served as evaluation candidates.
SParC construction prompts included earlier gold SQL from that dialogue, whereas evaluation excluded target SQL and future utterances.
The \hyperref[sec:exp-stateful-applied]{SParC study} reuses each seed's complete candidate family.

Generation used pinned model/tokenizer revisions, bfloat16 weights, greedy decoding without thinking, and limits of 8,192 prompt and 2,048 completion tokens.
The local executor required uncapped predictions to match reference results on every test-suite database, preserving SQL literals, \texttt{DISTINCT}, duplicate rows, and outer-query ordering.
Column-permutation comparison followed the pinned test-suite evaluator, but the executor was not the unmodified leaderboard script.
Invalid predictions, 30-second query timeouts, results exceeding 100,000 rows, and completion caps received unit loss, while reference or infrastructure failures stopped evaluation.
Whitespace-split comparison operators were normalized outside strings, identifiers, and comments, and SQLite decoding followed upstream malformed-UTF-8 handling.
All final evaluation references passed an audit before model evaluation.

\begin{table}[t]
    \centering
    \small
    \caption{Spider candidate test-suite accuracy (\%).
    Columns identify the base model and adapter learning rates.}
    \label{tab:sql-spider-candidates}
    \begin{tabular}{lrrrr}
        \toprule
        \textbf{Prompt} & \textbf{Base} & $\mathbf{5\times10^{-5}}$ & $\mathbf{10^{-4}}$ & $\mathbf{2\times10^{-4}}$ \\
        \midrule
        Direct $z_0$ & 66.05 & $69.21\pm0.28$ & $69.34\pm0.45$ & $69.50\pm1.01$ \\
        Verification $z_1$ & 65.28 & $68.73\pm0.45$ & $69.25\pm0.19$ & $69.54\pm0.88$ \\
        Demonstrations $z_2$ & 63.73 & $69.86\pm0.70$ & $70.15\pm0.87$ & $69.21\pm0.71$ \\
        Schema linking $z_3$ & 65.09 & $69.54\pm0.51$ & $69.60\pm0.35$ & $70.08\pm1.10$ \\
        Clause planning $z_4$ & 65.86 & $68.99\pm0.09$ & $69.60\pm0.21$ & $70.02\pm0.81$ \\
        Demonstrations + verification $z_5$ & 64.31 & $70.08\pm0.78$ & $70.25\pm0.86$ & $69.34\pm0.71$ \\
        \bottomrule
    \end{tabular}
\end{table}

\subsection{Selection and Family Restriction on Spider}
\label{sec:app-spider}

Within each training replication, the text family contains all six prompts with base weights, and the parameter family contains the direct prompt with base weights or any of that seed's three adapters.
Their union has nine configurations, and the joint family contains all 24 combinations.
All selectors in that replication share the best fixed joint-family comparator on the final evaluation questions, with restriction defined in Section~\ref{sec:protocol}.
These finite-family comparisons concern unequal candidate availability and do not isolate unrestricted adaptation modes.

We evaluated the full question--configuration matrix, then replayed selectors in frozen question order using only preceding rows.
EW used a uniform prior within its family and $\eta=\sqrt{8\log K/T}$ for family size $K$ and $T=1{,}034$.
Reported loss is the exact expectation $\sum_t\langle q_t,d_t\rangle$, without selector Monte Carlo error.
These full-diagonal expectations are also the lazy EW deployment losses: lazy and independent coupling share the same marginals, and Spider has no state effect.
Baselines were uniform selection, uniform draw-once, follow-the-leader with uniform tie breaking, and a development-selected candidate held fixed through final evaluation.
Uniform and draw-once have equal expected loss here.
The \hyperref[sec:exp-sql-feedback]{SQL feedback replay} separately restricts access to this matrix.


The best fixed joint candidates averaged $299.667\pm7.265$ failures across training seeds, whereas the text-family minimum was the shared fixed value $351$; see Table~\ref{tab:sql-spider-candidates}.
Text EW had mean selection regret $8.903$, compared with joint EW's $15.358\pm1.893$, but its mean restriction was $51.333\pm7.265$ failures.
Averaging within-seed accounting gives text selection plus restriction $8.903+51.333=60.236$, versus $15.358+0=15.358$ for the joint family; see Figure~\ref{fig:sql-spider-comparison}, Panel~\subref{fig:sql-spider-accounting}.
Joint EW also had lower mean expected loss than parameter and union EW, with a paired joint-minus-parameter contrast of $-4.329\pm2.270$ failures.
Its SE is substantial relative to the mean difference, so the observed ordering should not be treated as a general advantage.

\begin{table}[tbp]
    \centering
    \small
    \caption{Spider expected cumulative failures by family and selector.
    Uniform also gives the draw-once expectation.}
    \label{tab:sql-spider-selectors}
    \begin{tabular}{lrrrr}
        \toprule
        \textbf{Family} & \textbf{EW} & \textbf{Follow-the-leader} & \textbf{Uniform} & \textbf{Development-fixed} \\
        \midrule
        Text & $359.903\pm0.000$ & $358.517\pm0.000$ & $361.333\pm0.000$ & $351.000\pm0.000$ \\
        Parameter & $319.354\pm5.750$ & $316.222\pm8.814$ & $325.417\pm3.495$ & $313.667\pm9.404$ \\
        Union & $325.435\pm5.962$ & $319.139\pm8.585$ & $346.519\pm1.553$ & $313.667\pm9.404$ \\
        Joint & $315.024\pm6.332$ & $310.990\pm9.123$ & $326.306\pm3.513$ & $309.333\pm7.311$ \\
        \bottomrule
    \end{tabular}
\end{table}

EW was not the lowest-loss rule on these orders: development-fixed and follow-the-leader had lower mean loss in every family; see Table~\ref{tab:sql-spider-selectors}.

To control candidate count, we also evaluated every subset of size $K\in\{2,3,4\}$ containing $(z_0,w_0)$ within each family.
Each subset used the same uniform-prior EW rate for its size and the same full joint-family comparator for its seed.
We summarize the complete subset distribution without choosing a subset using final evaluation performance.
At $K=4$, there are 10 text, one parameter, 56 union, and 1,771 joint subsets per seed.
Overlapping subsets are not independent replications.

\begin{table}[t]
    \centering
    \small
    \caption{Spider excess failures after averaging all base-anchored subsets of each size within each seed, relative to the full joint-family comparator.}
    \label{tab:sql-spider-matched}
    \begin{tabular}{rrrrr}
        \toprule
        $\mathbf{K}$ & \textbf{Text} & \textbf{Parameter} & \textbf{Union} & \textbf{Joint} \\
        \midrule
        2 & $56.437\pm7.265$ & $26.627\pm5.059$ & $45.258\pm6.154$ & $31.479\pm4.488$ \\
        3 & $58.223\pm7.265$ & $21.693\pm4.498$ & $39.184\pm5.430$ & $23.931\pm3.665$ \\
        4 & $59.195\pm7.265$ & $19.687\pm4.162$ & $34.503\pm4.988$ & $20.851\pm3.293$ \\
        \bottomrule
    \end{tabular}
\end{table}

Parameter subsets have lower average excess loss than joint subsets at every tested size; see Table~\ref{tab:sql-spider-matched}.
At $K=4$, the paired parameter-minus-joint contrast is $-1.164\pm1.075$, with magnitude comparable to its SE; see Figure~\ref{fig:sql-spider-comparison}\subref{fig:sql-spider-matched}.

Candidate complementarity is a separate question from family size.
Relative to direct prompt $z_0$ and base weights $w_0$, define the interaction
\begin{equation}
    \Delta(z,w)=\frac{1}{T}\sum_{t=1}^T\bigl[d_t(z,w)-d_t(z,w_0)-d_t(z_0,w)+d_t(z_0,w_0)\bigr].
    \label{eq:sql-prompt-adapter-interaction}
\end{equation}
Negative values mean that joint use improves loss beyond additive effects.
Table~\ref{tab:sql-spider-interactions} shows larger negative mean interactions for demonstration prompts than for verification alone.
Demonstrations with base weights nevertheless have lower accuracy than direct prompting.

\begin{table}[t]
    \centering
    \small
    \caption{Spider prompt--adapter failure interactions in percentage points.
    Negative values indicate improvement beyond additive effects.}
    \label{tab:sql-spider-interactions}
    \begin{tabular}{lrrr}
        \toprule
        \textbf{Prompt} & $\mathbf{5\times10^{-5}}$ & $\mathbf{10^{-4}}$ & $\mathbf{2\times10^{-4}}$ \\
        \midrule
        Verification $z_1$ & $-0.29\pm0.17$ & $-0.68\pm0.30$ & $-0.81\pm0.17$ \\
        Demonstrations $z_2$ & $-2.97\pm0.47$ & $-3.13\pm0.59$ & $-2.03\pm0.31$ \\
        Schema linking $z_3$ & $-1.29\pm0.23$ & $-1.23\pm0.23$ & $-1.55\pm0.15$ \\
        Clause planning $z_4$ & $0.03\pm0.23$ & $-0.45\pm0.31$ & $-0.71\pm0.21$ \\
        Demonstrations + verification $z_5$ & $-2.61\pm0.53$ & $-2.64\pm0.49$ & $-1.58\pm0.31$ \\
        \bottomrule
    \end{tabular}
\end{table}

Each interaction is paired within a training bank before averaging across the three seeds.
Fourteen of the 15 mean interactions are negative, with clause planning ($z_4$) at the smallest training rate as the exception.
This descriptive pattern does not supply a multiple-comparison significance claim.
For the middle-rate comparison, see Figure~\ref{fig:sql-diagnostics}\subref{fig:sql-spider-interactions}.
The three Spider matrices each contain 24,816 logical candidate--question evaluations, with 34 completion-cap outcomes receiving unit loss across the three banks.
These logical counts include shared base configurations and do not measure a selector's physical deployment cost.

\subsection{State Effects on SParC}
\label{sec:app-sparc}

SParC~\citep{YuT2019acl} tests history effects under equal per-round configuration distributions.
We evaluated 1,203 turns in 422 complete dialogues on the same 20 final evaluation databases, using the complete 24-configuration family for each of the three training seeds.
Each seed and history condition uses its own best fixed comparator, evaluated on SParC trajectories rather than Spider questions.
Both conditions retained the schema and all raw user utterances through the current turn, excluding gold SQL and future utterances.
Recursive history additionally retained earlier generated SQL, allowing switches to alter inherited context, whereas cleared history removed it.

History reset at dialogue boundaries while EW statistics continued.
The longest dialogue had five turns, bounding all recursive influence by four rounds without assuming that deleting an old answer removes its indirect effects.
The prompt policy would drop the oldest generated answers at the token limit, but no history was dropped.
Model, decoding, token limits, test-suite scoring, and failure rules follow Appendix~\ref{sec:app-sql-protocol}.
Before final model evaluation, we added a missing table alias to one invalid released reference.
The frozen protocol recorded this correction, and no final evaluation records were excluded.

Each candidate's complete constant-configuration trajectory started every dialogue with empty generated history and supplied $d_t(a)$ in~\eqref{eq:common-diagonal-loss}.
Independent and lazy EW used preceding rows, the uniform prior, and~\eqref{eq:lazy-learning-rate} with $T=1{,}203$:
\begin{equation}
    \eta=\sqrt{\frac{\log 24}{T/8+4(T-1)}}.
    \label{eq:sql-sparc-rate}
\end{equation}
The cleared control retained this rate.
Mixed-history losses measured deployment effects without updating EW, preserving matched diagonal marginals.
This full diagonal oracle requires own-history losses unavailable from deployment alone.
The \hyperref[sec:exp-sql-feedback]{restricted-feedback replay} instead uses only the cleared condition, with its own zero-influence rate.

To estimate signed state effects, we generated independent and lazy sequences in sixteen paired repetitions on the frozen dialogues.
Summing each selected turn's mixed-history minus own-trajectory loss and averaging over repetitions estimates signed state effects.
Adding this mean to exact expected diagonal loss estimates expected deployment loss without sampled-diagonal noise.
For the deployed-loss contrast, paired state sums provide a direct estimate by Proposition~\ref{prop:matched-marginals}.
Bank averaging and paired conditional MCSEs follow Appendix~\ref{sec:app-uncertainty}, conditioning on the banks, benchmark, and model responses.

\begin{table}[t]
    \centering
    \small
    \caption{Recursive SParC accounting over 1,203 turns.
    Both couplings share exact selection regret, and restriction is zero.}
    \label{tab:sql-sparc-deployment}
    \begin{tabular}{lrrr}
        \toprule
        \textbf{Coupling} & \textbf{Selection Regret} & \textbf{State Effect} & \textbf{Deployed Loss} \\
        \midrule
        Independent & 26.108 & $-1.979\pm1.678$ & $489.795\pm1.678$ \\
        Lazy & 26.108 & $0.396\pm0.175$ & $492.170\pm0.175$ \\
        \bottomrule
    \end{tabular}
\end{table}

Under recursive history, the best fixed own-trajectory loss averaged $465.667$ across the three banks, and the exact expected diagonal loss averaged $491.774$; see Table~\ref{tab:sql-sparc-deployment}.
The paired lazy-minus-independent difference was $2.375\pm1.626$ failures, or $0.197\pm0.135$ percentage points; see Figure~\ref{fig:sql-diagnostics}\subref{fig:sql-sparc-state}.
Here $1.626$ is the MCSE of the paired mean, conditional on the evaluated banks, benchmark, and model responses.
The estimated difference is about 1.5~\mbox{MCSEs}, providing limited evidence for an expected-loss advantage of independent over lazy EW.
The exact expected loss of uniform draw-once averaged $500.500$ failures across the banks, compared with estimated losses of $489.795$ for independent EW and $492.170$ for lazy EW.
This draw-once expectation is computed directly over the possible initial configurations on the evaluated banks.

Conditional on the evaluated banks, exact expected switches averaged $1{,}144.669$ under independent coupling and $11.718$ under lazy coupling.
Within-dialogue expectations averaged $743.637$ and $7.637$, respectively.
Switches at reset boundaries, by contrast, carry no SQL forward.
Both couplings produced helpful and harmful changed-history turns.
Harmful-minus-helpful counts recover the signed effects for binary loss.
Thus fewer mixed histories remove beneficial and harmful effects together, so the large switch reduction does not imply lower deployment loss.

Cleared history supplies a negative control: every deployed turn has its selected diagonal loss.
Its best fixed loss averaged $479.667$, and both couplings had the same exact expected deployed loss of $508.846$, averaged over banks.
Clearing changes the diagonal table and its comparator, so differences in total loss across history conditions do not identify signed state effects.
The matched-coupling contrast is therefore defined within each condition.

Each seed and condition contains 28,872 logical oracle turns and 38,496 logical selector turns.
Because responses are reused, these counts do not represent physical generations.
One repeated base-model oracle outcome differs across recursive-history replications: the same generated SQL and token sequence for a \texttt{poker\_player} turn passed in two banks but incurred a prediction timeout in the third.
The original unit-loss timeout outcome is retained, so the bank-averaged result also includes this recorded scoring-runtime exception.
Spider and cleared-history SParC base columns agree exactly across all three banks.
The evaluated receipts and conditional selector errors do not measure deployment latency or adapter-switching savings.

Appendix~\ref{sec:app-limitations} collects the limits of these finite-family, fixed-benchmark studies.

\subsection{Restricted Feedback on Precomputed SQL Losses}
\label{sec:app-sql-feedback}

We specify the feedback rules, learning-rate formulas, and block lengths without using losses from the final evaluation set.

We apply the \hyperref[sec:exp-restricted-feedback]{feedback protocols from the controlled study} to the six completed 24-candidate loss matrices: Spider and cleared-history SParC for each training seed.
The replay preserves the original questions, candidate outcomes, fixed order, and uniform prior, and compares all regimes with the same best fixed joint candidate within each matrix.
Both tasks have zero state effect in this replay, so deployed excess loss equals diagonal selection regret and family restriction is zero.
Recursive SParC is excluded: its mixed-history deployment loss cannot be reconstructed from a fixed diagonal matrix.

\begin{table}[t]
    \centering
    \small
    \caption{SQL restricted-feedback excess failures above each bank's best fixed joint candidate, with separate conditional MCSEs.
    SParC uses cleared history and the zero-influence rate.}
    \label{tab:sql-feedback}
    \begin{tabular}{lrrrr}
        \toprule
        \multirow{2}{*}{\textbf{Feedback and Selection}} & \multicolumn{2}{c}{\textbf{Spider}} & \multicolumn{2}{c}{\textbf{Cleared SParC}} \\
        & \textbf{Mean $\pm$ SE} & \textbf{MCSE} & \textbf{Mean $\pm$ SE} & \textbf{MCSE} \\
        \midrule
        Full diagonal, lazy & $15.358\pm1.893$ & 0.000 & $17.674\pm0.443$ & 0.000 \\
        One probe, lazy & $24.165\pm4.312$ & 0.130 & $34.304\pm1.979$ & 0.241 \\
        Deployment only, $b=1$ & $24.340\pm4.536$ & 0.300 & $34.182\pm2.254$ & 0.336 \\
        Deployment only, $b=2$ & $25.232\pm4.541$ & 0.327 & $36.094\pm2.696$ & 0.342 \\
        Deployment only, $b=4$ & $25.431\pm4.694$ & 0.366 & $38.145\pm2.836$ & 0.368 \\
        Deployment only, $b=8$ & $25.382\pm3.865$ & 0.379 & $38.908\pm2.494$ & 0.388 \\
        Deployment only, $b=16$ & $25.461\pm4.276$ & 0.378 & $38.772\pm3.037$ & 0.415 \\
        Draw once & $26.639\pm4.549$ & 0.000 & $40.514\pm2.932$ & 0.000 \\
        \bottomrule
    \end{tabular}
\end{table}

For each matrix and feedback regime, we run 256 selector repetitions.
Full feedback exposes all 24 current diagonal entries only after deployment, while sparse feedback exposes one independent uniform probe.
Deployment-only feedback exposes the selected entry and uses $b\in\{1,2,4,8,16\}$.
Draw-once and the independent-coupling diagnostics match the controlled study.
The zero-influence setting gives full EW rate $\sqrt{8\log(24)/T}$, which differs from the rate~\eqref{eq:sql-sparc-rate} retained in SParC's recursive-versus-cleared comparison.
The two cleared-history experiments therefore test different selector protocols.

Full-feedback and draw-once expected losses are exact.
For sparse feedback, errors reflect probe randomness after integrating the deployed diagonal term, whereas deployment-only errors reflect actual trajectories.
Conditional MCSEs follow Appendix~\ref{sec:app-uncertainty}.
Logical diagonal access is $24T$ entries for full feedback and $T$ for one-probe feedback, while deployment-only learners receive no diagonal oracle and observe $T$ deployed outcomes.
All matrices had already been generated.

Full diagonal feedback has the lowest mean excess among these regimes on both tasks; see Table~\ref{tab:sql-feedback}.
Sparse feedback uses one twenty-fourth as many logical diagonal entries, with paired sparse-minus-full mean differences of $8.807\pm2.752$ expected failures on Spider and $16.630\pm1.872$ on cleared SParC.
The increase is larger and has a smaller training-seed SE on cleared SParC.
These differences compare the prescribed algorithms and their learning rates, not information alone.

The paired sparse-minus-deployment-only mean differences, with $b=1$ for deployment-only feedback, are $-0.175\pm0.320$ failures on Spider and $0.121\pm0.440$ on cleared SParC.
Both mean differences are smaller in magnitude than their training-seed SEs, so their observed ordering does not establish an advantage of either feedback type.
Increasing the deployment block length to 16 reduces mean switches from $972.034$ to $60.779$ on Spider and from $1{,}125.211$ to $71.332$ on cleared SParC, while increasing mean excess.
The paired $b=16$ minus $b=1$ excess-loss contrasts are $1.121\pm0.315$ and $4.590\pm0.787$, respectively, with a larger observed loss increase on cleared SParC.
Here switching cannot create beneficial or harmful state, so batching trades update opportunities for fewer configuration changes.
Figures~\ref{fig:sql-diagnostics}\subref{fig:sql-feedback-spider} and~\ref{fig:sql-diagnostics}\subref{fig:sql-feedback-sparc} show the training-seed averages, while Table~\ref{tab:sql-feedback} separately reports conditional selector uncertainty.

%% file: appendices/appendix_costs.tex
\section{Adaptation Costs and Amortization}
\label{sec:deployment}

After validating admissibility and target satisfaction as specified in Definitions~\ref{def:protocol} and~\ref{def:reachability}, compare preparation and operating costs under a common evaluator and scalarization.
Small regret alone does not establish target satisfaction.
A break-even comparison between two modes requires both to be feasible.
For $e\in\{Z,W\}$, let $\kappa_e$ denote one-time preparation and installation costs, and let $c_t(a_{t-1},a_t)$ denote transition costs, accounted for separately from the loss contamination in~\eqref{eq:pathwise-switch-contamination}.

For validated per-use realization losses $\varepsilon_Z,\varepsilon_W\geq0$ (zero for exact realization), expected scalarized costs over $N$ uses are
\begin{align}
    J_Z(N)&=\kappa_Z+N(r_Z+\varepsilon_Z)+I_Z,
    \label{eq:deployment-cost-text}\\
    J_W(N)&=\kappa_W+N(r_W+\varepsilon_W)+I_W.
    \label{eq:deployment-cost-parameter}
\end{align}
Here $r_e$ excludes realization loss, and $I_e$ charges interference, irreversibility, validation, or recovery once.
For existing candidates, the preparation component of $\kappa_e$ is a sunk cost.
The model excludes nonaffine and path-dependent costs.

\begin{proposition}[Feasibility-aware break-even threshold]
\label{prop:deployment-break-even}
Assume both realizations are admissible and satisfy the declared target
tolerance, and that
\begin{equation}
    r_W+\varepsilon_W<r_Z+\varepsilon_Z.
    \label{eq:deployment-per-use-advantage}
\end{equation}
A parameter-configuration update has no larger total cost than a
text-configuration update if and only if
\begin{equation}
    N
    \geq
    \frac{(\kappa_W-\kappa_Z)+(I_W-I_Z)}
    {(r_Z+\varepsilon_Z)-(r_W+\varepsilon_W)}.
    \label{eq:deployment-break-even}
\end{equation}
When the right-hand side is nonpositive, the parameter update is weakly
preferred for every $N\geq0$.
\end{proposition}

The proof is given in Appendix~\ref{sec:app-proofs-costs}.
If $r_W+\varepsilon_W\geq r_Z+\varepsilon_Z$, reuse alone creates no asymptotic parameter advantage.
Costs can enter episodic or bounded-influence losses only with the corresponding range and memory assumptions, which exclude arbitrary path dependence.

\paragraph{Illustrative Adaptation Settings.}
For revised instructions or retrieval, charge preparation per revision and retrieval per use, with $N$ ending at the next preparation or validation event.
For target-matched structured output, an adapter with higher fixed cost but lower per-use cost breaks even at~\eqref{eq:deployment-break-even}.
Per-use cost includes input processing, generation, and verification.
A model, task, or format change can invalidate this reuse horizon.
A retrieval collection and a format adapter may jointly satisfy a target that neither restricted candidate family meets, subject to the common resource budget.
This finite-family restriction does not establish nonreachability in the full protocol family.

\paragraph{Complexity-Adjusted Cost Certificates.}
For continued online selection, take validated, target-matched distributions $Q_Z,Q_W$ on the text and parameter slices in Section~\ref{sec:protocol}, and let
\begin{equation}
    D_Z=\operatorname{KL}(Q_Z\,\|\,\pi),\quad\quad
    D_W=\operatorname{KL}(Q_W\,\|\,\rho).
\end{equation}
Use the slice priors~\eqref{eq:text-slice-prior}--\eqref{eq:mode-specific-priors} and comparator-average costs, counting localization once in $\varepsilon_e$.
Assume scalarized losses in $[0,1]$, one mode-level setup charge, and no subsequent installation or transition costs.
For a general range of width $G$, multiply regret by $G$.
The stateless bound~\eqref{eq:general-kl-regret-optimized} gives
\begin{align}
    U_Z(N)&=\kappa_Z+N(r_Z+\varepsilon_Z)+I_Z
    +\sqrt{\frac{ND_Z}{2}},
    \label{eq:deployment-certificate-text}\\
    U_W(N)&=\kappa_W+N(r_W+\varepsilon_W)+I_W
    +\sqrt{\frac{ND_W}{2}}.
    \label{eq:deployment-certificate-parameter}
\end{align}
These selection-regret certificates satisfy $U_W(N)\leq U_Z(N)$ exactly when
\begin{equation}
    N\bigl((r_Z+\varepsilon_Z)-(r_W+\varepsilon_W)\bigr)
    \geq
    (\kappa_W-\kappa_Z)+(I_W-I_Z)
    +\sqrt{\frac{N}{2}}\bigl(\sqrt{D_W}-\sqrt{D_Z}\bigr).
    \label{eq:deployment-certificate-comparison}
\end{equation}
Equation~\eqref{eq:deployment-certificate-comparison} orders ex-post, comparator-tuned upper bounds, without establishing an ordering of realized costs.
Stateless switching adds $C_e^{\mathrm{sw}}=\mathbb E\sum_{t=2}^N c_t(A_{t-1},A_t)$ to $U_e$, including per-configuration installation and separately charging any unpaid initial transition.
Corollaries~\ref{cor:prior-atom-regret} and~\ref{cor:local-point-regret} give description-length and localized versions, with approximation counted in $\varepsilon_e$.

Count resource charges and signed state effects once each.
Persistent state additionally requires loss control, such as Theorem~\ref{thm:lazy-policy}, because transition charges alone do not justify these stateless certificates.

%% file: appendices/appendix_related_work.tex
\section{Additional Related Work}
\label{sec:app-related-work}

\paragraph{Automatic Optimization of Discrete Text.}
Prompt optimization uses token gradients~\citep{shin2020autoprompt}, generated or reinforcement-learned instructions~\citep{zhou2023ape,deng2022rlprompt}, critiques~\citep{pryzant2023protegi}, optimizer histories~\citep{yang2024opro}, evolution~\citep{guo2024evoprompt,fernando2024promptbreeder}, and textual gradients or pairwise feedback~\citep{yuksekgonul2025textgrad,feedbackdescent}.
MIPRO searches instructions and demonstrations~\citep{opsahlong2024mipro}, GEPA evolves textual system components~\citep{agrawal2026gepa}, and SPRIG evolves reusable prompts and evaluates transfer~\citep{ZhangL2026iclr}.
Appendix~\ref{sec:candidate-discovery} separates candidate discovery from selection.

\paragraph{Compound Systems, Context, and External Memory.}
Demonstrations and retrieval configure inference~\citep{brown2020language,min2022demonstrations,rubin2022retrieve,lewis2020rag}.
Text also governs tools~\citep{yao2023react,schick2023toolformer}, retained feedback and memory~\citep{shinn2023reflexion,packer2023memgpt}, and optimized pipelines or harnesses~\citep{khattab2024dspy,metaharness}.
A-Mem revises linked memories~\citep{XuW2025neurips}, LightMem separates online use from offline consolidation~\citep{FangJ2026iclr}, and LongMemEval evaluates sustained retention and updates~\citep{WuD2025iclr}.
These systems motivate reusable text with explicit state and resource accounting under protocol $\Xi$.

\paragraph{Prior Complexity, Description Length, and Language-Model Priors.}
Probably approximately correct (PAC)-Bayesian bounds charge prior KL~\citep{mcallester1999pacbayes}, while minimum description length connects coding and complexity~\citep{grunwald2007mdl}.
Language-model priors can also certify discrete prompts~\citep{akinwande2023prompt}.
Our online bounds use prior complexity without requiring independent and identically distributed samples.

\paragraph{Stateful and Adaptive Environments.}
Policy regret compares counterfactual trajectories~\citep{arora2012policy}.
Related approaches include learning with memory, online Markov decision processes, reactive experts, and stateful reference policies~\citep{anava2015memory,evendar2009online,neu2010online,defarias2006reactive,feige2017chasing}.
\citet{anava2015memory} obtain square-root rates with oblivious memory losses, counterfactual feedback, and low switching, the mechanism underlying our full-diagonal guarantee.
Decoupled exploration separates evaluation from deployment~\citep{AvnerO2012icml}, including under limited switching~\citep{ChengD2023icml}.
Deployment-only feedback instead motivates policy-regret mini-batching~\citep{arora2012policy}.

%% file: appendices/appendix_limitations.tex
\section{Limitations}
\label{sec:app-limitations}

Our analysis assumes a fixed, known candidate family constructed before evaluation and does not address how to construct or expand that family.
A practical extension is to expand the prompt and adapter banks systematically and repeat the comparisons, examining how candidate diversity affects selection performance and family restriction.
We leave this extension to future work.

Our language-model experiments focus on Spider and SParC with one base model.
Repeating the evaluation with additional models and task orders would help assess how consistently the observed patterns appear.
Longer dialogue sequences would also allow closer examination of how inherited state affects later decisions.

%% file: main.bbl
\begin{thebibliography}{80}
\providecommand{\natexlab}[1]{#1}
\providecommand{\url}[1]{\texttt{#1}}
\expandafter\ifx\csname urlstyle\endcsname\relax
  \providecommand{\doi}[1]{doi: #1}\else
  \providecommand{\doi}{doi: \begingroup \urlstyle{rm}\Url}\fi

\bibitem[Agrawal et~al.(2026)Agrawal, Tan, Soylu, Ziems, Khare, Opsahl-Ong,
  Singhvi, Shandilya, Ryan, Jiang, Potts, Sen, Dimakis, Stoica, Klein, Zaharia,
  and Khattab]{agrawal2026gepa}
L.~A. Agrawal, S.~Tan, D.~Soylu, N.~Ziems, R.~Khare, K.~Opsahl-Ong, A.~Singhvi,
  H.~Shandilya, M.~J. Ryan, M.~Jiang, C.~Potts, K.~Sen, A.~G. Dimakis,
  I.~Stoica, D.~Klein, M.~Zaharia, and O.~Khattab.
\newblock {GEPA}: Reflective prompt evolution can outperform reinforcement
  learning.
\newblock In \emph{Proceedings of the International Conference on Learning
  Representations (ICLR)}, 2026.

\bibitem[Akinwande et~al.(2024)Akinwande, Jiang, Sam, and
  Kolter]{akinwande2023prompt}
V.~Akinwande, Y.~Jiang, D.~Sam, and J.~Z. Kolter.
\newblock Understanding prompt engineering may not require rethinking
  generalization.
\newblock In \emph{Proceedings of the International Conference on Learning
  Representations (ICLR)}, 2024.

\bibitem[Altschuler \& Talwar(2018)Altschuler and
  Talwar]{altschuler2018limited}
J.~Altschuler and K.~Talwar.
\newblock Online learning over a finite action set with limited switching.
\newblock In \emph{Proceedings of the Conference on Learning Theory (COLT)},
  2018.

\bibitem[Anava et~al.(2015)Anava, Hazan, and Mannor]{anava2015memory}
O.~Anava, E.~Hazan, and S.~Mannor.
\newblock Online learning for adversaries with memory: Price of past mistakes.
\newblock In \emph{Advances in Neural Information Processing Systems
  (NeurIPS)}, 2015.

\bibitem[Arora et~al.(2012)Arora, Dekel, and Tewari]{arora2012policy}
R.~Arora, O.~Dekel, and A.~Tewari.
\newblock Online bandit learning against an adaptive adversary: From regret to
  policy regret.
\newblock In \emph{Proceedings of the International Conference on Machine
  Learning (ICML)}, 2012.

\bibitem[Asawa et~al.(2026)Asawa, Glaze, Orlanski, Ramakrishnan, Xu, Biswal,
  Chen, Sala, Zaharia, and Gonzalez]{continuallearningbench2026}
P.~Asawa, C.~M. Glaze, G.~Orlanski, R.~Ramakrishnan, B.~Xu, A.~Biswal, V.~S.
  Chen, F.~Sala, M.~Zaharia, and J.~E. Gonzalez.
\newblock Continual learning bench: Evaluating frontier {AI} systems in
  real-world stateful environments, 2026.
\newblock arXiv preprint arXiv:2606.05661.

\bibitem[Auer et~al.(2002)Auer, Cesa-Bianchi, Freund, and
  Schapire]{auer2002bandit}
P.~Auer, N.~Cesa-Bianchi, Y.~Freund, and R.~E. Schapire.
\newblock The nonstochastic multiarmed bandit problem.
\newblock \emph{SIAM Journal on Computing}, 32\penalty0 (1):\penalty0 48--77,
  2002.

\bibitem[Avner et~al.(2012)Avner, Mannor, and Shamir]{AvnerO2012icml}
O.~Avner, S.~Mannor, and O.~Shamir.
\newblock Decoupling exploration and exploitation in multi-armed bandits.
\newblock In \emph{Proceedings of the International Conference on Machine
  Learning (ICML)}, 2012.

\bibitem[Bo et~al.(2026)Bo, Li, Sun, Dai, Zhang, Tian, Chen, and
  Dong]{BoX2026iclr}
X.~Bo, R.~Li, Z.~Sun, Q.~Dai, Z.~Zhang, Z.~Tian, X.~Chen, and Z.~Dong.
\newblock Prompt and parameter co-optimization for large language models.
\newblock In \emph{Proceedings of the International Conference on Learning
  Representations (ICLR)}, 2026.

\bibitem[Brown et~al.(2020)Brown, Mann, Ryder, Subbiah, Kaplan, Dhariwal,
  Neelakantan, Shyam, Sastry, Askell, Agarwal, Herbert-Voss, Krueger, Henighan,
  Child, Ramesh, Ziegler, Wu, Winter, Hesse, Chen, Sigler, Litwin, Gray, Chess,
  Clark, Berner, McCandlish, Radford, Sutskever, and Amodei]{brown2020language}
T.~B. Brown, B.~Mann, N.~Ryder, M.~Subbiah, J.~Kaplan, P.~Dhariwal,
  A.~Neelakantan, P.~Shyam, G.~Sastry, A.~Askell, S.~Agarwal, A.~Herbert-Voss,
  G.~Krueger, T.~Henighan, R.~Child, A.~Ramesh, D.~M. Ziegler, J.~Wu,
  C.~Winter, C.~Hesse, M.~Chen, E.~Sigler, M.~Litwin, S.~Gray, B.~Chess,
  J.~Clark, C.~Berner, S.~McCandlish, A.~Radford, I.~Sutskever, and D.~Amodei.
\newblock Language models are few-shot learners.
\newblock In \emph{Advances in Neural Information Processing Systems
  (NeurIPS)}, 2020.

\bibitem[Bubeck \& Cesa-Bianchi(2012)Bubeck and Cesa-Bianchi]{bubeck2012regret}
S.~Bubeck and N.~Cesa-Bianchi.
\newblock Regret analysis of stochastic and nonstochastic multi-armed bandit
  problems.
\newblock \emph{Foundations and Trends in Machine Learning}, 5\penalty0
  (1):\penalty0 1--122, 2012.

\bibitem[Cesa-Bianchi \& Lugosi(2006)Cesa-Bianchi and
  Lugosi]{cesabianchi2006prediction}
N.~Cesa-Bianchi and G.~Lugosi.
\newblock \emph{Prediction, Learning, and Games}.
\newblock Cambridge University Press, 2006.

\bibitem[Cesa-Bianchi et~al.(2013)Cesa-Bianchi, Dekel, and
  Shamir]{cesabianchi2013switching}
N.~Cesa-Bianchi, O.~Dekel, and O.~Shamir.
\newblock Online learning with switching costs and other adaptive adversaries.
\newblock In \emph{Advances in Neural Information Processing Systems
  (NeurIPS)}, 2013.

\bibitem[Cheng et~al.(2023)Cheng, Zhou, and Ji]{ChengD2023icml}
D.~Cheng, X.~Zhou, and B.~Ji.
\newblock Understanding the role of feedback in online learning with switching
  costs.
\newblock In \emph{Proceedings of the International Conference on Machine
  Learning (ICML)}, 2023.

\bibitem[Cortes et~al.(2019)Cortes, DeSalvo, Gentile, Mohri, and
  Yang]{cortes2019sleeping}
C.~Cortes, G.~DeSalvo, C.~Gentile, M.~Mohri, and S.~Yang.
\newblock Online learning with sleeping experts and feedback graphs.
\newblock In \emph{Proceedings of the International Conference on Machine
  Learning (ICML)}, 2019.

\bibitem[de~Farias \& Megiddo(2006)de~Farias and Megiddo]{defarias2006reactive}
D.~P. de~Farias and N.~Megiddo.
\newblock Combining expert advice in reactive environments.
\newblock \emph{Journal of the ACM}, 53\penalty0 (5):\penalty0 762--799, 2006.

\bibitem[Dekel et~al.(2014)Dekel, Ding, Koren, and Peres]{dekel2014bandits}
O.~Dekel, J.~Ding, T.~Koren, and Y.~Peres.
\newblock Bandits with switching costs: {$T^{2/3}$} regret.
\newblock In \emph{Proceedings of the Annual ACM Symposium on Theory of
  Computing (STOC)}, 2014.

\bibitem[Deng et~al.(2022)Deng, Wang, Hsieh, Wang, Guo, Shu, Song, Xing, and
  Hu]{deng2022rlprompt}
M.~Deng, J.~Wang, C.-P. Hsieh, Y.~Wang, H.~Guo, T.~Shu, M.~Song, E.~Xing, and
  Z.~Hu.
\newblock {RLPrompt}: Optimizing discrete text prompts with reinforcement
  learning.
\newblock In \emph{Proceedings of the Conference on Empirical Methods in
  Natural Language Processing (EMNLP)}, 2022.

\bibitem[Dud{\'{\i}}k et~al.(2011)Dud{\'{\i}}k, Langford, and
  Li]{dudik2011doubly}
M.~Dud{\'{\i}}k, J.~Langford, and L.~Li.
\newblock Doubly robust policy evaluation and learning.
\newblock In \emph{Proceedings of the International Conference on Machine
  Learning (ICML)}, 2011.

\bibitem[Even-Dar et~al.(2009)Even-Dar, Kakade, and Mansour]{evendar2009online}
E.~Even-Dar, S.~M. Kakade, and Y.~Mansour.
\newblock Online {Markov} decision processes.
\newblock \emph{Mathematics of Operations Research}, 34\penalty0 (3):\penalty0
  726--736, 2009.

\bibitem[Fang et~al.(2026)Fang, Deng, Xu, Jiang, Tang, Xu, Deng, Yao, Wang,
  Qiao, Chen, and Zhang]{FangJ2026iclr}
J.~Fang, X.~Deng, H.~Xu, Z.~Jiang, Y.~Tang, Z.~Xu, S.~Deng, Y.~Yao, M.~Wang,
  S.~Qiao, H.~Chen, and N.~Zhang.
\newblock {LightMem}: Lightweight and efficient memory-augmented generation.
\newblock In \emph{Proceedings of the International Conference on Learning
  Representations (ICLR)}, 2026.

\bibitem[Feige et~al.(2017)Feige, Koren, and Tennenholtz]{feige2017chasing}
U.~Feige, T.~Koren, and M.~Tennenholtz.
\newblock Chasing ghosts: Competing with stateful policies.
\newblock \emph{SIAM Journal on Computing}, 46\penalty0 (1):\penalty0 190--223,
  2017.

\bibitem[Fernando et~al.(2024)Fernando, Banarse, Michalewski, Osindero, and
  Rockt{\"a}schel]{fernando2024promptbreeder}
C.~Fernando, D.~S. Banarse, H.~Michalewski, S.~Osindero, and
  T.~Rockt{\"a}schel.
\newblock {Promptbreeder}: Self-referential self-improvement via prompt
  evolution.
\newblock In \emph{Proceedings of the International Conference on Machine
  Learning (ICML)}, 2024.

\bibitem[Foster \& Vohra(1999)Foster and Vohra]{foster1999regret}
D.~P. Foster and R.~V. Vohra.
\newblock Regret in the on-line decision problem.
\newblock \emph{Games and Economic Behavior}, 29\penalty0 (1--2):\penalty0
  7--35, 1999.

\bibitem[Freund \& Schapire(1997)Freund and Schapire]{freund1997decision}
Y.~Freund and R.~E. Schapire.
\newblock A decision-theoretic generalization of on-line learning and an
  application to boosting.
\newblock \emph{Journal of Computer and System Sciences}, 55\penalty0
  (1):\penalty0 119--139, 1997.

\bibitem[Genewein et~al.(2025)Genewein, Li, Grau-Moya, Ruoss, Orseau, and
  Hutter]{genewein2025prompt}
T.~Genewein, K.~Li, J.~Grau-Moya, A.~Ruoss, L.~Orseau, and M.~Hutter.
\newblock Understanding prompt tuning and in-context learning via
  meta-learning.
\newblock In \emph{Advances in Neural Information Processing Systems
  (NeurIPS)}, 2025.

\bibitem[Geulen et~al.(2010)Geulen, V{\"o}cking, and Winkler]{geulen2010regret}
S.~Geulen, B.~V{\"o}cking, and M.~Winkler.
\newblock Regret minimization for online buffering problems using the weighted
  majority algorithm.
\newblock In \emph{Proceedings of the Conference on Learning Theory (COLT)},
  2010.

\bibitem[Ghosh et~al.(2026)Ghosh, Das, Speicher, Wu, Khan, Garg, Gummadi, and
  Terzi]{GhoshB2026acl}
B.~Ghosh, S.~Das, T.~Speicher, Q.~Wu, M.~A. Khan, D.~Garg, K.~P. Gummadi, and
  E.~Terzi.
\newblock Fine-tuning vs. in-context learning in large language models: A
  formal language learning perspective.
\newblock In \emph{Proceedings of the Annual Meeting of the Association for
  Computational Linguistics (ACL)}, 2026.

\bibitem[Gr{\"u}nwald(2007)]{grunwald2007mdl}
P.~D. Gr{\"u}nwald.
\newblock \emph{The Minimum Description Length Principle}.
\newblock MIT Press, 2007.

\bibitem[Guo et~al.(2024)Guo, Wang, Guo, Li, Song, Tan, Liu, Bian, and
  Yang]{guo2024evoprompt}
Q.~Guo, R.~Wang, J.~Guo, B.~Li, K.~Song, X.~Tan, G.~Liu, J.~Bian, and Y.~Yang.
\newblock Connecting large language models with evolutionary algorithms yields
  powerful prompt optimizers.
\newblock In \emph{Proceedings of the International Conference on Learning
  Representations (ICLR)}, 2024.

\bibitem[Hannan(1957)]{hannan1957bayes}
J.~Hannan.
\newblock Approximation to {Bayes} risk in repeated play.
\newblock In M.~Dresher, A.~W. Tucker, and P.~Wolfe (eds.), \emph{Contributions
  to the Theory of Games, Volume {III}}, pp.\  97--139. Princeton University
  Press, Princeton, NJ, 1957.

\bibitem[Herbster \& Warmuth(1998)Herbster and Warmuth]{herbster1998tracking}
M.~Herbster and M.~K. Warmuth.
\newblock Tracking the best expert.
\newblock \emph{Machine Learning}, 32:\penalty0 151--178, 1998.

\bibitem[Hoeffding(1963)]{hoeffding1963probability}
W.~Hoeffding.
\newblock Probability inequalities for sums of bounded random variables.
\newblock \emph{Journal of the American Statistical Association}, 58\penalty0
  (301):\penalty0 13--30, 1963.

\bibitem[Houlsby et~al.(2019)Houlsby, Giurgiu, Jastrzebski, Morrone, {De
  Laroussilhe}, Gesmundo, Attariyan, and Gelly]{houlsby2019adapters}
N.~Houlsby, A.~Giurgiu, S.~Jastrzebski, B.~Morrone, Q.~{De Laroussilhe},
  A.~Gesmundo, M.~Attariyan, and S.~Gelly.
\newblock Parameter-efficient transfer learning for {NLP}.
\newblock In \emph{Proceedings of the International Conference on Machine
  Learning (ICML)}, 2019.

\bibitem[Hu et~al.(2022)Hu, Shen, Wallis, Allen-Zhu, Li, Wang, Wang, and
  Chen]{hu2022lora}
E.~J. Hu, Y.~Shen, P.~Wallis, Z.~Allen-Zhu, Y.~Li, S.~Wang, L.~Wang, and
  W.~Chen.
\newblock {LoRA}: Low-rank adaptation of large language models.
\newblock In \emph{Proceedings of the International Conference on Learning
  Representations (ICLR)}, 2022.

\bibitem[Jiang \& Li(2016)Jiang and Li]{jiang2016doubly}
N.~Jiang and L.~Li.
\newblock Doubly robust off-policy value evaluation for reinforcement learning.
\newblock In \emph{Proceedings of the International Conference on Machine
  Learning (ICML)}, 2016.

\bibitem[Kalai \& Vempala(2005)Kalai and Vempala]{kalai2005efficient}
A.~Kalai and S.~Vempala.
\newblock Efficient algorithms for online decision problems.
\newblock \emph{Journal of Computer and System Sciences}, 71\penalty0
  (3):\penalty0 291--307, 2005.

\bibitem[Khattab et~al.(2024)Khattab, Singhvi, Maheshwari, Zhang, Santhanam,
  Vardhamanan, Haq, Sharma, Joshi, Moazam, Miller, Zaharia, and
  Potts]{khattab2024dspy}
O.~Khattab, A.~Singhvi, P.~Maheshwari, Z.~Zhang, K.~Santhanam, S.~Vardhamanan,
  S.~Haq, A.~Sharma, T.~T. Joshi, H.~Moazam, H.~Miller, M.~Zaharia, and
  C.~Potts.
\newblock {DSPy}: Compiling declarative language model calls into
  self-improving pipelines.
\newblock In \emph{Proceedings of the International Conference on Learning
  Representations (ICLR)}, 2024.

\bibitem[Kim et~al.(2026)Kim, Lee, Lee, Finn, and Lee]{KimH2026whale}
H.~Kim, Y.~Lee, G.~Lee, C.~Finn, and K.~Lee.
\newblock {WHALE}: A simple recipe for joint harness-weight optimization, 2026.
\newblock arXiv preprint arXiv:2609.00196.

\bibitem[Kirkpatrick et~al.(2017)Kirkpatrick, Pascanu, Rabinowitz, Veness,
  Desjardins, Rusu, Milan, Quan, Ramalho, Grabska-Barwinska, Hassabis, Clopath,
  Kumaran, and Hadsell]{kirkpatrick2017overcoming}
J.~Kirkpatrick, R.~Pascanu, N.~Rabinowitz, J.~Veness, G.~Desjardins, A.~A.
  Rusu, K.~Milan, J.~Quan, T.~Ramalho, A.~Grabska-Barwinska, D.~Hassabis,
  C.~Clopath, D.~Kumaran, and R.~Hadsell.
\newblock Overcoming catastrophic forgetting in neural networks.
\newblock \emph{Proceedings of the National Academy of Sciences}, 114\penalty0
  (13):\penalty0 3521--3526, 2017.

\bibitem[Lee et~al.(2026{\natexlab{a}})Lee, Boen, and Finn]{feedbackdescent}
Y.~Lee, J.~Boen, and C.~Finn.
\newblock Feedback descent: Open-ended text optimization via pairwise
  comparison.
\newblock In \emph{ICLR Workshop on AI with Recursive Self-Improvement},
  2026{\natexlab{a}}.

\bibitem[Lee et~al.(2026{\natexlab{b}})Lee, Nair, Zhang, Lee, Khattab, and
  Finn]{metaharness}
Y.~Lee, R.~S. Nair, Q.~Zhang, K.~Lee, O.~Khattab, and C.~Finn.
\newblock {Meta-Harness}: End-to-end optimization of model harnesses.
\newblock In \emph{Proceedings of the Conference on Language Modeling (COLM)},
  2026{\natexlab{b}}.

\bibitem[Lester et~al.(2021)Lester, Al-Rfou, and Constant]{lester2021prompt}
B.~Lester, R.~Al-Rfou, and N.~Constant.
\newblock The power of scale for parameter-efficient prompt tuning.
\newblock In \emph{Proceedings of the Conference on Empirical Methods in
  Natural Language Processing (EMNLP)}, 2021.

\bibitem[Lewis et~al.(2020)Lewis, Perez, Piktus, Petroni, Karpukhin, Goyal,
  K{\"u}ttler, Lewis, t.~Yih, Rockt{\"a}schel, Riedel, and Kiela]{lewis2020rag}
P.~Lewis, E.~Perez, A.~Piktus, F.~Petroni, V.~Karpukhin, N.~Goyal,
  H.~K{\"u}ttler, M.~Lewis, W.~t.~Yih, T.~Rockt{\"a}schel, S.~Riedel, and
  D.~Kiela.
\newblock Retrieval-augmented generation for knowledge-intensive {NLP} tasks.
\newblock In \emph{Advances in Neural Information Processing Systems
  (NeurIPS)}, 2020.

\bibitem[Li \& Liang(2021)Li and Liang]{li2021prefix}
X.~L. Li and P.~Liang.
\newblock {Prefix-Tuning}: Optimizing continuous prompts for generation.
\newblock In \emph{Proceedings of the Annual Meeting of the Association for
  Computational Linguistics and the International Joint Conference on Natural
  Language Processing (ACL-IJCNLP)}, 2021.

\bibitem[Littlestone \& Warmuth(1994)Littlestone and
  Warmuth]{littlestone1994weighted}
N.~Littlestone and M.~K. Warmuth.
\newblock The weighted majority algorithm.
\newblock \emph{Information and Computation}, 108\penalty0 (2):\penalty0
  212--261, 1994.

\bibitem[Liu et~al.(2022)Liu, Tam, Muqeeth, Mohta, Huang, Bansal, and
  Raffel]{liu2022fewshot}
H.~Liu, D.~Tam, M.~Muqeeth, J.~Mohta, T.~Huang, M.~Bansal, and C.~A. Raffel.
\newblock Few-shot parameter-efficient fine-tuning is better and cheaper than
  in-context learning.
\newblock In \emph{Advances in Neural Information Processing Systems
  (NeurIPS)}, 2022.

\bibitem[McAllester(1999)]{mcallester1999pacbayes}
D.~A. McAllester.
\newblock {PAC}-{Bayesian} model averaging.
\newblock In \emph{Proceedings of the Conference on Learning Theory (COLT)},
  1999.

\bibitem[Meng et~al.(2022)Meng, Bau, Andonian, and Belinkov]{meng2022rome}
K.~Meng, D.~Bau, A.~Andonian, and Y.~Belinkov.
\newblock Locating and editing factual associations in {GPT}.
\newblock In \emph{Advances in Neural Information Processing Systems
  (NeurIPS)}, 2022.

\bibitem[Min et~al.(2022)Min, Lyu, Holtzman, Artetxe, Lewis, Hajishirzi, and
  Zettlemoyer]{min2022demonstrations}
S.~Min, X.~Lyu, A.~Holtzman, M.~Artetxe, M.~Lewis, H.~Hajishirzi, and
  L.~Zettlemoyer.
\newblock Rethinking the role of demonstrations: What makes in-context learning
  work?
\newblock In \emph{Proceedings of the Conference on Empirical Methods in
  Natural Language Processing (EMNLP)}, 2022.

\bibitem[Mosbach et~al.(2023)Mosbach, Pimentel, Ravfogel, Klakow, and
  Elazar]{mosbach2023fewshot}
M.~Mosbach, T.~Pimentel, S.~Ravfogel, D.~Klakow, and Y.~Elazar.
\newblock Few-shot fine-tuning vs. in-context learning: A fair comparison and
  evaluation.
\newblock In \emph{Findings of the Association for Computational Linguistics:
  ACL}, 2023.

\bibitem[Mourtada \& Maillard(2017)Mourtada and Maillard]{mourtada2017growing}
J.~Mourtada and O.-A. Maillard.
\newblock Efficient tracking of a growing number of experts.
\newblock In \emph{Proceedings of the International Conference on Algorithmic
  Learning Theory (ALT)}, 2017.

\bibitem[Neu et~al.(2010)Neu, Antos, Gy{\"o}rgy, and
  Szepesv{\'a}ri]{neu2010online}
G.~Neu, A.~Antos, A.~Gy{\"o}rgy, and C.~Szepesv{\'a}ri.
\newblock Online {Markov} decision processes under bandit feedback.
\newblock In \emph{Advances in Neural Information Processing Systems
  (NeurIPS)}, 2010.

\bibitem[{OpenAI}(2026{\natexlab{a}})]{OpenAI2026gpt56}
{OpenAI}.
\newblock {GPT-5.6}: Frontier intelligence that scales with your ambition.
\newblock \url{https://openai.com/index/gpt-5-6/}, 2026{\natexlab{a}}.
\newblock Accessed September 25, 2026.

\bibitem[{OpenAI}(2026{\natexlab{b}})]{OpenAI2026gpt6astra}
{OpenAI}.
\newblock {GPT-6 Astra}: A new generation of intelligence.
\newblock \url{https://openai.com/index/gpt-6-astra/}, 2026{\natexlab{b}}.
\newblock Accessed September 25, 2026.

\bibitem[Opsahl-Ong et~al.(2024)Opsahl-Ong, Ryan, Purtell, Broman, Potts,
  Zaharia, and Khattab]{opsahlong2024mipro}
K.~Opsahl-Ong, M.~J. Ryan, J.~Purtell, D.~Broman, C.~Potts, M.~Zaharia, and
  O.~Khattab.
\newblock Optimizing instructions and demonstrations for multi-stage language
  model programs.
\newblock In \emph{Proceedings of the Conference on Empirical Methods in
  Natural Language Processing (EMNLP)}, 2024.

\bibitem[Orabona \& P{\'a}l(2016)Orabona and P{\'a}l]{orabona2016coin}
F.~Orabona and D.~P{\'a}l.
\newblock Coin betting and parameter-free online learning.
\newblock In \emph{Advances in Neural Information Processing Systems
  (NeurIPS)}, 2016.

\bibitem[Packer et~al.(2023)Packer, Wooders, Lin, Fang, Patil, Stoica, and
  Gonzalez]{packer2023memgpt}
C.~Packer, S.~Wooders, K.~Lin, V.~Fang, S.~G. Patil, I.~Stoica, and J.~E.
  Gonzalez.
\newblock {MemGPT}: Towards {LLM}s as operating systems, 2023.
\newblock arXiv preprint arXiv:2310.08560.

\bibitem[Pryzant et~al.(2023)Pryzant, Iter, Li, Lee, Zhu, and
  Zeng]{pryzant2023protegi}
R.~Pryzant, D.~Iter, J.~Li, Y.~Lee, C.~Zhu, and M.~Zeng.
\newblock Automatic prompt optimization with ``gradient descent'' and beam
  search.
\newblock In \emph{Proceedings of the Conference on Empirical Methods in
  Natural Language Processing (EMNLP)}, 2023.

\bibitem[{Qwen Team}(2026)]{qwen35modelcard}
{Qwen Team}.
\newblock {Qwen3.5-9B} model card.
\newblock \url{https://huggingface.co/Qwen/Qwen3.5-9B}, 2026.
\newblock Accessed August 30, 2026.

\bibitem[Razumovskaia et~al.(2025)Razumovskaia, Vuli{\'c}, and
  Korhonen]{RazumovskaiaE2025tacl}
E.~Razumovskaia, I.~Vuli{\'c}, and A.~Korhonen.
\newblock Analyzing and adapting large language models for few-shot
  multilingual {NLU}: Are we there yet?
\newblock \emph{Transactions of the Association for Computational Linguistics},
  13:\penalty0 1096--1120, 2025.

\bibitem[Rubin et~al.(2022)Rubin, Herzig, and Berant]{rubin2022retrieve}
O.~Rubin, J.~Herzig, and J.~Berant.
\newblock Learning to retrieve prompts for in-context learning.
\newblock In \emph{Proceedings of the Conference of the North American Chapter
  of the Association for Computational Linguistics (NAACL)}, 2022.

\bibitem[Schick et~al.(2023)Schick, Dwivedi-Yu, Dess{\`i}, Raileanu, Lomeli,
  Hambro, Zettlemoyer, Cancedda, and Scialom]{schick2023toolformer}
T.~Schick, J.~Dwivedi-Yu, R.~Dess{\`i}, R.~Raileanu, M.~Lomeli, E.~Hambro,
  L.~Zettlemoyer, N.~Cancedda, and T.~Scialom.
\newblock {Toolformer}: Language models can teach themselves to use tools.
\newblock In \emph{Advances in Neural Information Processing Systems
  (NeurIPS)}, 2023.

\bibitem[Shin et~al.(2020)Shin, Razeghi, {Logan IV}, Wallace, and
  Singh]{shin2020autoprompt}
T.~Shin, Y.~Razeghi, R.~L. {Logan IV}, E.~Wallace, and S.~Singh.
\newblock {AutoPrompt}: Eliciting knowledge from language models with
  automatically generated prompts.
\newblock In \emph{Proceedings of the Conference on Empirical Methods in
  Natural Language Processing (EMNLP)}, 2020.

\bibitem[Shinn et~al.(2023)Shinn, Cassano, Gopinath, Narasimhan, and
  Yao]{shinn2023reflexion}
N.~Shinn, F.~Cassano, A.~Gopinath, K.~Narasimhan, and S.~Yao.
\newblock {Reflexion}: Language agents with verbal reinforcement learning.
\newblock In \emph{Advances in Neural Information Processing Systems
  (NeurIPS)}, 2023.

\bibitem[Shuttleworth et~al.(2025)Shuttleworth, Andreas, Torralba, and
  Sharma]{ShuttleworthR2025neurips}
R.~Shuttleworth, J.~Andreas, A.~Torralba, and P.~Sharma.
\newblock {LoRA} vs full fine-tuning: An illusion of equivalence.
\newblock In \emph{Advances in Neural Information Processing Systems
  (NeurIPS)}, 2025.

\bibitem[Soylu et~al.(2024)Soylu, Potts, and Khattab]{SoyluD2024emnlp}
D.~Soylu, C.~Potts, and O.~Khattab.
\newblock Fine-tuning and prompt optimization: Two great steps that work better
  together.
\newblock In \emph{Proceedings of the Conference on Empirical Methods in
  Natural Language Processing (EMNLP)}, 2024.

\bibitem[Tiwari et~al.(2026)Tiwari, Sareen, Agrawal, Gonzalez, Zaharia,
  Keutzer, Dhillon, Agarwal, and Khatri]{tiwari2026fastslow}
R.~Tiwari, K.~Sareen, L.~A. Agrawal, J.~E. Gonzalez, M.~Zaharia, K.~Keutzer,
  I.~S. Dhillon, R.~Agarwal, and D.~Khatri.
\newblock Learning, fast and slow: Towards {LLM}s that adapt continually, 2026.
\newblock arXiv preprint arXiv:2605.12484.

\bibitem[Wu et~al.(2025)Wu, Wang, Yu, Zhang, Chang, and Yu]{WuD2025iclr}
D.~Wu, H.~Wang, W.~Yu, Y.~Zhang, K.-W. Chang, and D.~Yu.
\newblock {LongMemEval}: Benchmarking chat assistants on long-term interactive
  memory.
\newblock In \emph{Proceedings of the International Conference on Learning
  Representations (ICLR)}, 2025.

\bibitem[Xu et~al.(2025)Xu, Liang, Mei, Gao, Tan, and Zhang]{XuW2025neurips}
W.~Xu, Z.~Liang, K.~Mei, H.~Gao, J.~Tan, and Y.~Zhang.
\newblock {A-Mem}: Agentic memory for {LLM} agents.
\newblock In \emph{Advances in Neural Information Processing Systems
  (NeurIPS)}, 2025.

\bibitem[Yang et~al.(2024{\natexlab{a}})Yang, Wang, Lu, Liu, Le, Zhou, and
  Chen]{yang2024opro}
C.~Yang, X.~Wang, Y.~Lu, H.~Liu, Q.~V. Le, D.~Zhou, and X.~Chen.
\newblock Large language models as optimizers.
\newblock In \emph{Proceedings of the International Conference on Learning
  Representations (ICLR)}, 2024{\natexlab{a}}.

\bibitem[Yang et~al.(2024{\natexlab{b}})Yang, Jimenez, Wettig, Lieret, Yao,
  Narasimhan, and Press]{YangJ2024neurips}
J.~Yang, C.~E. Jimenez, A.~Wettig, K.~Lieret, S.~Yao, K.~Narasimhan, and
  O.~Press.
\newblock {SWE-agent}: Agent-computer interfaces enable automated software
  engineering.
\newblock In \emph{Advances in Neural Information Processing Systems
  (NeurIPS)}, 2024{\natexlab{b}}.

\bibitem[Yao et~al.(2023)Yao, Zhao, Yu, Du, Shafran, Narasimhan, and
  Cao]{yao2023react}
S.~Yao, J.~Zhao, D.~Yu, N.~Du, I.~Shafran, K.~Narasimhan, and Y.~Cao.
\newblock {ReAct}: Synergizing reasoning and acting in language models.
\newblock In \emph{Proceedings of the International Conference on Learning
  Representations (ICLR)}, 2023.

\bibitem[Yu et~al.(2018)Yu, Zhang, Yang, Yasunaga, Wang, Li, Ma, Li, Yao,
  Roman, Zhang, and Radev]{YuT2018emnlp}
T.~Yu, R.~Zhang, K.~Yang, M.~Yasunaga, D.~Wang, Z.~Li, J.~Ma, I.~Li, Q.~Yao,
  S.~Roman, Z.~Zhang, and D.~Radev.
\newblock {Spider}: A large-scale human-labeled dataset for complex and
  cross-domain semantic parsing and text-to-{SQL} task.
\newblock In \emph{Proceedings of the Conference on Empirical Methods in
  Natural Language Processing (EMNLP)}, 2018.

\bibitem[Yu et~al.(2019)Yu, Zhang, Yasunaga, Tan, Lin, Li, Er, Li, Pang, Chen,
  Ji, Dixit, Proctor, Shim, Kraft, Zhang, Xiong, Socher, and Radev]{YuT2019acl}
T.~Yu, R.~Zhang, M.~Yasunaga, Y.~C. Tan, X.~V. Lin, S.~Li, H.~Er, I.~Li,
  B.~Pang, T.~Chen, E.~Ji, S.~Dixit, D.~Proctor, S.~Shim, J.~Kraft, V.~Zhang,
  C.~Xiong, R.~Socher, and D.~Radev.
\newblock {SParC}: Cross-domain semantic parsing in context.
\newblock In \emph{Proceedings of the Annual Meeting of the Association for
  Computational Linguistics (ACL)}, 2019.

\bibitem[Y{\"u}ksekg{\"o}n{\"u}l et~al.(2025)Y{\"u}ksekg{\"o}n{\"u}l, Bianchi,
  Boen, Liu, Lu, Huang, Guestrin, and Zou]{yuksekgonul2025textgrad}
M.~Y{\"u}ksekg{\"o}n{\"u}l, F.~Bianchi, J.~Boen, S.~Liu, P.~Lu, Z.~Huang,
  C.~Guestrin, and J.~Zou.
\newblock Optimizing generative {AI} by backpropagating language model
  feedback.
\newblock \emph{Nature}, 639:\penalty0 609--616, 2025.

\bibitem[Zhang et~al.(2026)Zhang, Ergen, Logeswaran, Lee, and
  Jurgens]{ZhangL2026iclr}
L.~Zhang, T.~Ergen, L.~Logeswaran, M.~Lee, and D.~Jurgens.
\newblock {SPRIG}: Improving large language model performance by system prompt
  optimization.
\newblock In \emph{Proceedings of the International Conference on Learning
  Representations (ICLR)}, 2026.

\bibitem[Zhang et~al.(2022)Zhang, Cutkosky, and
  Paschalidis]{zhang2022switching}
Z.~Zhang, A.~Cutkosky, and I.~Ch. Paschalidis.
\newblock Optimal comparator adaptive online learning with switching cost.
\newblock In \emph{Advances in Neural Information Processing Systems
  (NeurIPS)}, 2022.

\bibitem[Zhong et~al.(2020)Zhong, Yu, and Klein]{ZhongR2020emnlp}
R.~Zhong, T.~Yu, and D.~Klein.
\newblock Semantic evaluation for text-to-{SQL} with distilled test suites.
\newblock In \emph{Proceedings of the Conference on Empirical Methods in
  Natural Language Processing (EMNLP)}, 2020.

\bibitem[Zhou et~al.(2023)Zhou, Muresanu, Han, Paster, Pitis, Chan, and
  Ba]{zhou2023ape}
Y.~Zhou, A.~I. Muresanu, Z.~Han, K.~Paster, S.~Pitis, H.~Chan, and J.~Ba.
\newblock Large language models are human-level prompt engineers.
\newblock In \emph{Proceedings of the International Conference on Learning
  Representations (ICLR)}, 2023.

\end{thebibliography}
